# Deep Convolutional Neural Networks for Short-Term Multi-Energy Demand Prediction of Integrated Energy Systems

Corneliu Arsene , Alessandra Parisio
*Department of Electrical and Electronic Engineering*
*The University of Manchester*
Manchester, United Kingdom
CorneliuArsene@gmx.com, ArseneCorneliu@tutanota.de, Alessandra.Parisio@manchester.ac.uk

***Abstract*—Forecasting power consumptions of integrated electrical, heat or gas network systems is essential in order to operate more efficiently the whole energy network. Multi-energy systems are increasingly seen as a key component of future energy systems, and a valuable source of flexibility, which can significantly contribute to a cleaner and more sustainable whole energy system. Therefore, there is a stringent need for developing novel and performant models for forecasting multi-energy demand of integrated energy systems, which to account for the different types of interacting energy vectors and of the coupling between them. Previous efforts in demand forecasting focused mainly on the single electrical power consumption or, more recently, on the single heat or gas power consumptions. In order to address this gap, in this paper six novel prediction models based on Convolutional Neural Networks (CNNs) are developed, for either individual or joint prediction of multi-energy power consumptions: the single input/single output CNN model with determining the optimum number of epochs (CNN_1), the multiple input/single output CNN model (CNN_2), the single input/ single output CNN model with training/validation/testing datasets (CNN_3), the joint prediction CNN model (CNN_4), the multiple-building input/output CNN model (CNN_5) and the federated learning CNN model (CNN_6). All six novel CNN models are applied in a comprehensive manner on a novel integrated electrical, heat and gas network system, which only recently has started to be used for forecasting. The forecast horizon is short-term (next half an hour) and all the predictions results are evaluated in terms of the Signal to Noise Ratio (SNR) and the Normalized Root Mean Square Error (NRMSE), while the Mean Absolute Percentage Error (MAPE) is used for comparison purposes with other existent results from literature. The numerical results show that the single input/single output CNN model with training/validation/testing datasets (CNN_3) is able to equal the performances of the single input/single output CNN model with determining the optimum number of epochs (CNN_1), and to outperform the other four prediction models. The prediction accuracy of the multi-energy buildings loads is shown to significantly depend on the level of non-linearity and scarcity existent in the input training dataset(s). Furthermore, this extensive multi-model study reveals that the characteristics (i.e. connections between the different networks, correlations between the different energy vectors) of the considered integrated system need to be explored when designing the predictive models.**

**Keywords—Multi-Energy vector, Deep Learning models, Convolutional Neural Networks, Integrated Energy Systems, Electrical-Heat-Gas Networks, Federated Learning**

## I. Introduction

Predicting the future state (i.e. power load consumption) of network systems is a key point in the operation of any such networks (e.g. electrical, heat) as it would allow the most appropriate decisions to be taken at the right time and at minimum costs [1]. Predicting the electrical power consumption has been studied by using different prediction models including Deep Learning (DL) models [2-20]. More recently, forecasting of the gas demand has been also implemented by using machine learning models [21-24]. Several methods to predict the heat demand have been also investigated such as Multiple Linear Regression (MLR), Seasonal Autoregressive Integrated Moving Average (SARIMA) [25], support vector machines [26,27], Artificial Neural Network (ANN) [28], wavelet methods [29], genetic algorithm or DL such as Temporal Convolutional Neural Network (TCN) [30].

The simulation and state estimation [33-41] of combined electrical, heat and gas network systems have received attention in the last decade [42-53]. However, the individual prediction (i.e. a single type of energy is predicted) or joint prediction (e.g. simultaneous prediction of electric, heat and gas energy vectors) of power consumption of combined electrical, heat and gas network systems is still at the beginning [54-60]. Therefore, there is a strong need to further develop and investigate predictive models, so that to gain a deeper understanding of the coupling and the interdependence between the different energy vectors (e.g. electric, heat or gas), their impact on forecasting as well as the need for forecasting the multiple-energy demands vectors, and not just one type of energy vector. One or multiple buildings are usually served by a single node of an electric, heat and/or gas network. Multiple interconnected nodes form the respective electric, heat or gas network, while multi-demand networks form an integrated energy system.

Furthermore, not only the literature covering the forecasting of multi-energy demand is not developed yet, but just a few relevant models are commonly explored and compared (i.e. maximum two [55] or three [57]). For example, cooling/heating loads were predicted in [54] by using a multi-layer perceptron neural network. In [55] for single output forecasting of electric, cooling and heating loads, there were used static and non-linear ARX (Autoregressive Model with Exogenous Inputs). A model based on Deep Belief Networks (DBN) was developed in [56] for individual electric, heat and gas prediction of an industrial park, without considering the different energy network interconnections. In [57], there were compared mainly three models, the Long Short-Term Memory (LSTM) model, the NBeats model and a combination of a

The author would like to thank the Engineering and Physical Sciences Research Council (EPSRC) grants no. EP/T021969/1 (Multi-energy Control of Cyber-Physical Urban Energy System), EP/S00078X/2 (Supergen Energy Networks hub 2018) and EP/Y016114/1 (Supergen Energy Networks Impact Hub 2023) for supporting this work.

Temporal Convolutional Network (TCN) and the NBeats model for prediction of electric, cooling and heating loads but without gas prediction. In [58] it was compared a two-layer LSTM model with two-layer Convolutional Neural Network (CNN) model and an ANN model on simulated data covering over 900 cities from United States, but again without accounting for the physical network connections between the different energy networks. In [59] multiple-load forecasting of electric, heating and cooling loads based on a particular Bi-directional LSTM (BiLSTM) model were proposed.

The CNN models have shown to obtain very good performances when predicting time sequence data within several domains [31,32] due to their very large parallelism and approximation capabilities including in power systems research. Therefore, there is a high interest in developing novel and performant CNN models whose structures and the way of applying will be different from the approaches in [57-59]. As a consequence, we propose to develop different CNN architectures from what has been implemented until now, which will result in six advanced CNN models and structures for single or joint prediction of multi-energy demands, and by taking also into account the network connections.

Furthermore, the conclusions drawn in the previous works [56-59] were usually for a given integrated energy system and were based on a rather reduced number of predictive methods, which makes it not easy to draw more general conclusions with regard to the forecasting of multi-energy demands in a more larger context and not only with regard to a specific example. Such specific examples of integrated energy systems can have different sizes (e.g. number of nodes, pipes, etc), the power loads can be simulated or real loads, or the buildings forming a network can have different characteristics (e.g. restaurants, homes, etc). Within this context, it can be underlined that in this paper the integrated energy system on which the forecasting models are developed and applied is new and it has just very recently started to be investigated for forecasting the multi-energy demands

It will be developed novel CNN-based models for individual and joint prediction, and it will be provided also several interesting guidelines for the design of such predictive models, which will need to be devised based on the following main factors:

(i) Type and number of input variables;
(ii) Type and number of output variables;
(iii) Most effective combination of input and output variables of CNN models in order to take into account the multi-energy networks interconnections (e.g. multiple input energy variables corresponding to energy networks interconnections), the size of integrated energy systems (e.g. number of nodes, pipes);
(iv) Optimal CNN model architectures, which would involve deciding on the number of convolutional hidden layers, the number of filters and kernel sizes, the size of the images input layer, the size of regression layer output;
(v) Type of learning of parameters of CNN models, which can be centralized or distributed (e.g. federated learning);
(vi) Most effective combination of all the factors listed above to increase the accuracy of prediction results and decrease of the computational times.

The developed novel CNN models are listed below and will be fully described the following sections:

1) Single input variable and single output (i.e. individual prediction) variable CNN model and based on the determination of the number of optimization epochs obtained by simulation (CNN_1: section IV.B1);
2) Multiple input variables and single output (i.e. individual prediction) variable CNN model (CNN_2: section IV.B2);
3) Single input variable and single output (i.e. individual prediction) variable CNN model together with using training/validation/testing datasets (CNN_3: section IV.B3);
4) Multiple input variables and multiple output (i.e. joint prediction) variables CNN model (CNN_4: section IV.B4);
5) Multiple-building input/output variable (i.e. simultaneous multiple building predictions) CNN model, which can be also called single CNN model for a single energy network (CNN_5: section IV.B5);
6) Federated learning for an integrated electric, heat and gas networks system (CNN_6: section IV.B6);

The federated learning is a type of distributed learning, which will be also investigated for multi-energy building prediction.

Before developing the new forecasting models, it is of high interest to identify the input variables of these CNN models based on the input variables prognostic importance. Therefore, correlations between the next and the previous 24 hours power consumptions, the solar radiance and the weather temperature ($^{\circ}$C) data will be calculated for each building and each type of energy vector (electric, heat and gas). For the federated learning, it will be calculated the correlation coefficients between the power consumptions of the different buildings, which have non-zero power consumptions, for the electric, the gas and the heat energy vectors. Based on these correlation studies, it will be possible to select the input variables for the different CNN models.

The paper is structured as follows: the datasets are described in section II. In section III, the CNN models and the other computational methods used in the paper are presented. In section IV the results are shown. The results section is structured in several subsections: section A presents the selection of the input variables for the CNNs models, while section B presents the short-term predictions. Finally, conclusions and directions for future work are discussed.

## II. DATASETS

In this work, the multi-energy buildings predictions (next half an hour) of a combined electrical, heat and gas network system will be done either as an individual prediction (i.e. single output variable of DL model) of the multi-energy vectors (electrical, heat and gas) [57], or as a joint prediction (i.e. multiple output variables of DL model) of the multi-energy vectors [56,57]. There can be used a large number of relevant input variables such as the weather, the time and/or the previous power consumptions usually covering the previous 24 hours. Although this will be discussed later in the results section, it can be mentioned here that we identified the most important input variables as the previous 24 hours electrical, heat and gas power consumption.

From the point of view of the number of input variables of the CNN models, there could be identified two cases. The first case consists of using a single input variable of the same type as the predicted output variable and covering the previous 24 hours power consumptions. For example, if the predicted output variable is the heat power consumption, then the input variable is the previous 24 hours heat power consumption. The second case is when using multiple input variables for a CNN model, such as when there exists one or two network connections between the different electrical, heat and gas networks. In this second case the multiple input variables are including an input variable of the same type as the predicted output variable and covering the previous 24 hours power consumptions, and also the other previous 24 hours power consumptions for which there exists a network link.

The datasets [42] used in this paper cover the power consumptions (kW) of a multi-energy vector system belonging to the University of Manchester (UoM) in United Kingdom (UK) and consisting of combined electric, heat and gas networks for 39 interconnected buildings (i.e. integrated energy system). Fig.1 shows each of the three electric, heat and gas networks with their network nodes numbering [42]. Fig 2(a) shows the buildings with their numbers, while Fig.2(b) shows the correspondence between the buildings numbers and the electric, the gas and the heating network nodes numbers. For example, building number 1 in Fig.2(a), is connected to node 4 of the electrical network in Fig.1(a), to node 9 of the heat network in Fig.1(b), and to node 32 of the gas network in Fig. 1(c). In this work, it is predicted the next half an hour electric, heat and gas power consumptions for each of the 39 buildings. Fig.3 shows how the CNN predicted power consumption of each building and of each type of energy vector (electric, heat or gas) are summed up over all 39 buildings (Fig. 3) to calculate the total power consumption of each energy network, which can stretch longer than half an hour (e.g. 2 months).

The CNN models are obtained by using three datasets (electric, heat and gas), which contain power consumptions over one year (i.e. 2013) and the data is available at each half an hour, which for a day results in 48 data points, and in total for the full year there are 17,520 data points. Our work is to predict the next half an hour power consumptions for each of the 39 buildings and for each type of power consumption (electric, heat and gas), and then to apply the CNN models for longer periods of time.

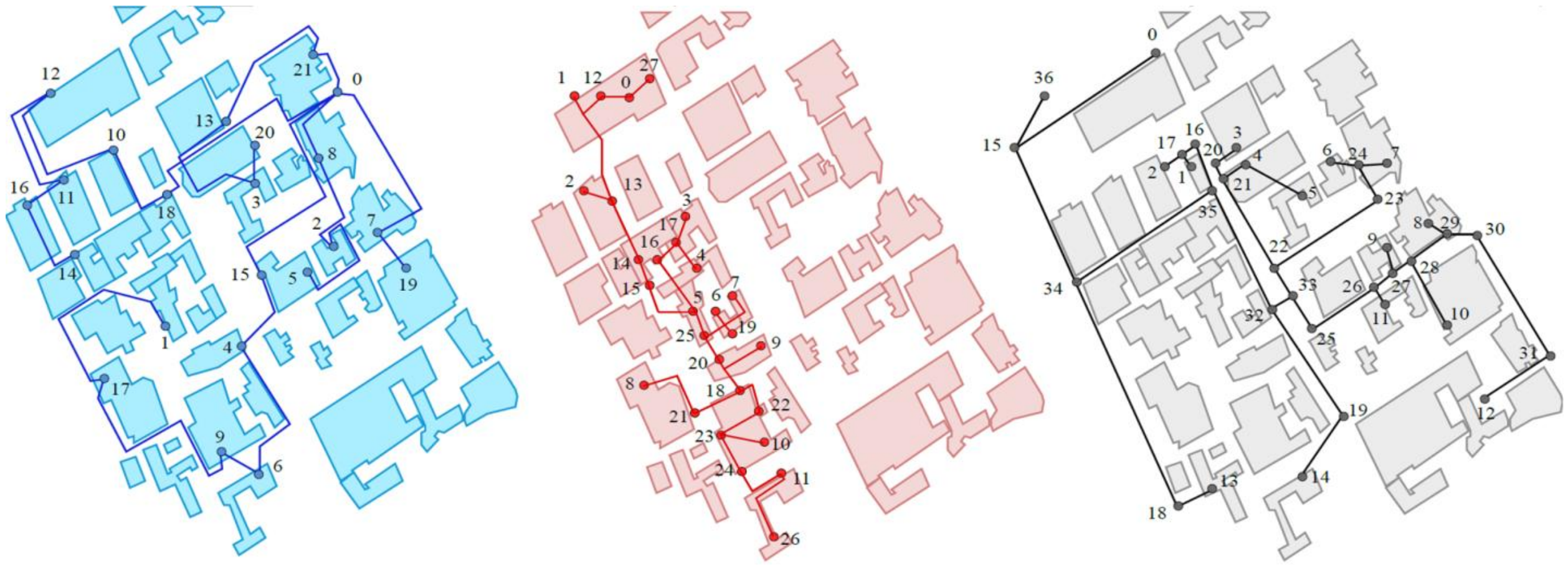

a) b) c)

Fig.1. Integrated energy system consisting of three combined electric, heat and gas networks of UoM where the numbers represent the network nodes: a) electric network; b) heat network; c) gas network, as shown in [42].

The chosen study period of time is the cold and the beginning of spring season, which is January, February and March (i.e. more power consumptions data available during this time period): average mean daily temperatures were similar during the three months 5.27ºC (January), 5.51ºC (February) and 5.15ºC (March), while average mean nightly temperatures were also close 1.57 ºC (January), 0.26ºC (February) and -0.8ºC (March).

The network connections are available for each building and therefore it is possible to know at which nodes the networks connect each other. There is also a slack node in each network denoted as 0 where there are connected the electricity and the gas supply points as well as the largest boiler for the heat network. This multi-energy vector system may take electrical energy from the electrical network and produce heat (i.e. EHP-electrical heat pump) or take gas from the gas network and produce heat and hot water (i.e. gas boiler). Fig.4 shows the total power consumptions obtained by summing up the predictions obtained for each of the 39 buildings and for each electric, gas and heat network. In this paper the focus is mainly on the period of time [0 7248], which is January to May.

The total annual electrical demand is 30.23 GW, 35.96 GW for the heat demand and 1.58 GW for the gas demand. The gas power consumptions are smaller than the electrical and the heat consumptions for this combined network system, and it can be expected some difficulties when predicting the gas power consumptions because of their non-linearity, scarcity and reduced (kW) consumption values (Fig.4c).

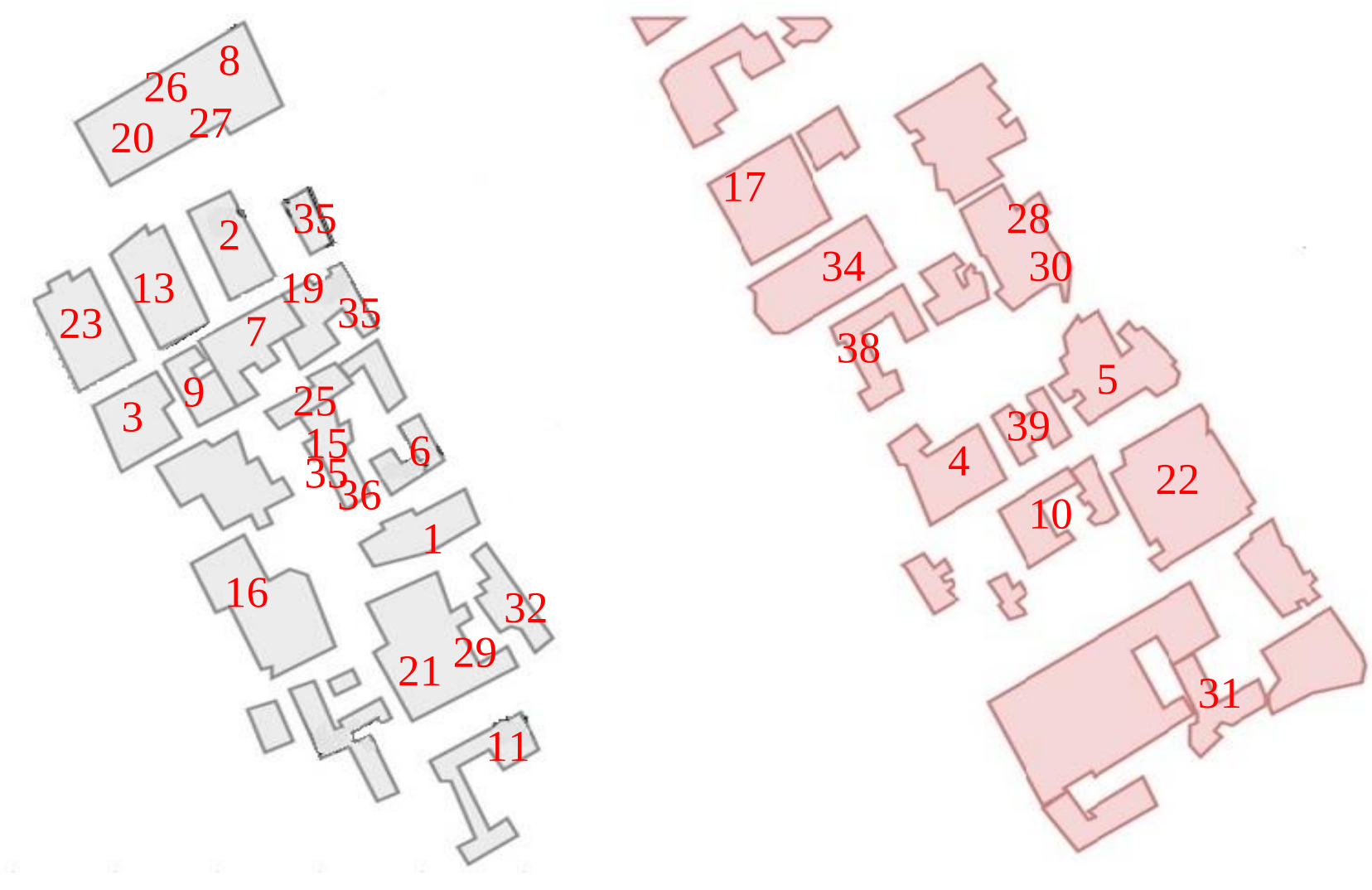

a)

| Building number | Network connection | | |
|---|---|---|---|
| | Electricity | Heat | Gas |
| 1 | 4 | 9 | 32 |
| 2 | 10 | 2 | 2 |
| 3 | 14 | 4 | 34 |
| 4 | 5 | - | - |
| 5 | 7 | - | 8 |
| 6 | 1 | 6 | 32 |
| 7 | 14 | 16 | 34 |
| 8 | 12 | 27 | . |
| 9 | 14 | 15 | . |
| 10 | 5 | - | 11 |
| 11 | 6 | 26 | 14 |
| 12 | - | - | . |
| 13 | 11 | 2 | 2 |
| . | . | . | . |
| . | . | . | . |
| . | . | . | . |
| . | . | . | . |

b)

Fig.2. 39 interconnected buildings of UoM: a) Buildings numbers; b) Correspondence between buildings numbers and networks nodes numbers shown in Fig.1 [42].

As there is more data available during the cold season due to the heat and gas power consumptions, the power consumptions from the months of January and February are used for training the CNN models (i.e. training dataset), while the consumptions from month of March are used for testing the CNN models (i.e. testing datasets). From the 39 buildings, 11 buildings have zero electrical power consumptions throughout the year and not only between January and March. There are also 9 buildings with zero heat power consumption for the same period of time and 18 buildings with zero gas power consumptions, as well as another 13 buildings with small gas power consumptions of less than 2000 kW per month. It is assumed also that the data available [42] is cleaned from any errors, but otherwise typical data cleaning procedures could have been applied.

## III. Methods and Convolutional Neural Networks

### *A. Selection of the input variables of the CNN models*

In the context of energy systems, previous literature [61] investigated the selection of the input variables for the predictions models by using various techniques such as correlation study, Kendall rank and Copula functions, Least Absolute Shrinkage and Selection Operator (LASSO). The identified input variables were usually either the previous energy consumptions or both the previous energy consumptions and the weather [61]. The weather data may include the temperature but it could be also the solar radiance or the solar energy or the Ultra-Violet (UV) index depending on the data available. In this work, it will be used the correlation study, which is an established method, in order to select the relevant input variables of the CNN models. It will be studied the correlation between the next 24 hours energy consumptions, which would play the role of the output predicted variables, and the previous 24 hours energy consumptions, which would be the input variables. It will also be investigated the correlation between the next 24 hours energy consumptions and the previous 24 hours weather data consisting of the temperature and the solar radiance. For the federated learning, it is of interest to look to the correlations of the energy vectors between the different buildings. There will be produced two-dimensional correlation matrixes, which will be plotted, and lighter colours will represent high correlations, while darker colours will represent weak correlations. The high correlations will be for to the selected input variables of the CNNs models.

### *B. Convolutional Neural Networks for short-term power prediction*

As mentioned before, CNN models obtained very good performances when predicting time sequence data and therefore they are chosen here to predict the multi-energy demands. The DL software toolbox from MATLAB is used for this purpose. Depending on the number of input and output variables, the type of DL learning (centralized or distributed), as well as the other factors mentioned earlier in the Introduction, it is possible to devise 6 DL modelling frameworks: single input variable and single output (i.e. individual prediction) variable CNN model with determining the optimum number of epochs by simulation (CNN_1), multiple input variables and single output variable (i.e. individual prediction) CNN model (CNN_2), single input variable and single output (i.e. individual prediction) variable CNN model together with training/validation/testing datasets (CNN_3), multiple input variables and multiple output (i.e. joint prediction) variables CNN model (CNN_4), multiple-building input/output variable (i.e. multiple building prediction for a single energy vector) CNN model (CNN_5), federated learning (distributed learning) based on CNN model (CNN_6). Each of these modelling frameworks is presented in the following sections.

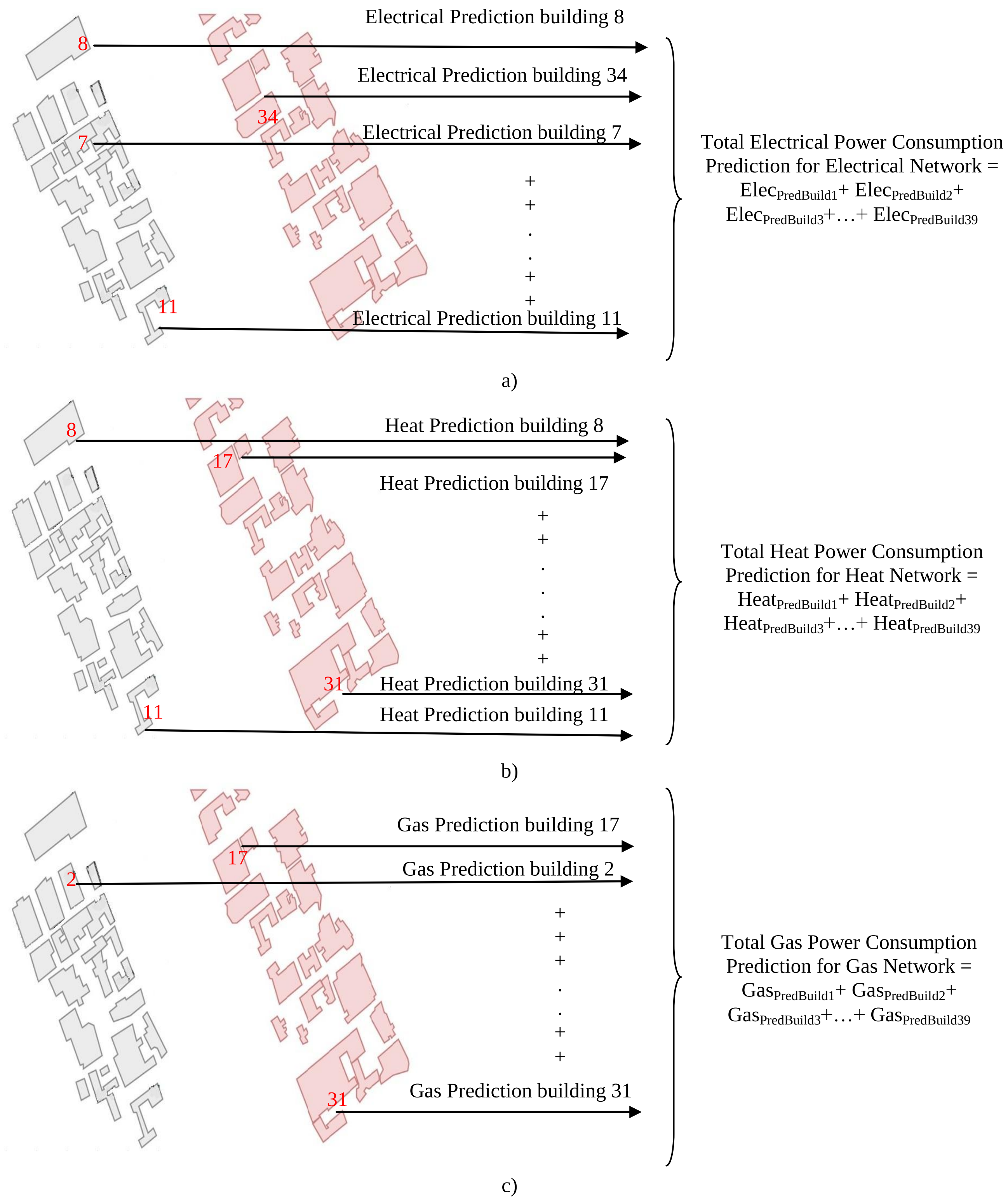

Fig.3. Calculation of the total predicted electrical, heat and gas power network consumptions obtained by summing the predictions obtained for each building: a) electric network; b) heat network; c) gas network; $Elec_{PredBuild1}$ is the electrical power consumption prediction for building no.1, $Heat_{PredBuild1}$ is the heat power consumption prediction for building no.1, $Gas_{PredBuild1}$ is the gas power consumption prediction for building 1.

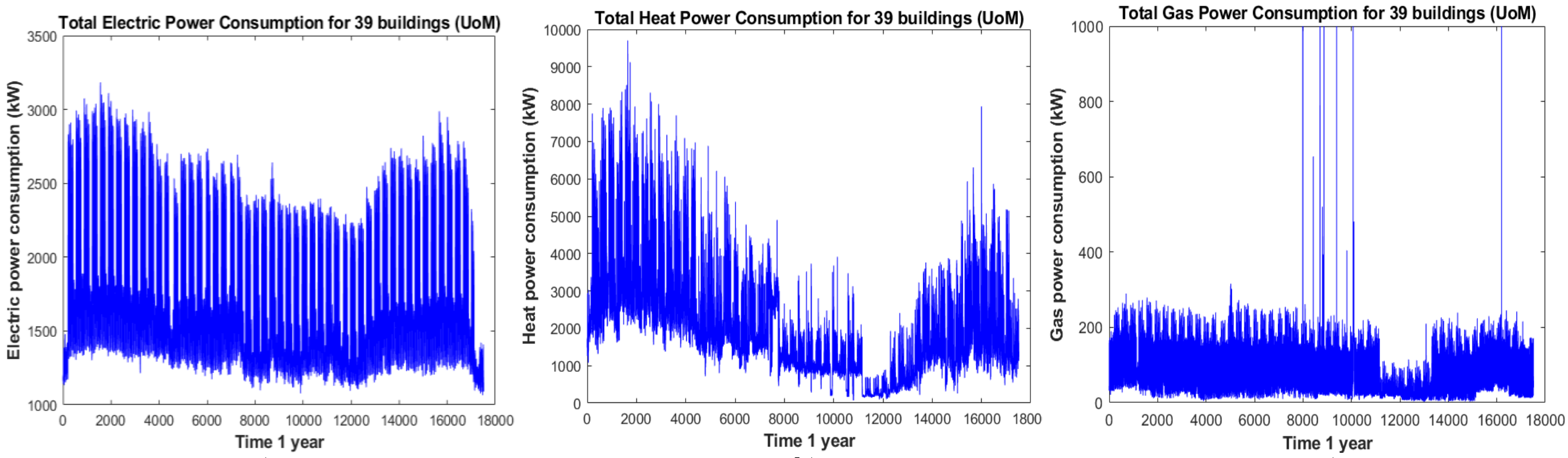

a) b) c)

Fig. 4. Total power consumptions summed up over all 39 buildings during a year: a) electric network; b) heat network; c) gas network (i.e. six gas energy spikes are limited for visualization purposes).

*B.1. Single input variable and single output (i.e. individual prediction) variable CNN model and based on the determination of the number of optimization epochs obtained by simulation (CNN_1)*

CNN_1 model is used to train and to predict either the electrical, the heat or the gas power consumptions for each of the 39 buildings. It has a single input variable and a single output variable. The single input variable represents the previous 24 hours electrical, heat or gas power consumptions while the single output variable is the predicted next half an hour power consumption (i.e. single point value), and which will be either the electrical, the heat or the gas power consumption. It results 39 CNN models corresponding to the 39 buildings belonging to the electric network. Similarly for the heat and the gas networks and therefore there will be a total of 117 CNN models covering all the 39 buildings and all the three energy vectors.

The months January and February are used as the training dataset, which results in a training matrix of size [48 1 1 2928] (i.e. 66% of all input data). The testing matrix representing month of March has a size of [48 1 1 1488] (i.e. 33% of all input data).

The CNN_1 model includes three 2-dimensional convolutional layers, each having 136 filters with kernel size of 146x1 per filter (Fig.5a). The first layer is an input layer of size 48x1x1 with 'zerocenter' normalization and therefore $M$ equals 48 is the number of samples per input signal sequence. The neurons of each convolutional layer connect to parts of the input feature or connect to the outputs of the previous layer. The step size (i.e. stride) for the kernels is [1 1] while the padding is used so that the output is the same size as the input. Each convolutional layer is followed by a batch normalization layer with 136 channels, a Rectified Linear Unit (ReLU) layer and an average pooling layer with stride of 4 and pooling size of 1. Before the final regression output layer, the signal goes through a fully connected layer. The CNN model is able to learn suitable filters that are used for predicting the electric, heat or gas power consumptions. The training of the CNN models uses the Adam optimizer with a batch training data size of 700. It is also noticed that the performance of the CNN models does not change much with increasing batch sizes. A constant learning rate of 0.01 and a gradient threshold of 'inf' are used. During the training, an epoch goes through the entire dataset, while an iteration is the calculation of the gradient and the network parameters for the mini-batch data. Tables I details all the 15 layers of CNN_1.

The training of the CNN model(s) (i.e. calculation of CNN model parameters) is performed for a maximum number of epochs (i.e. the iterative updating of DL model) so that to avoid the danger of under or overfitting the DL models. The number of epochs is determined by simulation and it is shown in supplementary material (section AA1).

*B.2. Multiple input variable and single output (i.e. individual prediction) variable CNN model (CNN_2)*

CNN_2 model has multiple input variables (e.g. 2 or 3 input variables) and it is used to train and to predict the heat or the gas power consumptions for the buildings where there are located network connections (Fig.5(b)). For example, for building 3, which has a network connection between the electrical, the heat and the gas networks, in order to predict the next half an hour gas power consumption (i.e. single point value), three input variables are used representing the previous 24 hours heat power consumptions, the previous 24 hours electrical power consumptions and the previous 24 hours gas power consumptions.

The input training matrix is of size [48 3 1 2928[, while the testing matrix for month March has size [48 3 1 1488]. The first layer is an input image layer of size [48 2 1] for when there are two input variables, or of size [48 3 1] for when there are 3 input variables. There is 'zerocenter' normalization and again M equals 48 is the number of samples per input signal sequence. The next layer is the convolutional layer followed by the normalization layer with 30 channels, a ReLU layer and an average pooling layer with stride of [1 1] and pooling size of 1. Fig.5(b) shows the structure of the CNN_2 model. Tables II details all 15 layers of CNN_2 model. The same as in the previous section, the number of optimization epochs is determined by simulation so that to avoid over or underfitting.

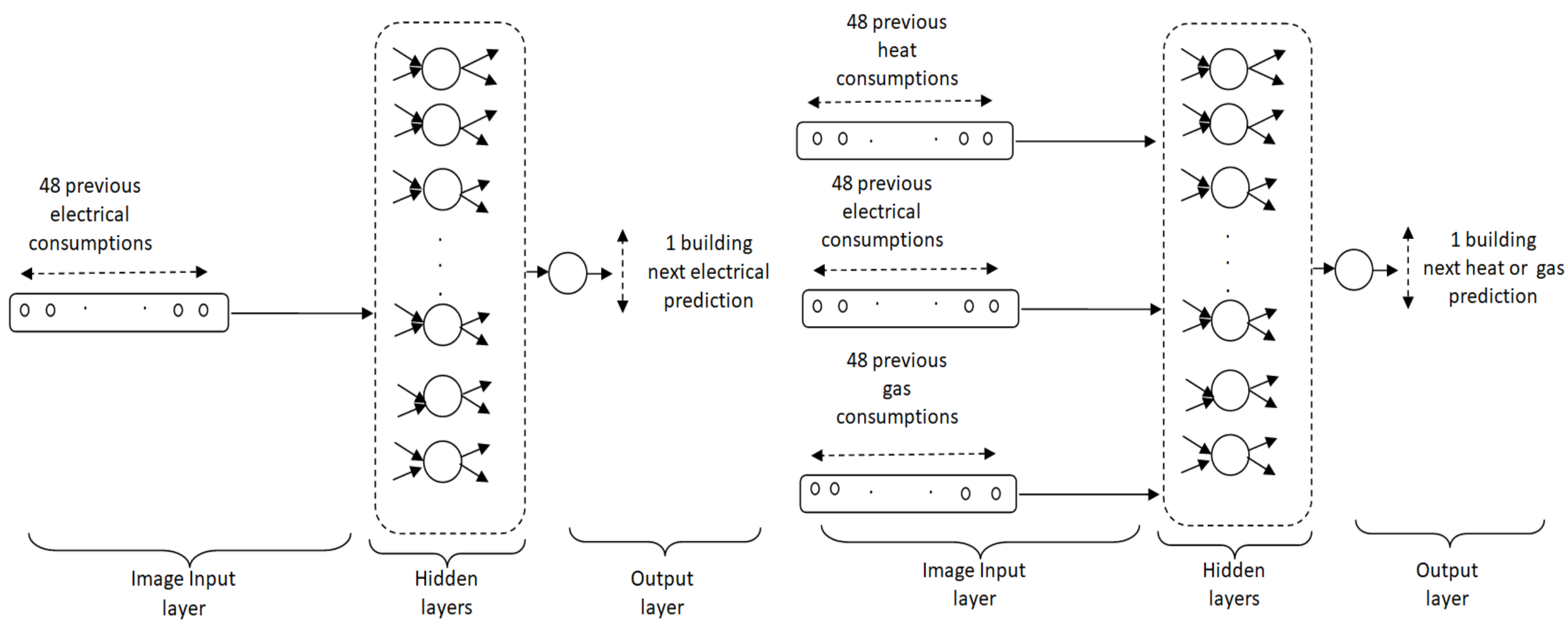

(a) (b)

Fig. 5. Structure of the CNN models: a) CNN_1 model with single input variable and single output variable; b) CNN_2 model with multiple input variables and single output variable: 48 stands for the previous 24 hours electric power consumptions multiplied by two half hours; 117 different CNN models are necessary to determine the total electric, heat and gas energy consumption for the 39 buildings.

TABLE I. LISTING OF THE CNN LAYERS FOR MODEL CNN_1: *M = 48* IS THE NUMBER OF SAMPLES PER INPUT SIGNAL

| nr | name and type | activations | learnable |
|---|---|---|---|
| 1 | Imageinput: 48x1x1 images with 'zerocenter' normalization | 48x1x1x1 | - |
| 2 | conv_1: 136 146x1x1 convolutions with stride [1 1] and padding 'same' | 48x1x136x1 | Weights: 146x1x1x136 Bias: 1x1x136 |
| 3 | batchnorm_1: Batch normalization with 136 channels | 48x1x136x1 | Offset 1x1x136 Scale 1x1x136 |
| 4 | relu_1: ReLu | 48x1x136x1 | - |
| 5 | avgpool_1: 1x1 average pooling with stride [4 4] and padding [0 0 0 0] | 12x1x136x1 | - |
| 6 | conv_2: 136 146x1x136 convolutions with stride [1 1] and padding 'same' | 12x1x136x1 | Weights 146x1x136x136 Bias: 1x1x136 |
| 7 | batchnorm_2: Batch normalization with 136 channels | 12x1x136x1 | Offset 1x1x136 Scale 1x1x136 |
| 8 | relu_2: ReLu | 12x1x136x1 | - |
| 9 | avgpool_2: 1x1 average pooling with stride [4 4] and padding [0 0 0 0] | 3x1x136x1 | - |
| 10 | conv_3: 136 146x1x136 convolutions with stride [1 1] and padding 'same' | 3x1x136x1 | Weights 146x1x136x136 Bias: 1x1x136 |
| 11 | batchnorm_3: Batch normalization with 136 channels | 3x1x136x1 | Offset 1x1x136 Scale 1x1x136 |
| 12 | relu_3: ReLu | 3x1x136x1 | - |
| 13 | avgpool_3: 1x1 average pooling with stride [4 4] and padding [0 0 0 0] | 1x1x136x1 | - |
| 14 | fc: 1 fully connected layer | 1x1x1x1 | Weights 1x136 Bias: 1x1 |
| 15 | regressionoutput: mean-squared-error with response | 1x1x1x1 | |

TABLE II.
LISTING OF THE CNN LAYERS FOR MODEL CNN_2: *M = 48* IS THE NUMBER OF SAMPLES PER INPUT SIGNAL

| nr | name and type | Activations | learnable |
|---|---|---|---|
| 1 | Imageinput: 48x3x1 images with 'zerocenter' normalization | 48x3x1x1 | - |
| 2 | conv_1: 30 100x1x1 convolutions with stride [1 1] and padding 'same' | 48x3x30x1 | Weights: 100x1x1x30 Bias: 1x1x30 |
| 3 | batchnorm_1: Batch normalization with 30 channels | 48x3x30x1 | Offset 1x1x30 Scale 1x1x30 |
| 4 | relu_1: ReLu | 48x3x30x1 | - |
| 5 | avgpool_1: 1x1 average pooling with stride [1 1] and padding [0 0 0 0] | 48x3x30x1 | - |
| 6 | conv_2: 30 100x1x30 convolutions with stride [1 1] and padding 'same' | 48x3x30x1 | Weights 100x1x30x30 Bias: 1x1x30 |
| 7 | batchnorm_2: Batch normalization with 30 channels | 48x3x30x1 | Offset 1x1x30 Scale 1x1x30 |
| 8 | relu_2: ReLu | 48x3x30x1 | - |
| 9 | avgpool_2: 1x1 average pooling with stride [1 1] and padding [0 0 0 0] | 48x3x30x1 | - |

| nr | name and type | Activations | learnable |
|---|---|---|---|
| 1 | Imageinput: 48x3x1 images with 'zerocenter' normalization | 48x3x1x1 | - |
| 10 | conv_3: 30 100x1x30 convolutions with stride [1 1] and padding 'same' | 48x3x30x1 | Weights 100x1x30x30 Bias: 1x1x30 |
| 11 | batchnorm_3: Batch normalization with 30 channels | 48x3x30x1 | Offset 1x1x30 Scale 1x1x30 |
| 12 | relu_3: ReLu | 48x3x30x1 | - |
| 13 | avgpool_3: 1x1 average pooling with stride [1 1] and padding [0 0 0 0] | 48x3x30x1 | - |
| 14 | fc: 1 fully connected layer | 1x1x1x1 | Weights 1x4320 Bias: 1x1 |
| 15 | regressionoutput: mean-squared-error with response | 1x1x1x1 | |

*B.3 Single input variable and single output (i.e. individual prediction) variable CNN model together with using training/validation/testing datasets (CNN_3)*

The standard training method used to avoid overfitting is by employing validation datasets in addition to the training and the testing datasets [62-64]. The validation datasets ensure that robust DL models are obtained by updating iteratively the hyperparameters of the DL model during the training. Fig. 6 shows the usual training of DL models based on the training, validation and testing datasets. The use of validation datasets in the training stage could be a bit cumbersome as it requires additional datasets or to reduce the size of the training and the testing datasets.

In this work, the validation datasets have the size corresponding to the power consumptions covering 10 days [48 480], which is about 10% of all data (i.e. training data, validation data and testing data). The training data will be 60% of all data and the testing data will be 30% of all data [65]. This will form also the third CNN modelling framework denoted as CNN_3.

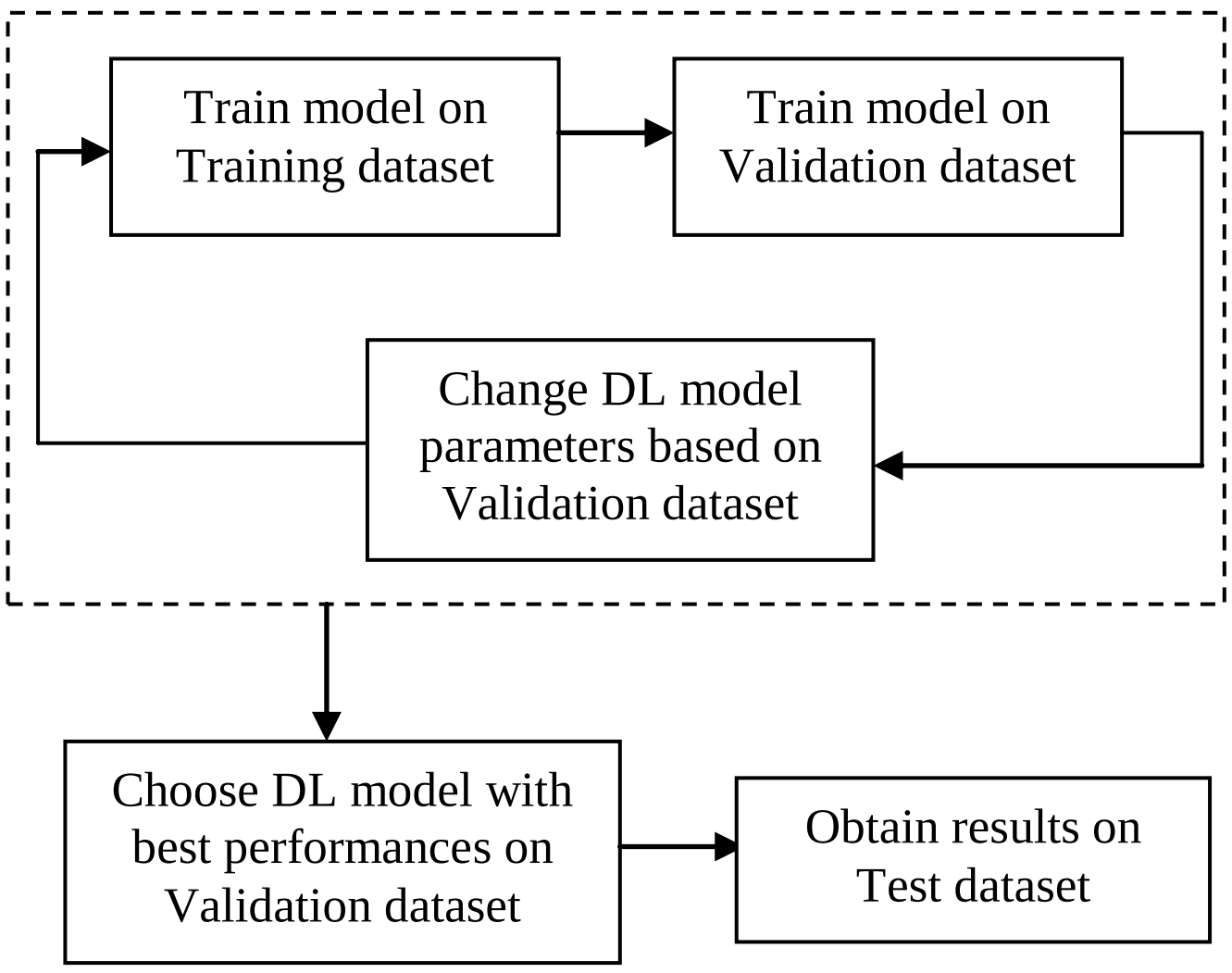

Fig. 6. Training, validation and test of DL/CNN models.

Three metrics will be used to check the accuracy of all the results:

$$NRMSE = \frac{\left(\sqrt{\frac{1}{N}\sum_{n=0}^{N}[y_{pred}(n)-y_{output}(n)]^2}\right)}{\max(y_{output}(n))} \quad (1)$$

$$SNR = 10log_{10}\frac{\sum_{n=1}^{N}[y_{output}(n)]^2}{\sum_{n=1}^{N}[y_{pred}(n)-y_{output}(n)]^2} \quad (2)$$

$$MAPE = \frac{1}{N}\sum_{n=1}^{N} abs\left(\frac{y_{output}(n)-y_{pred}(n)}{y_{output}(n)}\right)x100\% \quad (3)$$

where NRMSE is the Normalized Root Mean Square Error to the maximum value of the real output signal, SNR is the Signal to Noise Ratio, MAPE is the Mean Absolute Percentage Error, *abs* is the absolute value function, $y_{output}$ is the real output signal, $y_{pred}$ is the predicted output signal, *n* is an index, *N* is the total number of sample points of the output signal with a fixed size of 48 representing 24 hours (i.e. 24 hours times two half hours), *max* is the maximum value. An SNR higher than 8 dB and NRMSE smaller than 0.15, would be desirable [66].

*B.4. Multiple input variables and multiple output (i.e. joint prediction) variables CNN model (CNN_4)*

The fourth framework (CNN_4) will use a single CNN model with multiple inputs and multiple outputs, which is also called joint prediction. CNN_4 has an input image layer of size [48 39 3] and an output layer of size [1 117]. Therefore, it includes in the training data all the electric, heat and gas data covering two months of data (i.e. January and February) of the 39 interconnected buildings.

The single CNN_4 model predicts simultaneously the electric, the heat and the gas power consumptions for all 39 buildings and it is depicted in Fig.7.

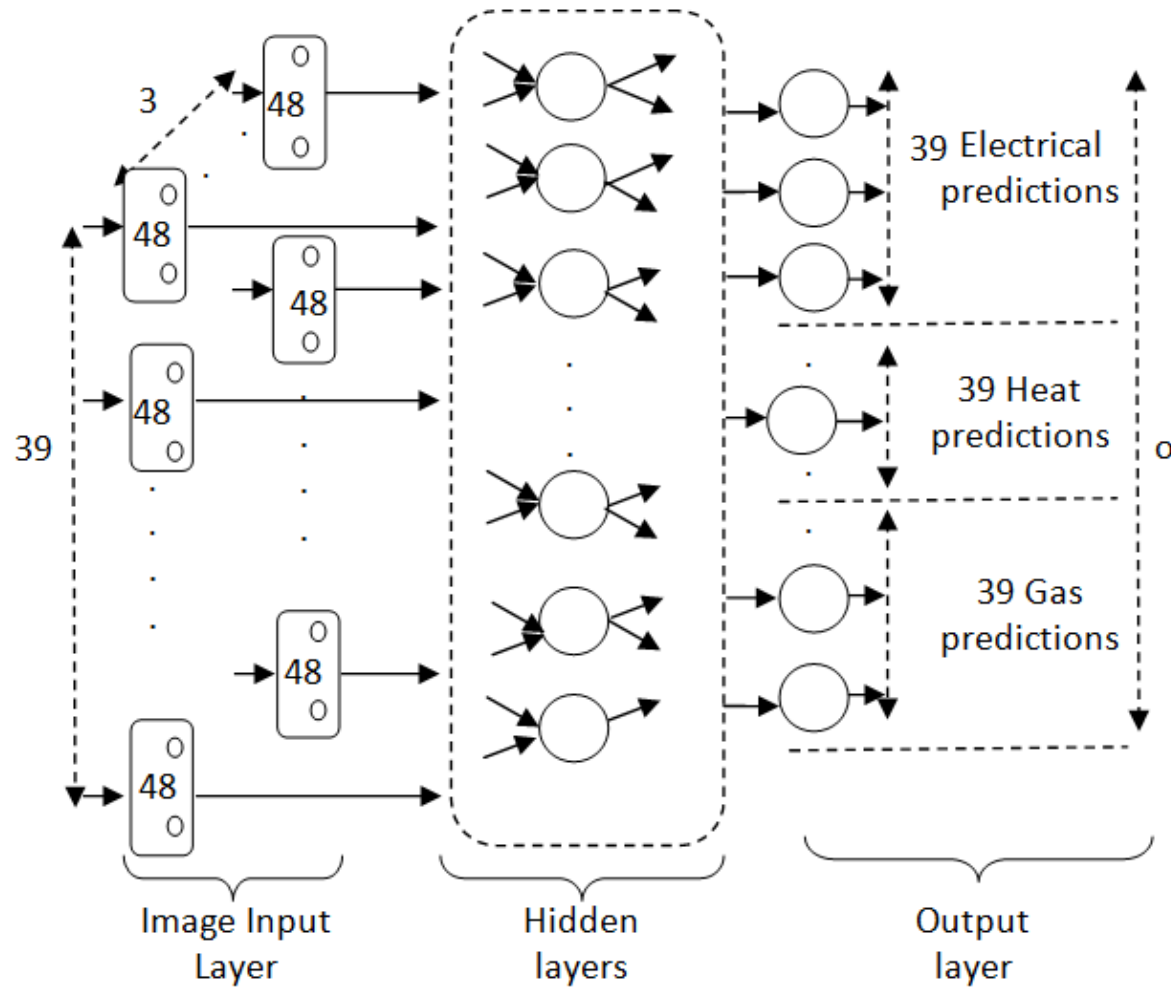

Fig.7. Single CNN_4 model with multiple inputs [48 39 3] and multiple outputs [1 117] predicting simultaneously the electric, the heat and the gas power consumptions for all 39 buildings and for all the three energy vectors (i.e. joint prediction).

There are two convolutional layers in the hidden layers. The input data size (i.e. [48 39 3]) corresponds to the 48 time intervals (i.e. the previous 24 hours multiplied by 2 half hours), the 39 interconnected buildings and the 3 types of energy vectors (electric, heat and gas). The size of the output regression layer (i.e. [1 117]), which corresponds to the power consumption predictions of the next half hours for each of the 39 buildings and for each of the three types of energy vectors (electric, heat and gas energy), equals to 117 outputs of CNN_4 model (i.e. 39 interconnected buildings multiplied by 3 types of energy vectors).

*B.5. Multiple-building input/output variable (i.e. simultaneous multiple building prediction) CNN model: single CNN model for a single energy network (CNN_5)*

The fifth framework (CNN_5) has a CNN model for each type of energy vector and is covering all the 39 buildings (i.e. multiple-building prediction). This results in a total of three different CNN_5 models. Each such CNN_5 model has an image input layer of size [48 39 1], where 48 stands again for the training data covering the previous 24 hours and multiplied by 2 half hours, and 39 stands for the 39 interconnected buildings. The regression output layer corresponding to a single energy network has the size [1 39], where 39 stands again for the 39 buildings (Fig.8). Therefore, the input data for each CNN model (i.e. [48 39 1]) is of the same type (e.g. electric-electric) as the output predicted data (i.e. electric network prediction of size [1 39]). There are used two convolutional layers in the hidden layers, each convolutional layer with 136 filters, and each filter of size [146 1].

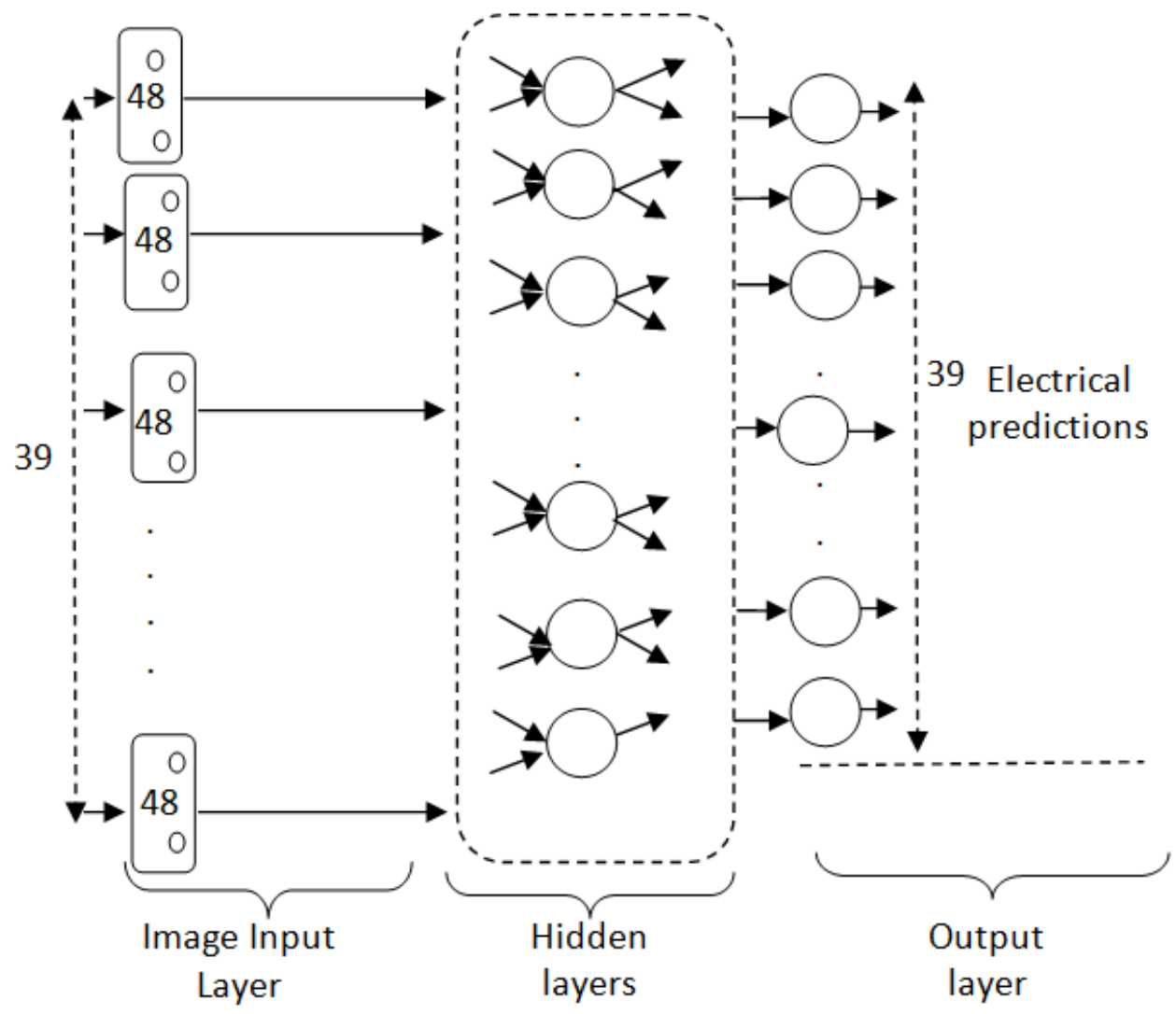

Fig. 8. CNN model (CNN_5) with multiple inputs [48 39 1] and multiple outputs [1 39] predicting for example simultaneously the electric power consumptions for all 39 buildings (i.e. multiple-building prediction).

*B.6. Federated Learning for combined energy systems (CNN_6)*

Federated learning is a method that trains a DL model in a distributed mode [67,68]. Federated learning consists of training a DL model without moving all the data to a central computer, and in addition the separate data sources do not have to meet the overall probability distribution of the entire dataset. This type of dataset is called non-Independent and Identically Distributed (non-IID) data. Federated learning can be used when the training data is big in size, or when there are security problems with moving the training data to a central computer.

In this work (CNN_6), federated learning can be used to train multiple local CNNs models at each node of an energy network (electric, heat or gas). Then, the weights of a global CNN model are updated at each iteration of the iterative optimization procedure, for example by averaging the local weights calculated by the local CNN models at each node. This is also called average federated learning and the global CNN model will be used to make the predictions.

The average federated learning could be useful, because multiple CNN models can be trained simultaneously while keeping the data at the individual nodes and in the end, a global CNN model will emerge for each energy network system (electric, heat and gas). However, as investigations from literature pointed out [67], the overall optimization procedure which is solved in order to calculate the weights of the global CNN model, has to be a convex function with a global minimum point in order to converge and to produce reliable predictions. This means that in the context of the non-IID datasets, there have to be some bounds on the dissimilarity of the data located to the different network nodes so that to ensure that a global CNN model can be obtained.

Fig.9 shows the electric network (Fig.1a) of our integrated energy system and how the local and the global CNN models are trained with the average federated learning scheme. The number of nodes N equals to 20. The average federated learning of the global CNN model is performed at node 0 ($CNN_0$).

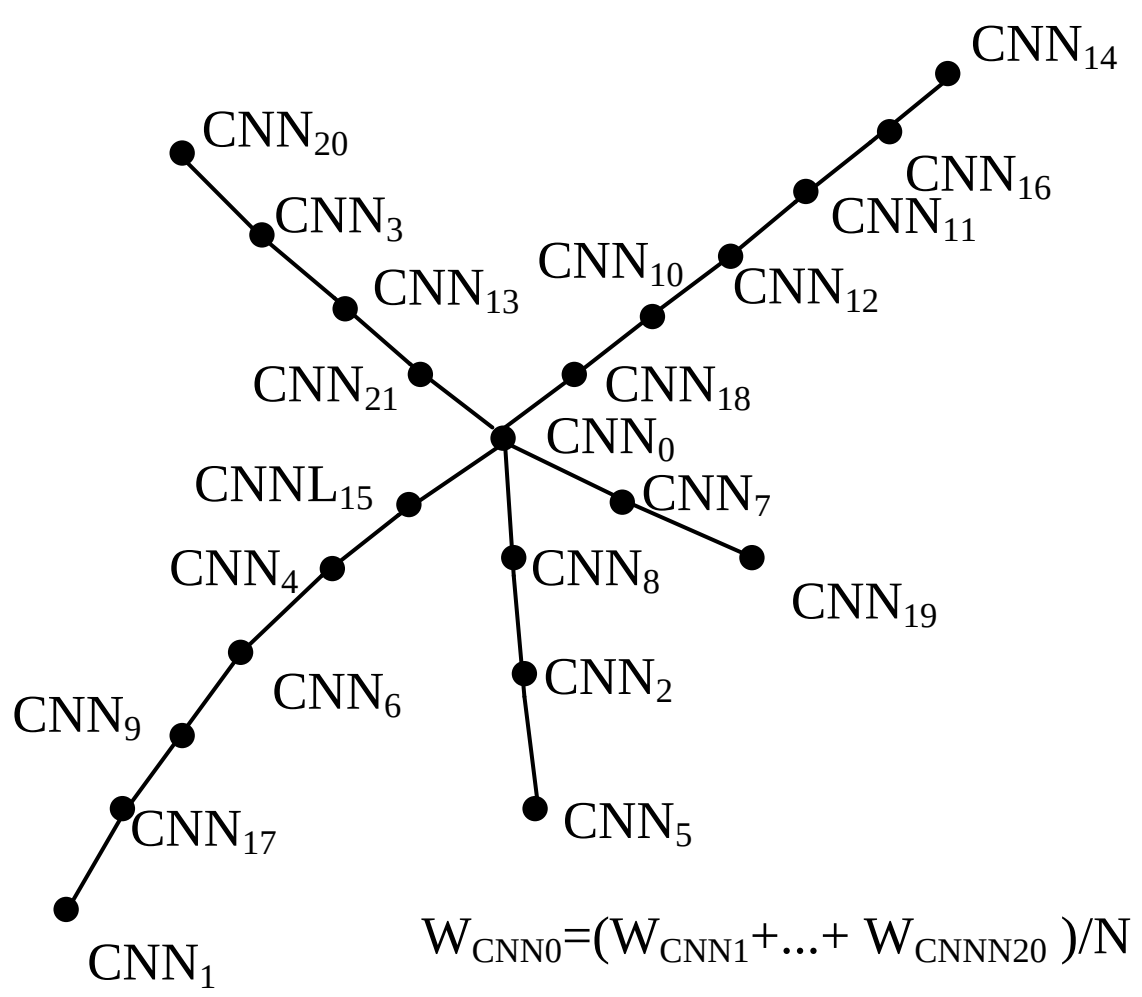

Fig. 9. Electric network shown in Fig.1(a): average federated learning based on CNN models running locally on each electric node (N nodes) for prediction of the electric power consumptions of buildings forming an electric network.

At each iteration of the optimization procedure is performed the averaging of the weights (i.e. $W_{CNN1}$,...,$W_{CNN20}$) of the local CNN models ($CNN_1$ ,…, $CNN_N$) trained at each node for example of the electric network. As already mentioned, this is called average federated learning, and more recently there have been made attempts to improve this type of federated learning [69], although these attempts may still require some bounds on the dissimilarity of the data used to train the local CNN models.

## IV. Results

All simulations are implemented in MATLAB software by using the DL toolbox and on an Intel i7 5.10 GHz processor with 32 GB RAM.

### *A. Selection of the input variables of the CNN models*

Fig.10 shows the temperature (°C) for the city of Manchester (U.K.) between January and May 2013. It can be observed that the temperatures are increasing gradually along March, April and May. Therefore, there can be identified two distinct periods in Fig.10: a colder period for the months January, February and March and a warmer period corresponding to April and May.

For the purpose of the selection of the input variables for the CNN models, correlations are calculated between the next (i.e. output variables) and the previous (i.e. input variables) 24 hours power consumptions, solar radiance and weather temperatures (°C). The correlations are calculated as a mean over two months of data (January and February): a “sliding” window moves along two months of data (January and February) at each 24 hours, and the correlation produce values at each successive 24 hours intervals. The mean of these successive 24 hours intervals are shown for each 39 buildings in Fig.11. There are implemented also correlation studies for different time intervals such as 6 hours, 12 hours, 24 hours and 48 hours. However, the most meaningful and clear correlations are obtained for the 24 hours time intervals.

In Fig.11, on the x-axis is located the building number, while on the y-axis are the previous 24 hours power consumptions, the previous 24 hours solar radiance and the previous 24 hours weather temperatures (i.e. year 2013, city of Manchester, UK).

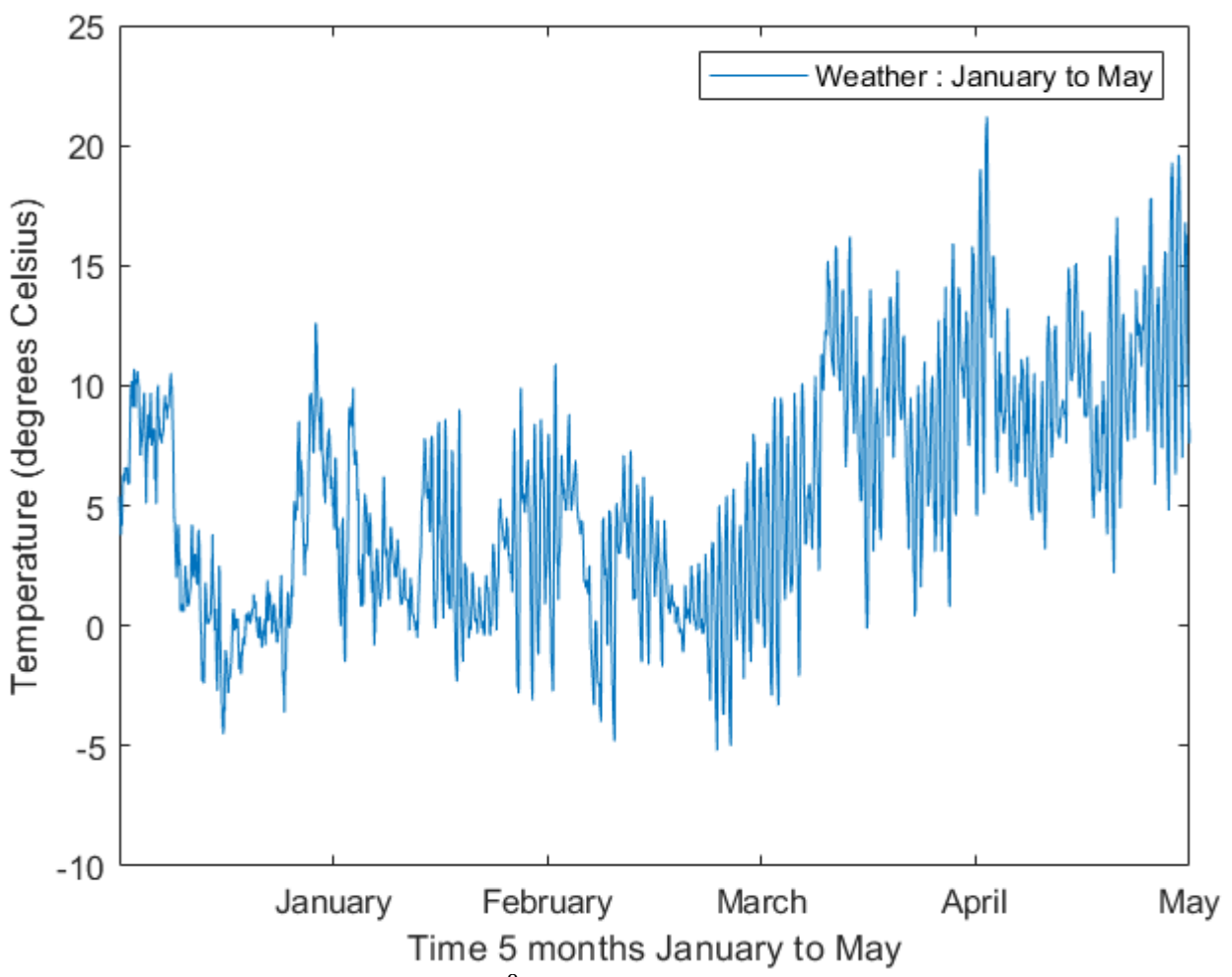

Fig.10. Temperatures (°C) for 5 months January to May 2013: temperatures increased gradually during April and May.

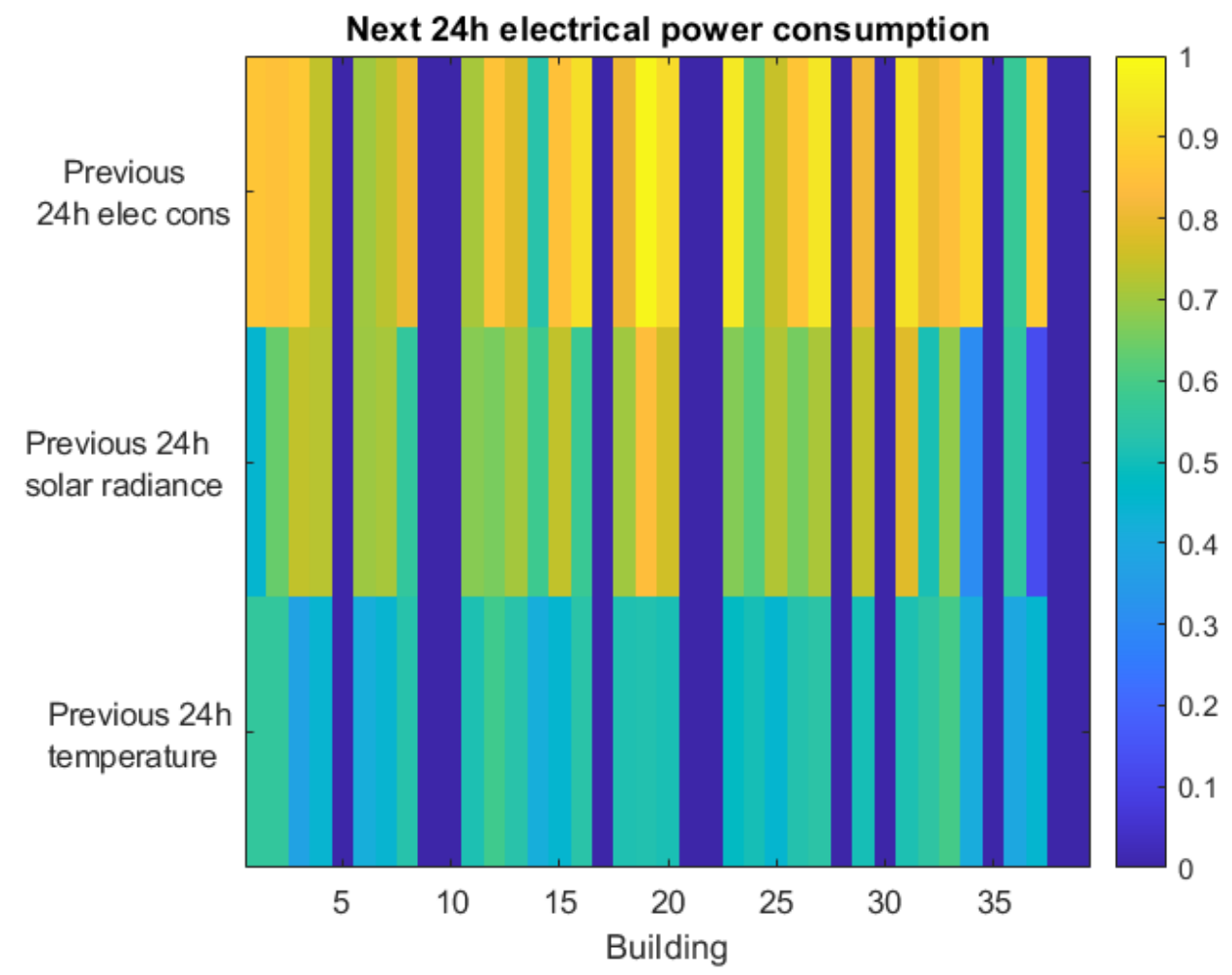

a) electric

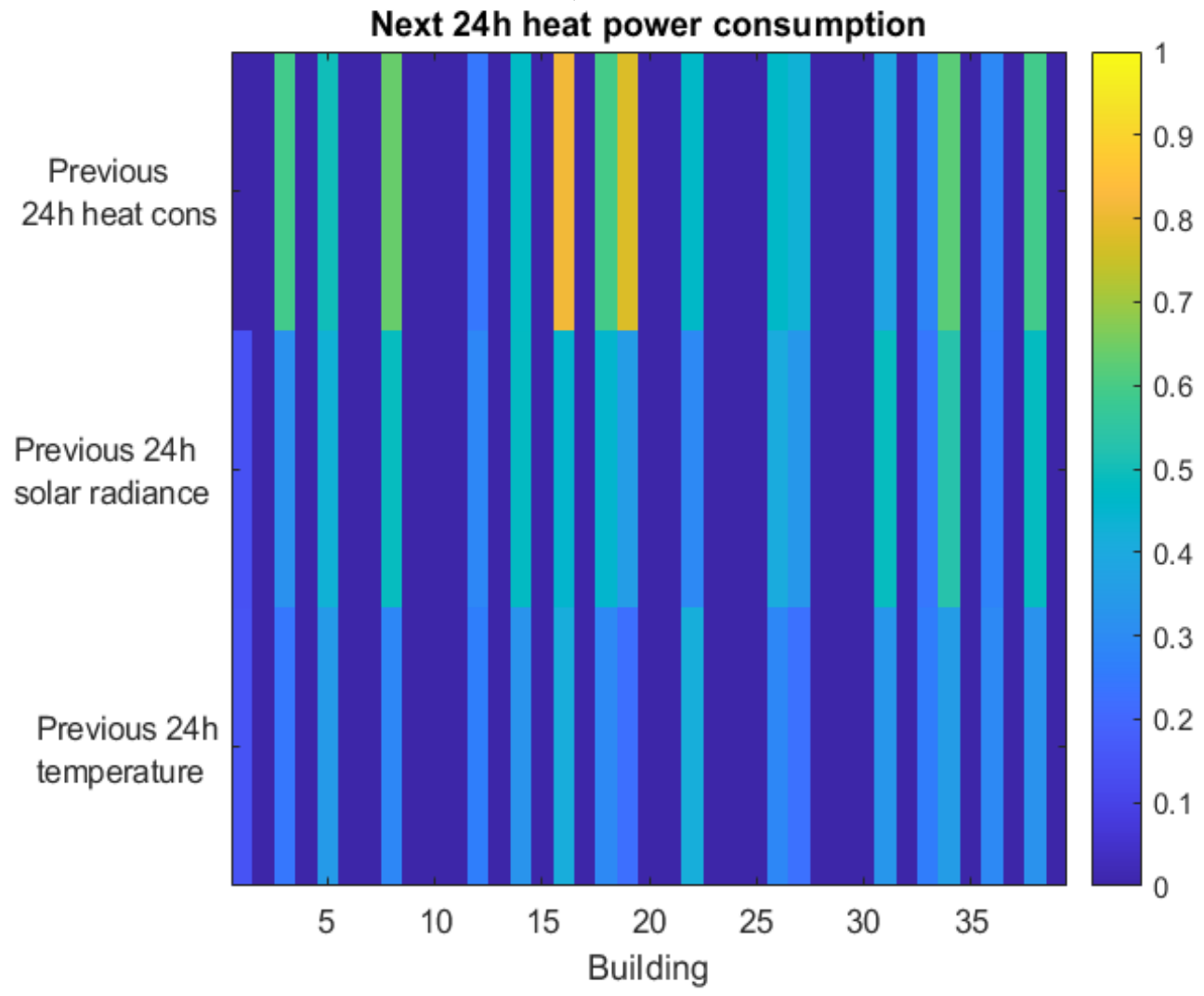

b) heat

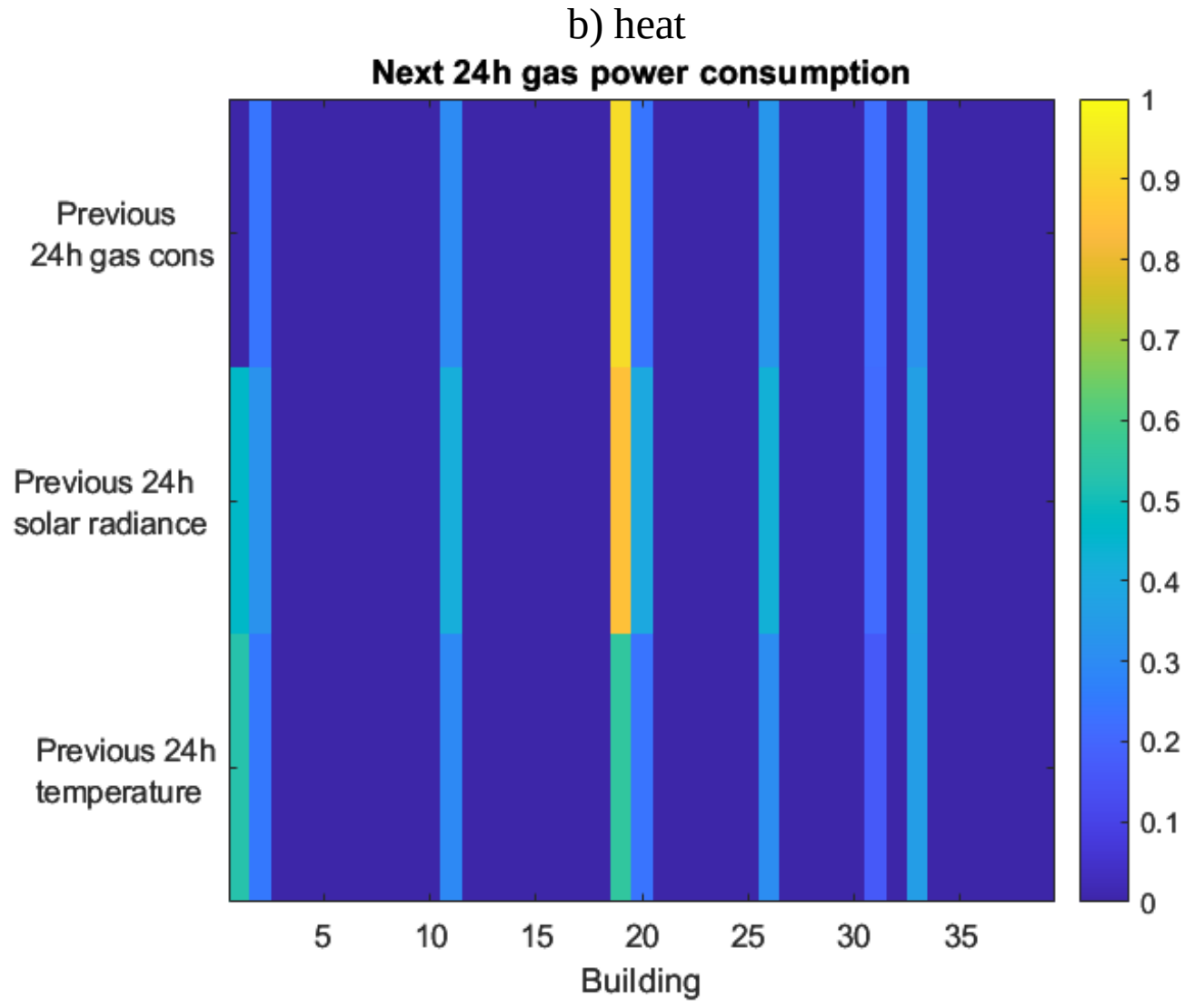

c) gas

Fig.11. Correlations between the next and the previous 24 hours power consumptions, solar radiance and temperature (°C) data and calculated as an average figure over two months of data using a “sliding” time window along two months of data and shown as mean values for each building: a) electric; b) heat; c) gas.

For the buildings with zero power consumptions (electric, heat or gas), there are shown dark blue vertical lines (Fig.11). When there is high correlation between the next and the previous 24 hours electrical consumptions such as for building 20 on the x-axis, then there is a yellow vertical line for the respective building on the 20th column shown in Fig.11(a). There are higher correlations for the next and the previous 24 hours electrical power consumptions compared to the weather data such as the previous 24 hours solar radiance or the previous 24 hours temperatures (°C). In Fig.11, as the correlations become weaker, then the colour moves gradually from yellow to dark blue. Table III shows that there is high correlation between the next and the previous energy consumptions of the same type (e.g. electric-electric), and rather than the previous energy consumptions of different types (e.g. electric-heat or electric-gas).

TABLE III.
CORRELATION FACTORS BETWEEN NEXT AND PREVIOUS 24 HOURS POWER CONSUMPTIONS FOR AN INTEGRATED ELECTRIC, HEAT AND GAS NETWORK SYSTEMS: BUILDING 3.

| | **Next 24 hours electric power cons.** | **Next 24 hours heat power cons.** | **Next 24 gas power cons.** |
|---|---|---|---|
| Previous 24 hours electric power consumption | **0.8709** | 0.3322 | 0.5792 |
| Previous 24 hours heat power consumption | 0.3484 | **0.5024** | 0.2604 |
| Previous 24 hours gas power consumption | 0.5923 | 0.2509 | **0.4703** |

Based on the Fig.11 and Table III, the input variables for the CNN models (e.g. CNN_1) are the previous 24 hours power consumptions (electric, heat or gas), while the outputs variables of the CNN models (e.g. CNN_1) are the predicted energy power consumption of the same type as the input variables. For example, if the electric power consumption is predicted then the time series consisting of the previous 24 hours electric power consumption are used as single input variable for the CNN models. The CNN predictions (i.e. output variable) are single point values of the predicted power consumptions (electric, heat or gas) for the next half an hour.

For federated learning, it would be of interest to determine the correlations between the non-zero power consumptions of the different buildings as shown in Fig.12. For the electric power consumptions, there are high correlations between almost all buildings as it can be seen with the yellow colour and the values above 0.5 in Fig.12: only building 23 makes an exception and it does not show any correlation because of its low electric power consumptions.

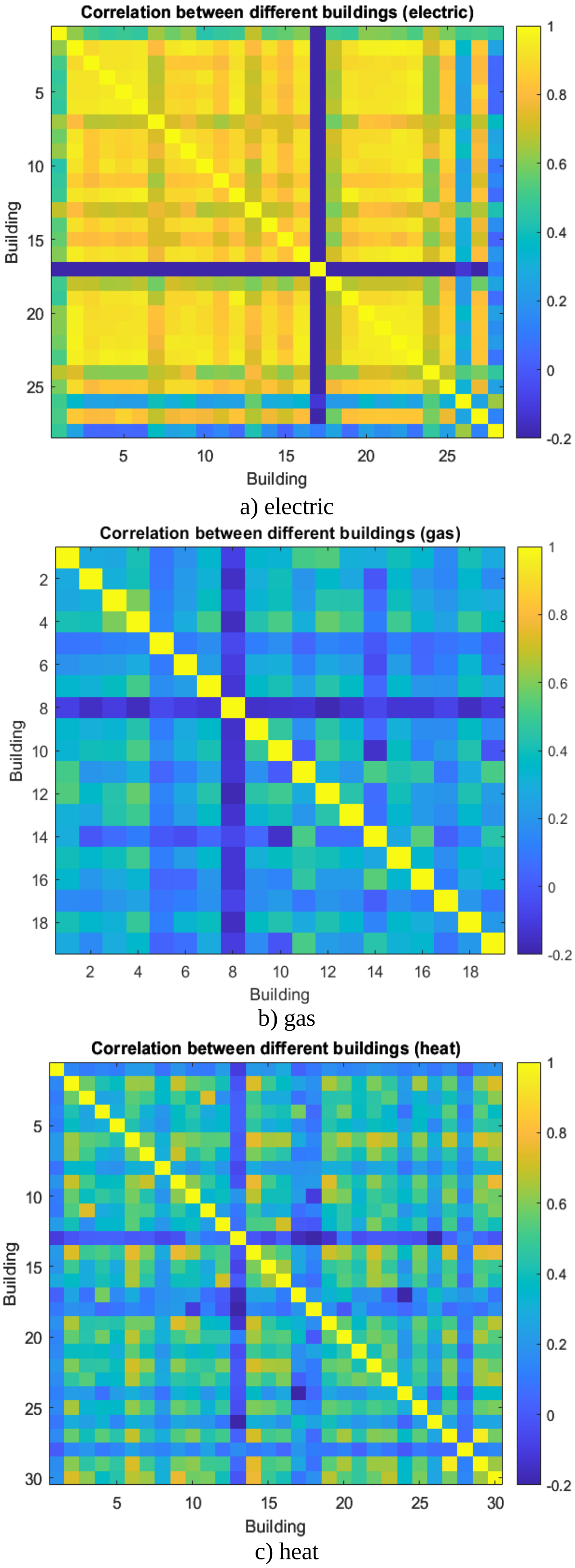

Fig.12. Correlation coefficients between the power consumptions of the different buildings with non-zero power consumptions: a) electric; b) gas; c) heat.

For the gas power consumptions, there are almost no correlations between the different buildings as it can be seen that the blue colour is everywhere in Fig.12(b). Obviously that there is auto-correlation (i.e. correlation for the same building) as shown on the main diagonal of Fig.12(b). Finally, for heat power consumptions (Fig.12(c)) there is some correlation between very few buildings and most of buildings exhibit average correlation (i.e. around 0.4 shown with green colour).

### *B. Short-term CNNs predictions*

The short-term CNN predictions (i.e. next half an hour) will be shown and discussed in this section *B*, which predictions will be obtained with the 6 CNN frameworks described in the methods section III.

#### *B1. Single input variable and single output variable results (CNN_1) and based on the determination of the number of optimization epochs obtained by simulation*

CNN_1 model framework is developed with a single input variable consisting of the previous 24 hours power consumptions (either electric, heat or gas) and a single output variable (i.e. electric input-electric output, heat input-heat output or gas input-gas output). A maximum number of 400 epochs is used to train the CNN model(s), which took on average around 7 minutes. The maximum number of epochs was determined by simulation and by plotting the SNR and NRMSE for several buildings function of the number of iterations and for the testing datasets: the simulation results are shown in section AA.1 of the supplementary material.

The testing predictions of the total power consumptions (electric, heat and gas) are shown in Figs.13 together with the SNR and the NRMSE values. The time on the x-axis is shown in counts of half hours instead of days or months and therefore for the testing predictions of month of March, the x-axis it stretches up to 1488.

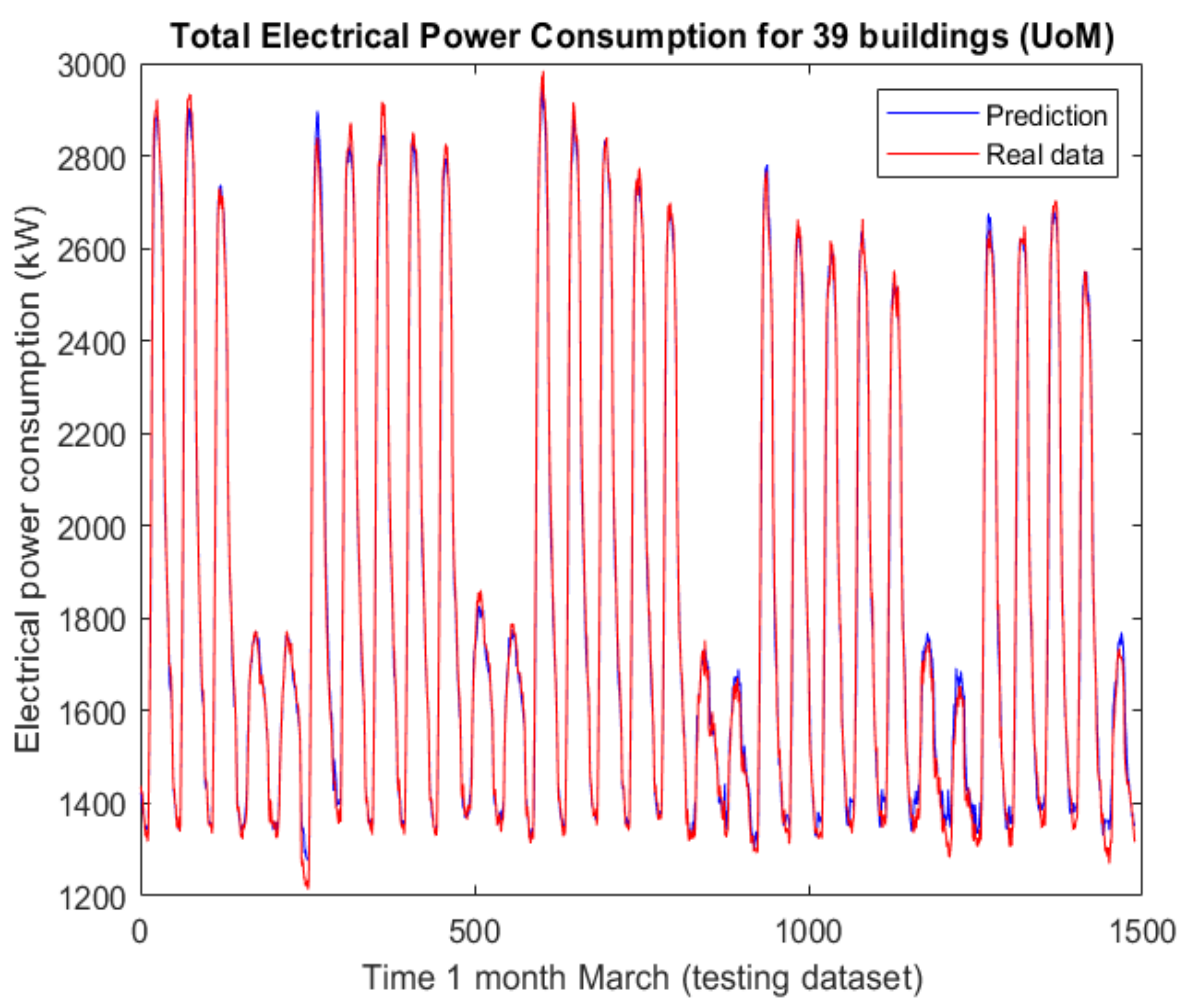

a) electric

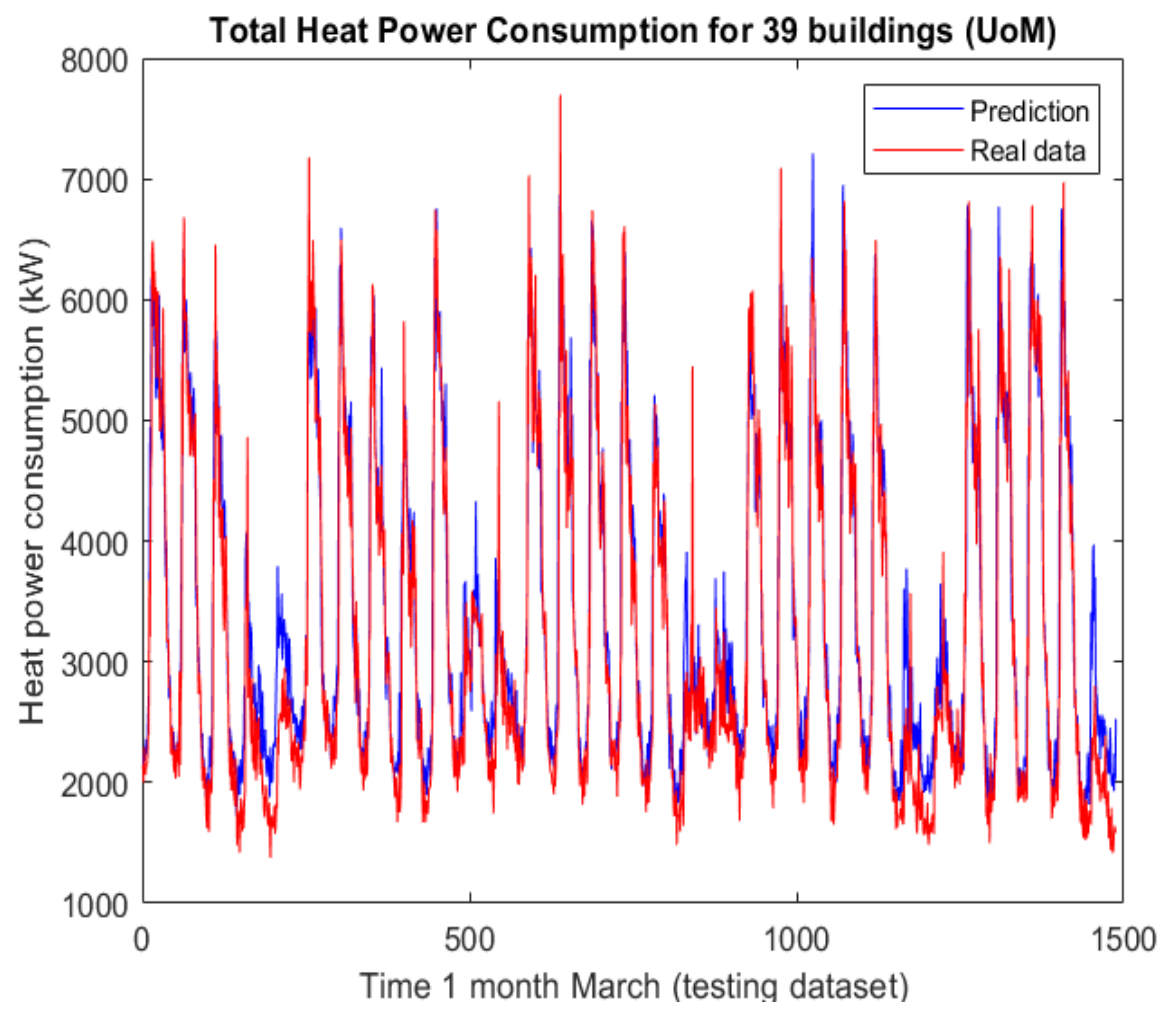

b) heat

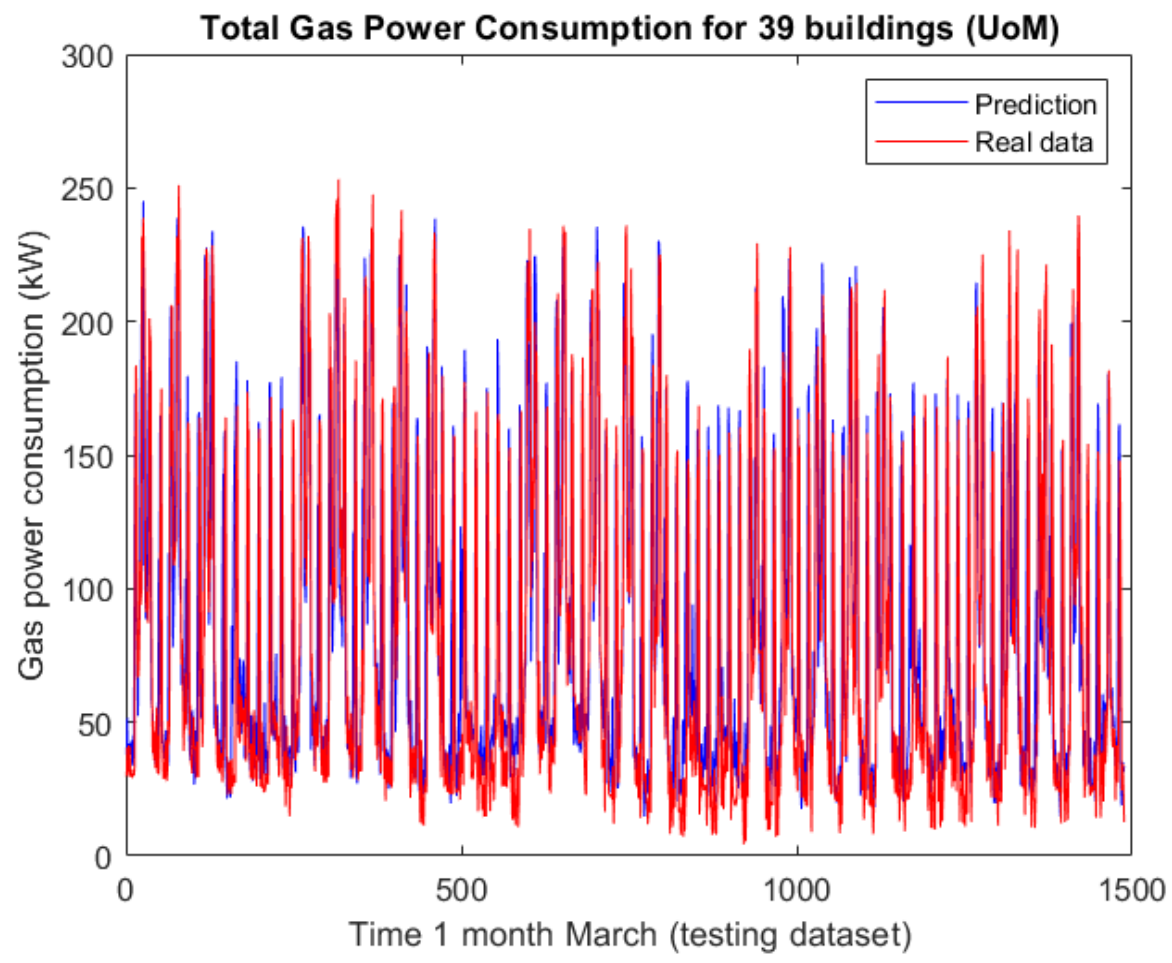

c) gas

Fig.13. Prediction of total power consumptions (CNN_1: single input variable/single output variable) for 39 buildings and the testing dataset: a) electric network (SNR = 34.63dB, NRMSE = 0.0085); b) heat network (SNR =17.22 dB, NRMSE =0.0628); c) gas network (SNR = 8.68dB, NRMSE =0.1477).

The electric network predictions seem to be always very good for the testing datasets. The training results although are not shown, they are also similarly good for the electric network (SNR = 38.84dB, NRMSE = 0.007), for the heat network (SNR= 26.77 dB , NRMSE =0.0195) and for the gas network (SNR =22.59 dB, NRMSE = 0.0293).

For each of the 28 buildings with non-zero electric power consumptions, the accuracy of testing predictions (i.e. single input variable) are shown in Fig.14 (CNN_1). SNR_average is SNRs values averaged over the 28 buildings. For each of the 30 buildings with non-zero heat power consumptions, the SNR and NRMSE for the testing predictions are shown in Fig.15.

There is a decrease in the accuracy of the heat power predictions because they are more non-linear than the electric power consumptions (Fig.4). For the gas testing dataset and for 8 buildings with gas power consumptions higher than 2000 kW per month, the SNR and NRMSE values are shown in Fig.16.

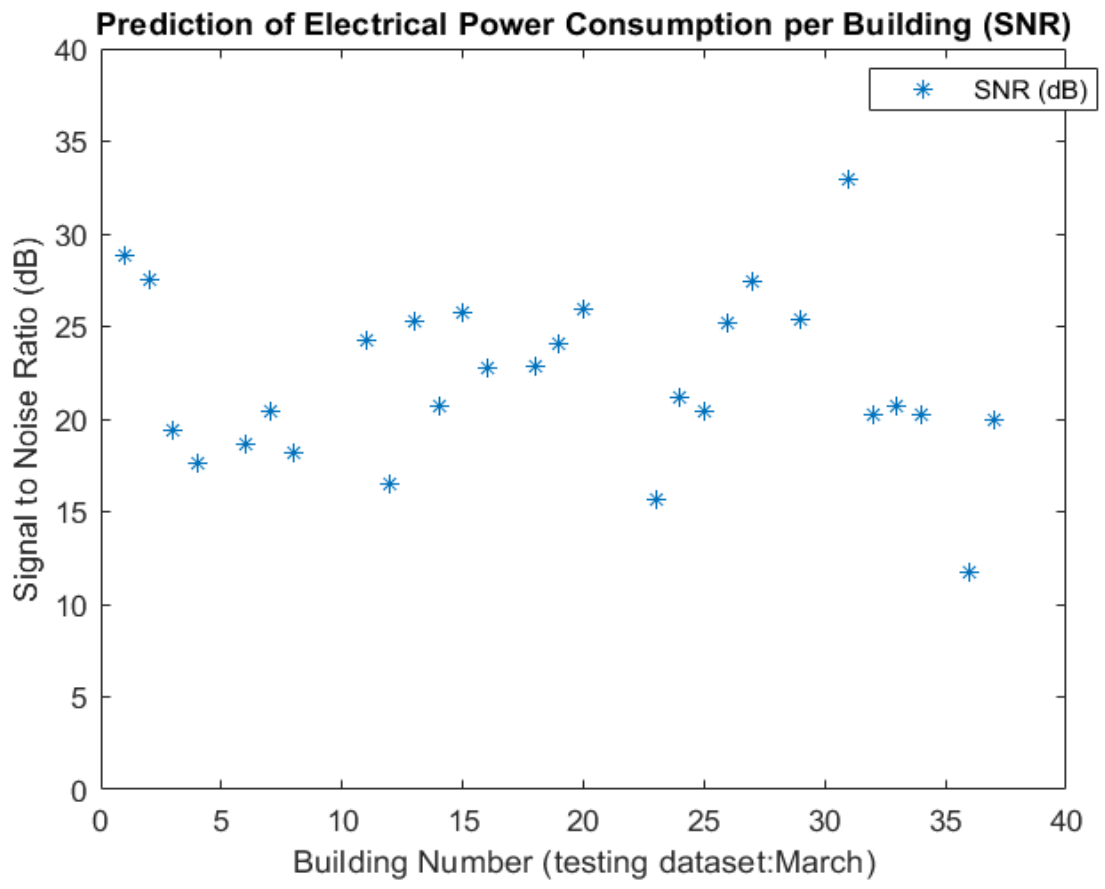

a) SNR

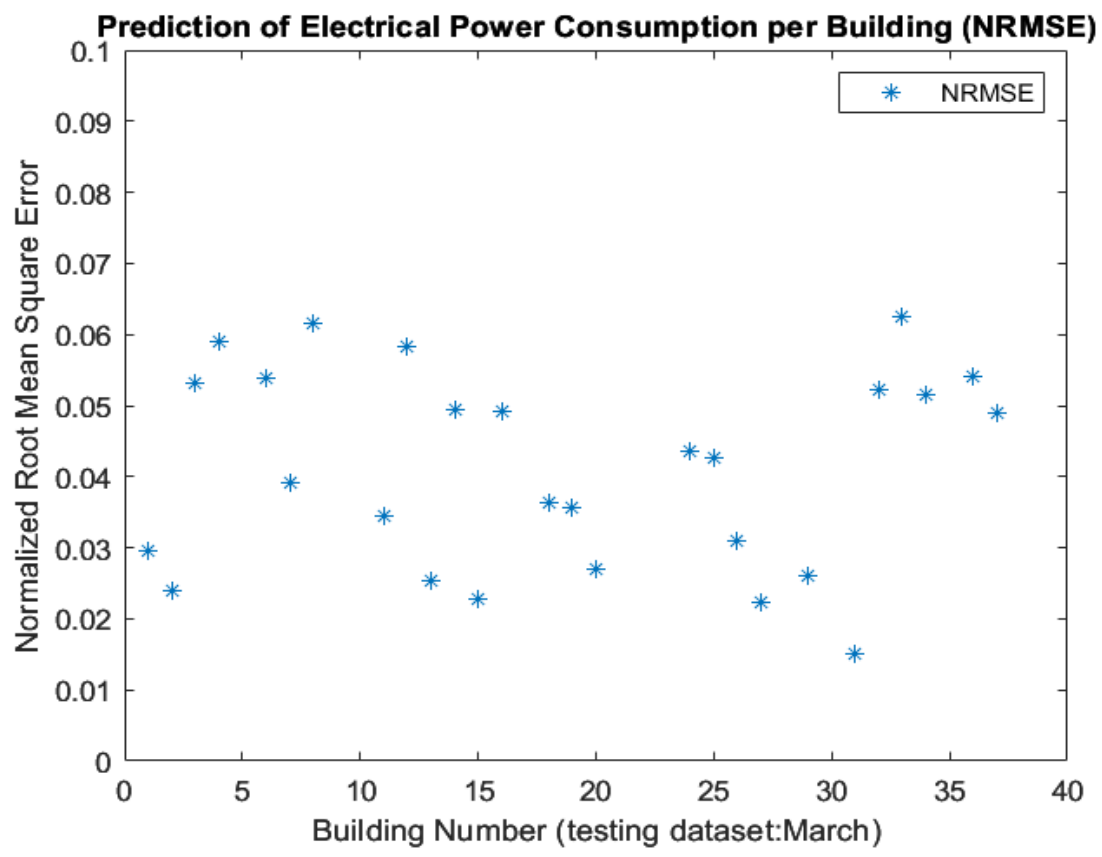

b) NRMSE

Fig.14. SNR and NRMSE for testing datasets for 28 buildings with non-zero electric power consumptions - (CNN_1): a) SNR_average = 22.12 dB; b) NRMSE_average =0.043.

*B2. Multiple input variables and single output variable results (CNN_2) and based on the determination of the number of optimization epochs obtained by simulation*

CNN_2 model is used for buildings, which have a network connection between the heat network and the electric network or between the heat network and the electric network and the gas network, or between the gas network and the electric network.

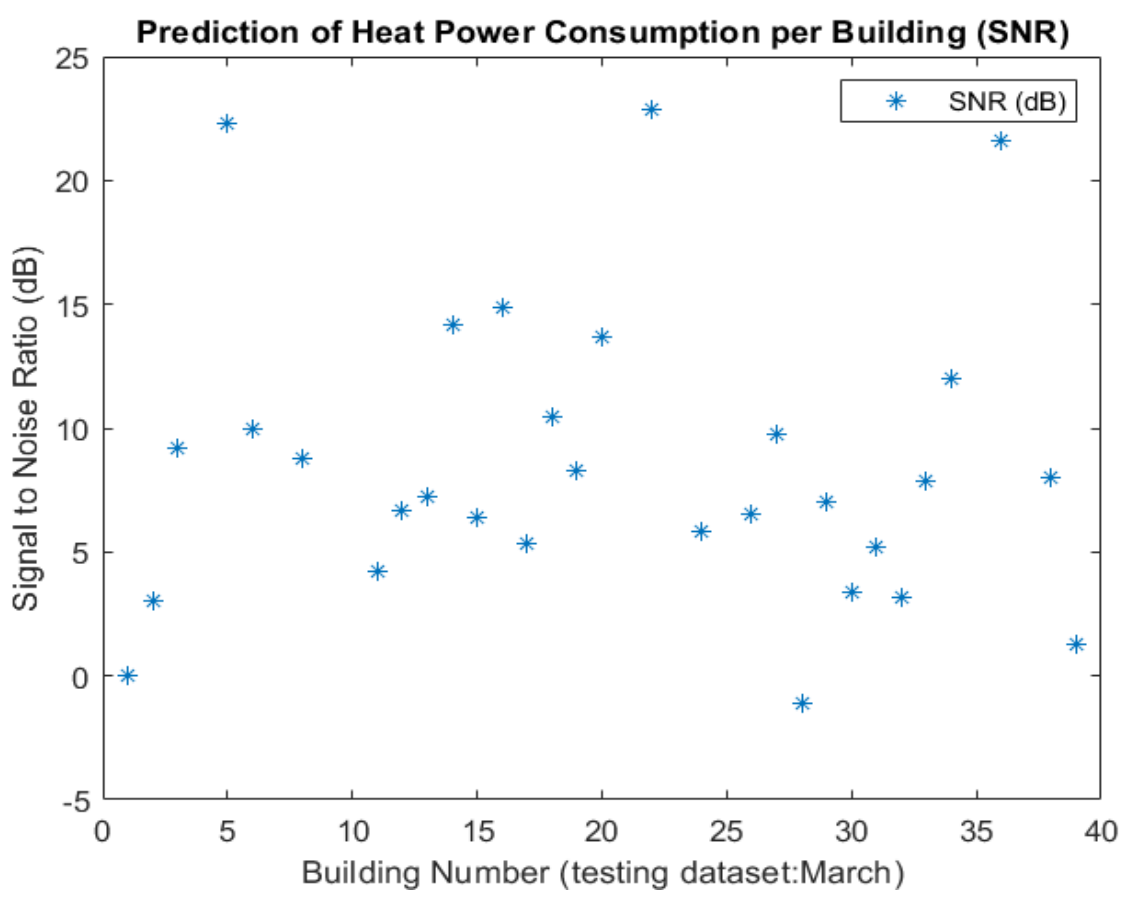

a) SNR

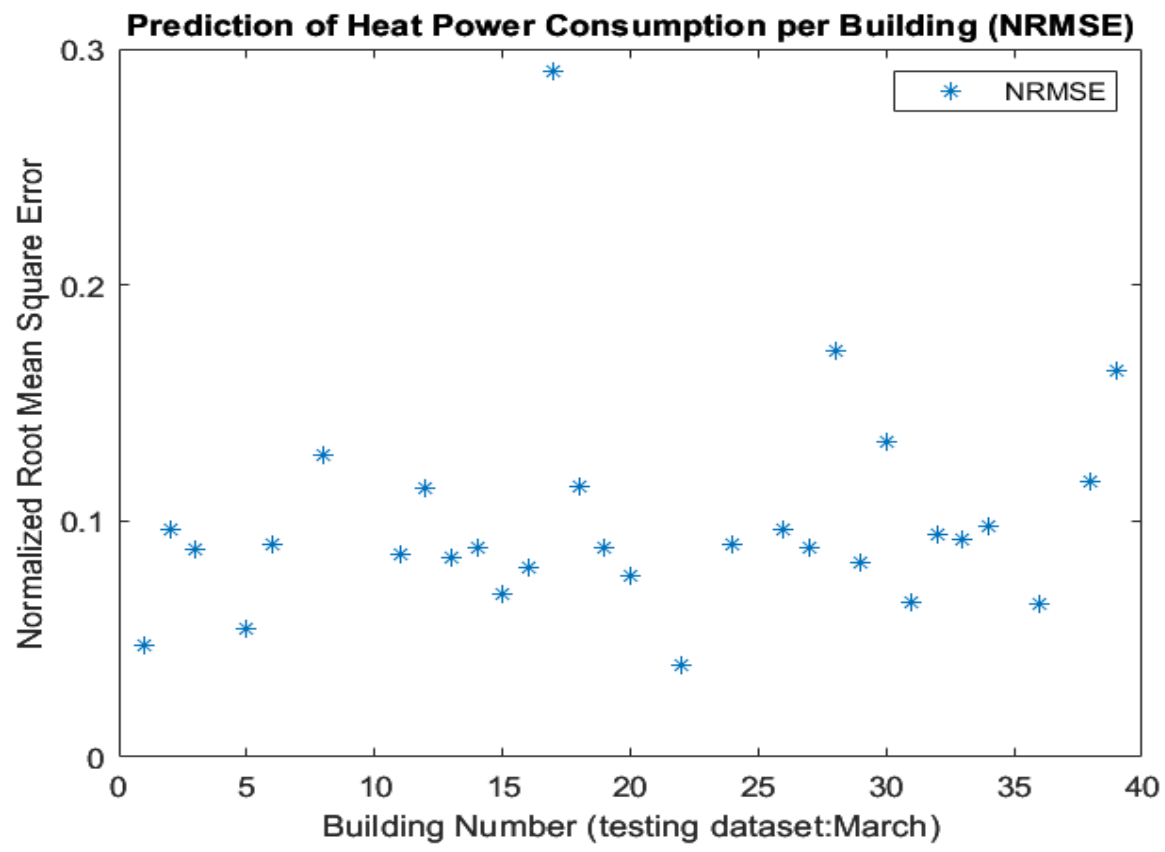

b) NRMSE

Fig.15. SNR and NRMSE for testing datasets for 30 buildings with non-zero heat power consumptions (CNN_1): a) SNR_Average = 8.59dB; b) NRMSE_Average =0.0997.

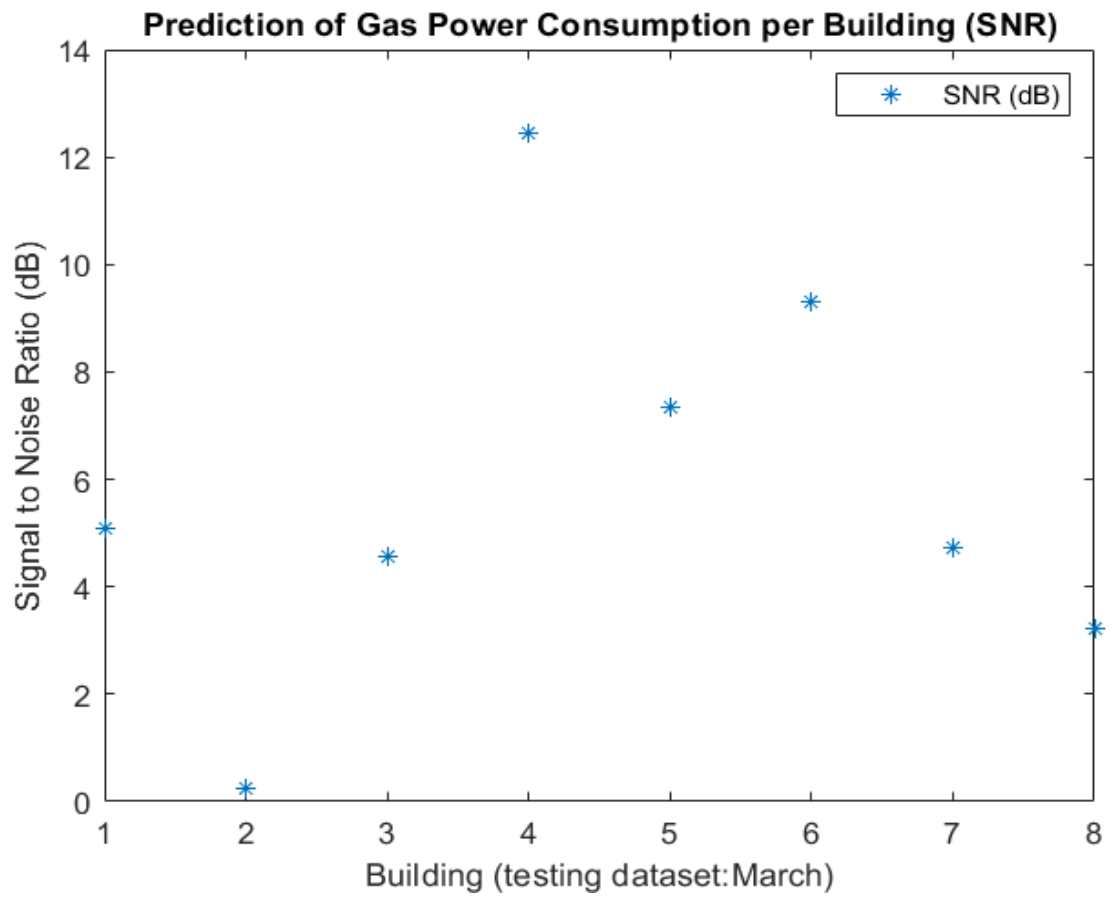

a) SNR

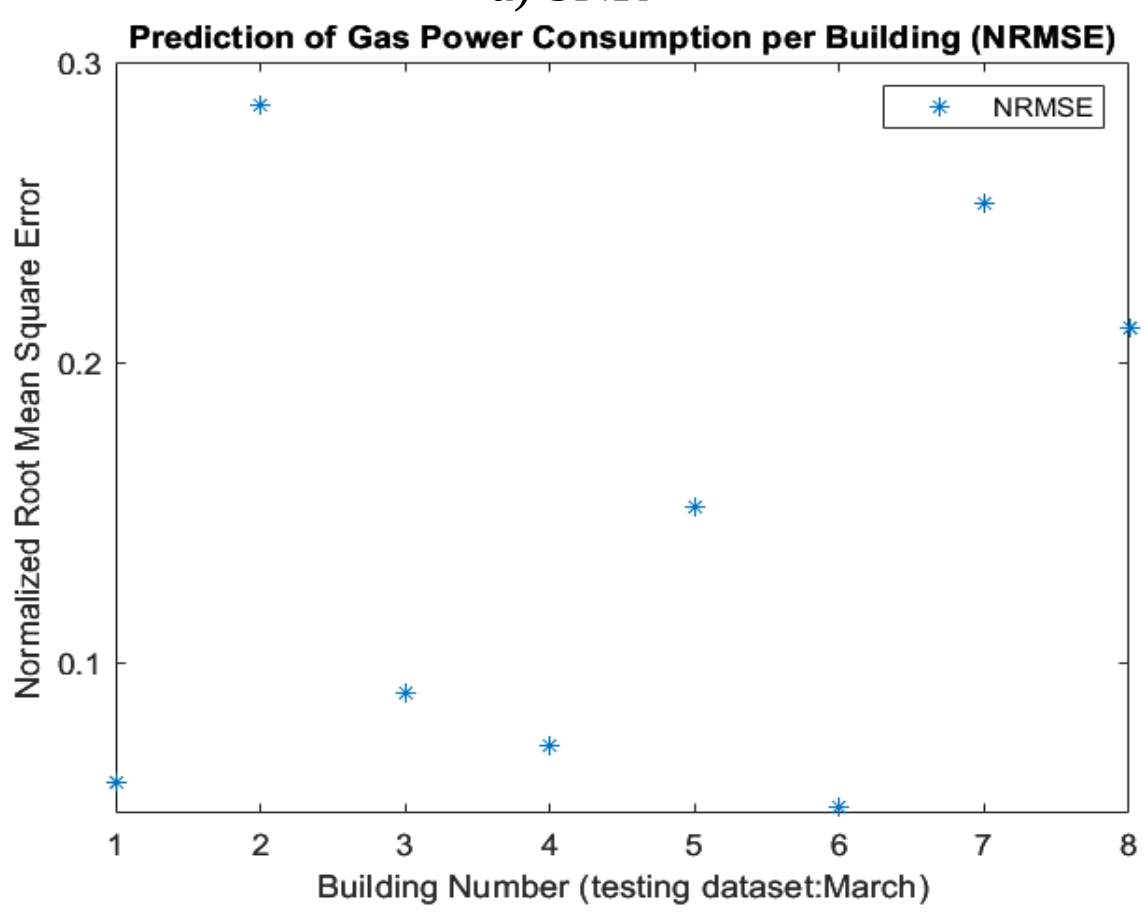

b) NRMSE

Fig.16. SNR and NRMSE for testing datasets for 8 buildings with higher gas power consumptions (CNN_1): a) SNR_Average = 6.01 dB; b) NRMSE_Average =0.14.

The testing results are shown in Fig.17, and in the training stage, a maximum number of 400 epochs is used. The single output variable of CNN_2 model is the next half an hour heat or gas power consumption.

Overall, the SNR and NRMSE differences between when using a single input variable (Fig.13(b), Fig.13(c)) and multiple input variables (Fig. 17) for the heat and the gas testing predictions are small but in favour of using a single input variable of the same type of power consumption as the output variable (heat or gas). The differences are in favour of using a single input variable based also on the high correlations shown for example in Table III on the main diagonal (i.e. 0.8709, 0.5024, 0.4703). The gas testing predictions (i.e. Fig.17(b)) can be regarded very challengeable as they are non-linear and scarce.

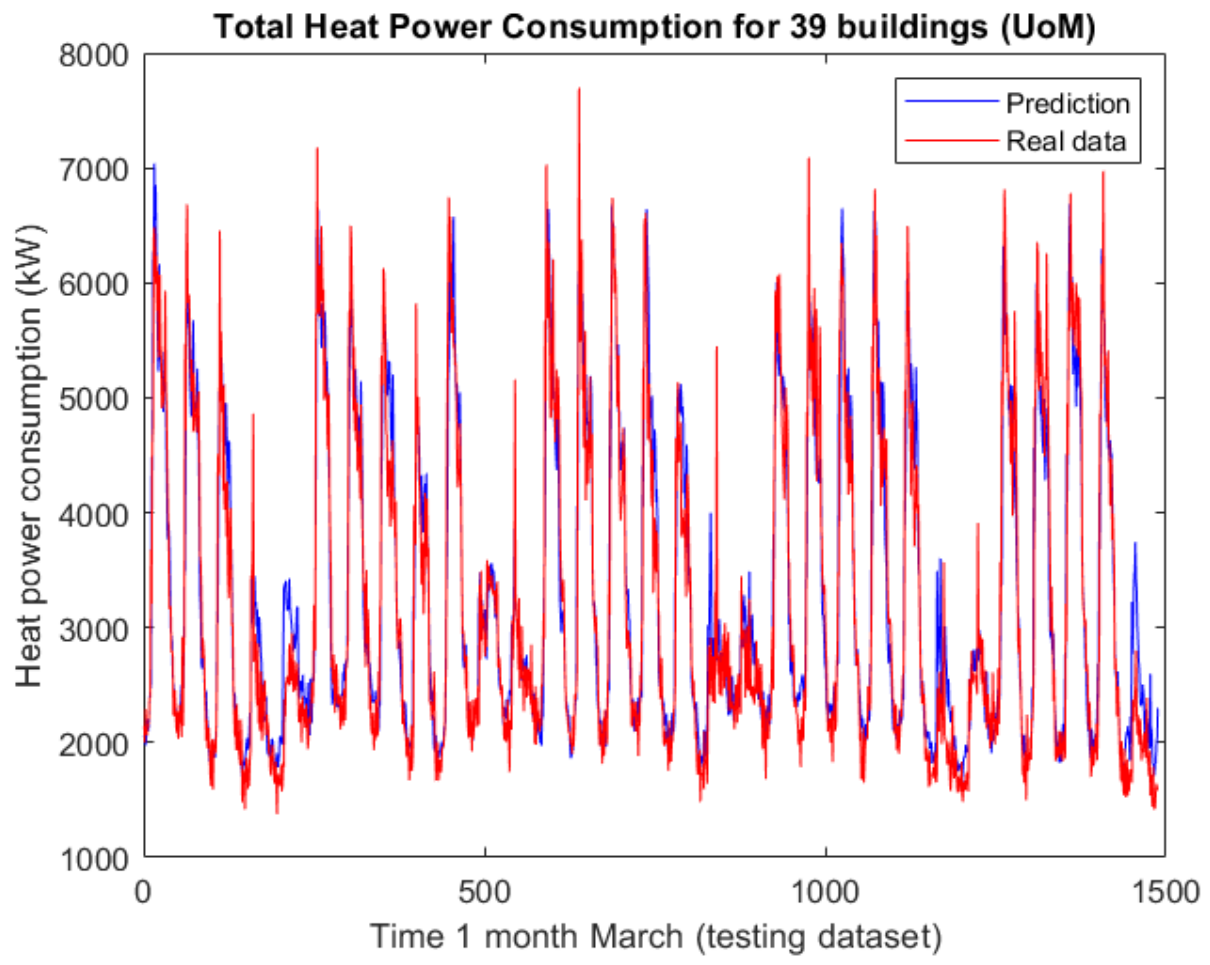

a) heat

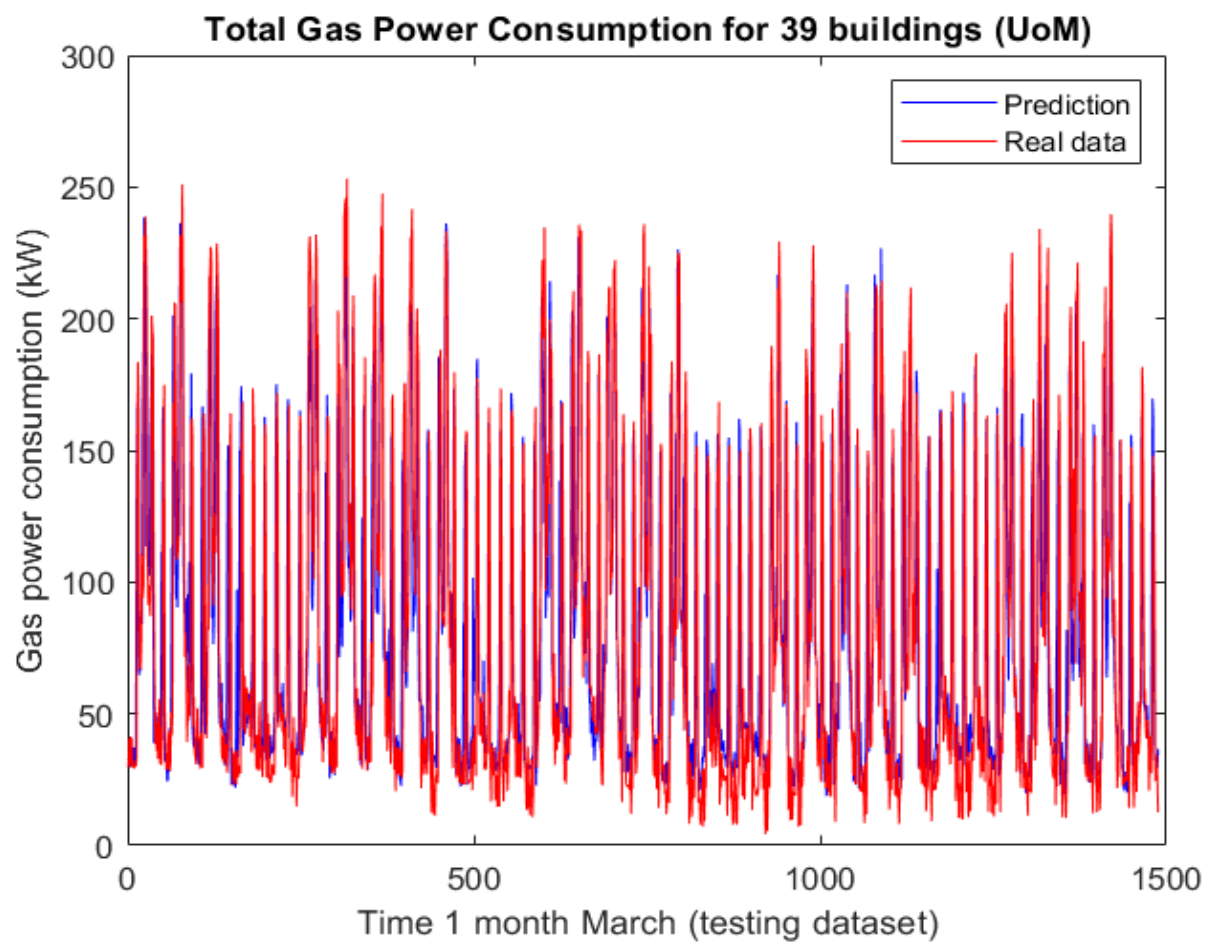

b) gas

Fig.17. Prediction of total power consumptions (CNN_2: multiple input variables/single output variable) for the testing dataset: a) heat network (SNR 17.42dB, NRMSE = 0.061); b) gas network (SNR = 8.86 dB, NRMSE = 0.14).

In general, it is not easy to compare the prediction results between different integrated network systems as they have their specific characteristics (i.e. network topologies, etc). However, a comparative study between the present results and the ones shown in [57] in terms of MAPE, produces the following metrics: 1.52 versus 3.43 [57] for the electric load testing predictions and 9.27 versus 5.16 [57] for the heat test predictions.

### *B.3 Short-term CNNs predictions using training, validation and testing datasets (CNN_3)*

Validation dataset(s) is used to update the hyperparameters of the DL model, as described in Fig.6. There are used similar settings as before, with 400 epochs used during the training/validation and the Adam optimizer with a batch training data size of 700. CNN_3 uses a single input variable and a single output variable of the same energy type as the input variable (i.e. electric-electric, heat-heat, gas-gas).

The results for the training datasets are comparable with the results shown in the previous section IVB.1 when using a single input variable: electric network (SNR = 38.18 dB, NRMSE =0.0077), heat network (SNR = 19.68 dB, NRMSE =0.0443) and gas network (SNR =15.264dB, NRMSE =0.0682).

Figs. 18, 19 and 20 show the predictions for the testing datasets for the electric, heat and gas power consumptions, when using validation datasets for training of the CNN_3 models. The predictions show similar quality with the testing results from section IVB.1.

In the section AA.6 of the supplementary material there are also shown for several buildings the training loss and the validation loss function of the number of epochs for several buildings and energy vectors.

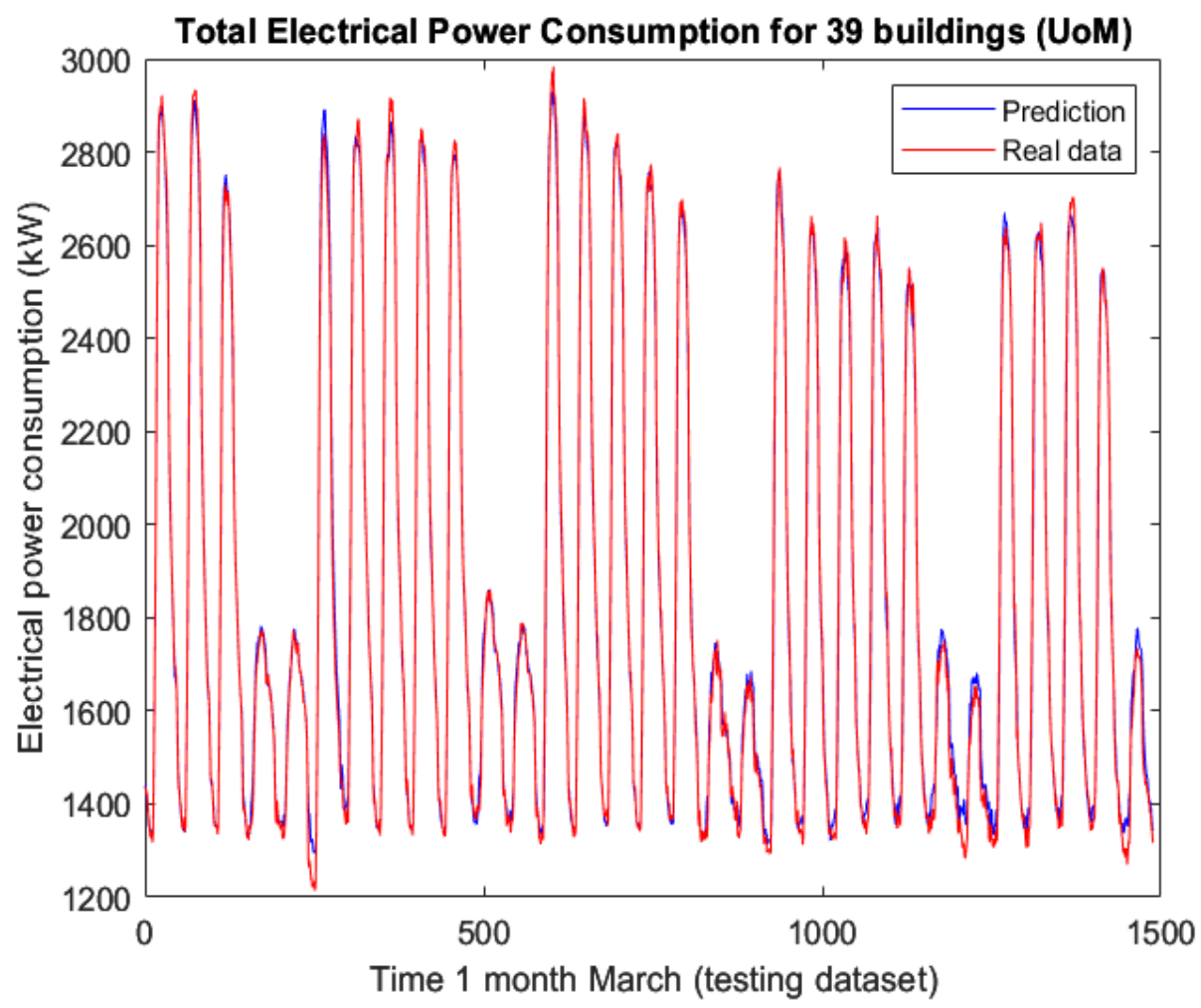

Fig. 18. Prediction of total electric power consumption for 39 buildings of UoM and using training/validation/testing datasets (CNN_3). (testing results: SNR= 34.87 dB, NRMSE=0.0116)

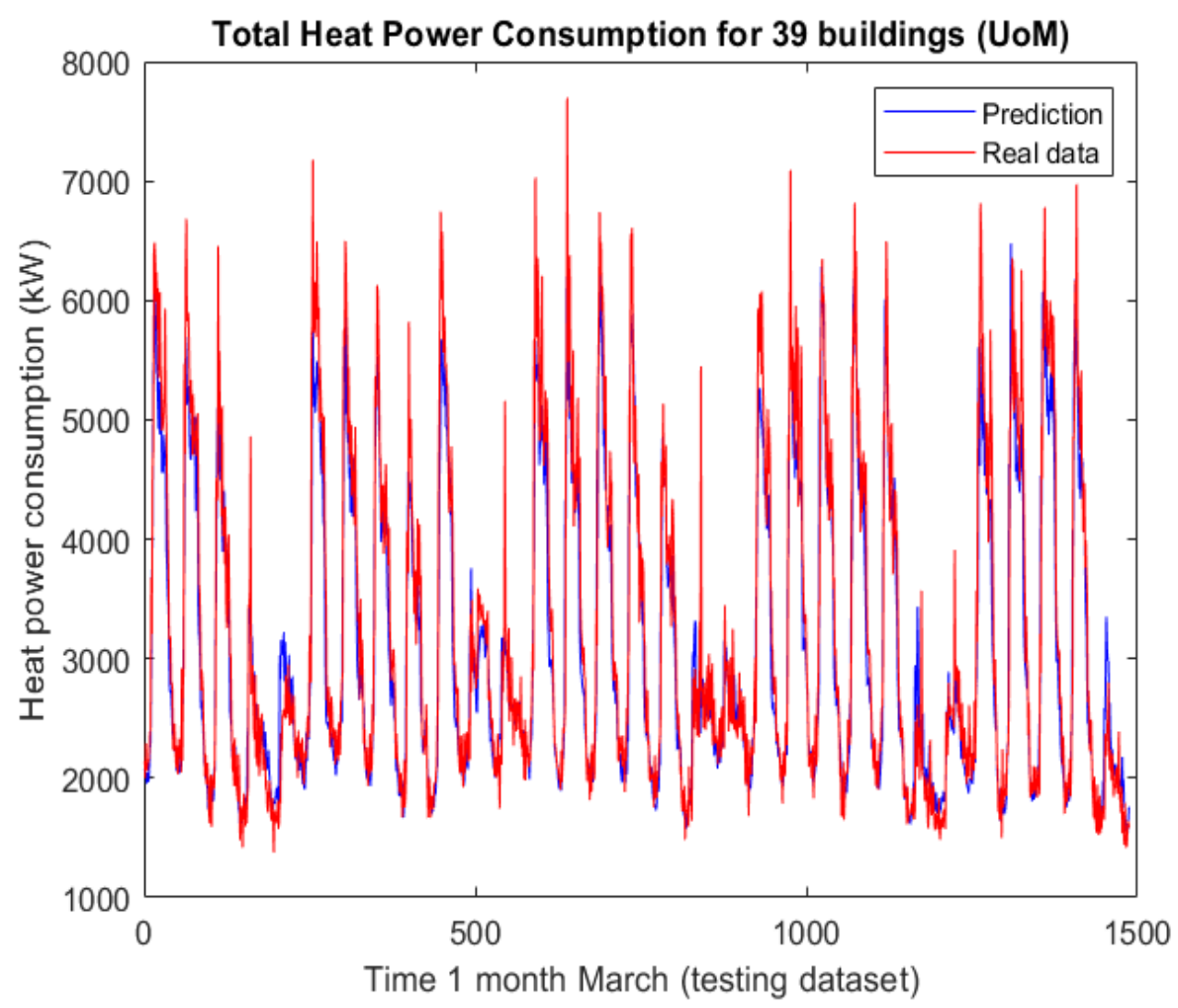

Fig. 19. Prediction of total heat power consumption for 39 buildings of UoM and using training/validation/testing datasets (CNN_3). (testing results: SNR= 17.84 dB, NRMSE=0.0584)

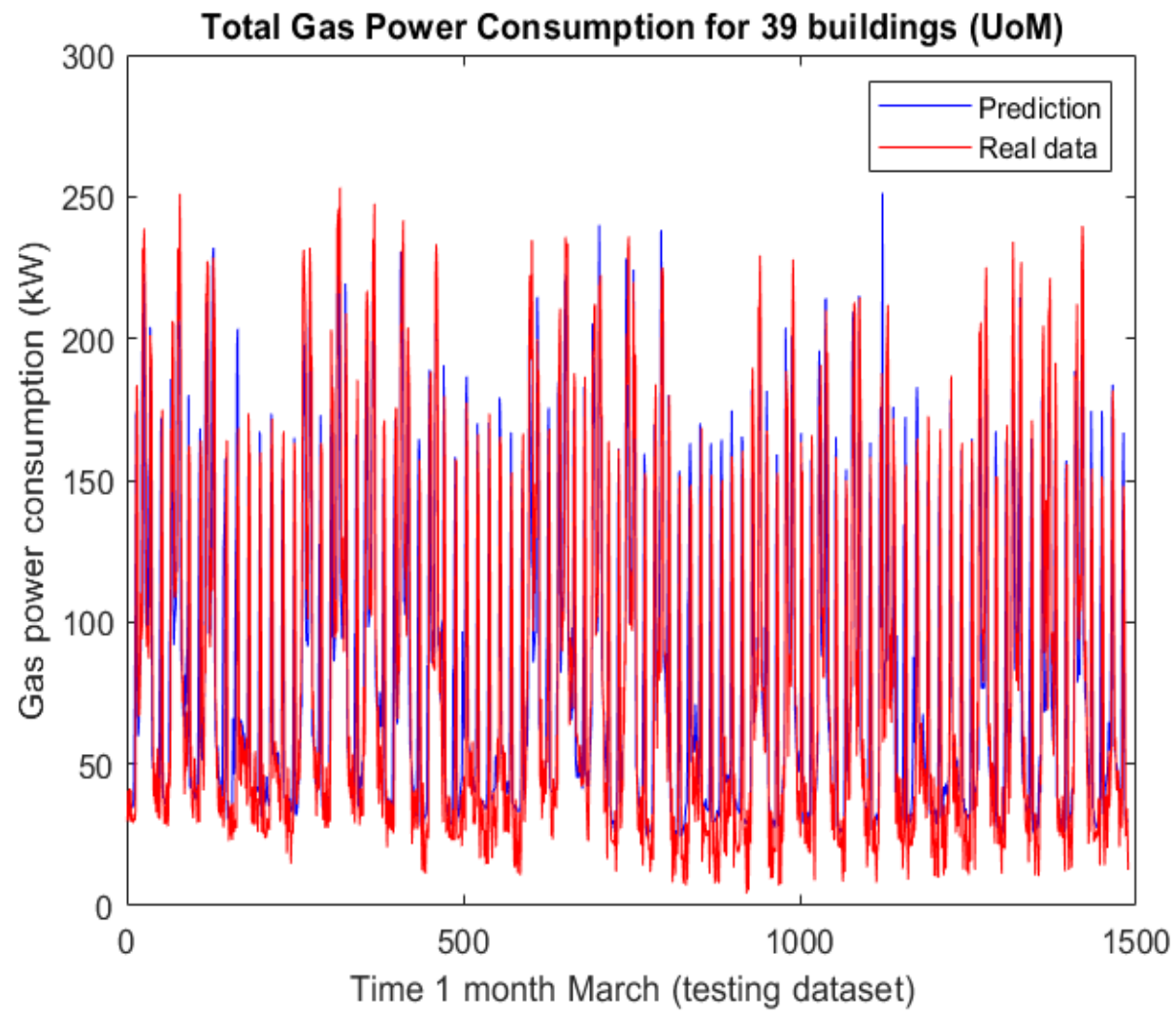

Fig. 20. Prediction of total gas power consumption for 39 buildings of UoM and using training/validation/testing datasets (CNN_3). (testing results: SNR= 9.18dB, NRMSE=0.151)

### *B.4 Short-term CNNs predictions based on a single CNN model for the entire integrated energy system (CNN_4)*

In this section, a single CNN model is used with multiple inputs and multiple outputs. CNN_4 model includes in the training data all the electric, heat and gas data and predicts simultaneously the electric, the heat and the gas power consumptions for the month of March. This is also called joint multi-energy prediction. The training settings are 2000 epochs during the training and the Adam optimizer with a batch training data size of 700. The results for the training datasets are good for the electric network (SNR = 31.43dB, NRMSE = 0.017), for the heat network (SNR = 22.76dB, NRMSE = 0.031) and for the gas network (SNR = 13.43dB, NRMSE = 0.017). Fig. 21 shows the predictions of the total electric, heat and gas power consumptions for the 39 buildings and for the testing dataset covering month of March (2013). The SNR and the NRMSE values are good for the electric and the heat predictions, but are not as good as the predictions obtained in section IVB.1 and which are shown underlined in Fig. 21.

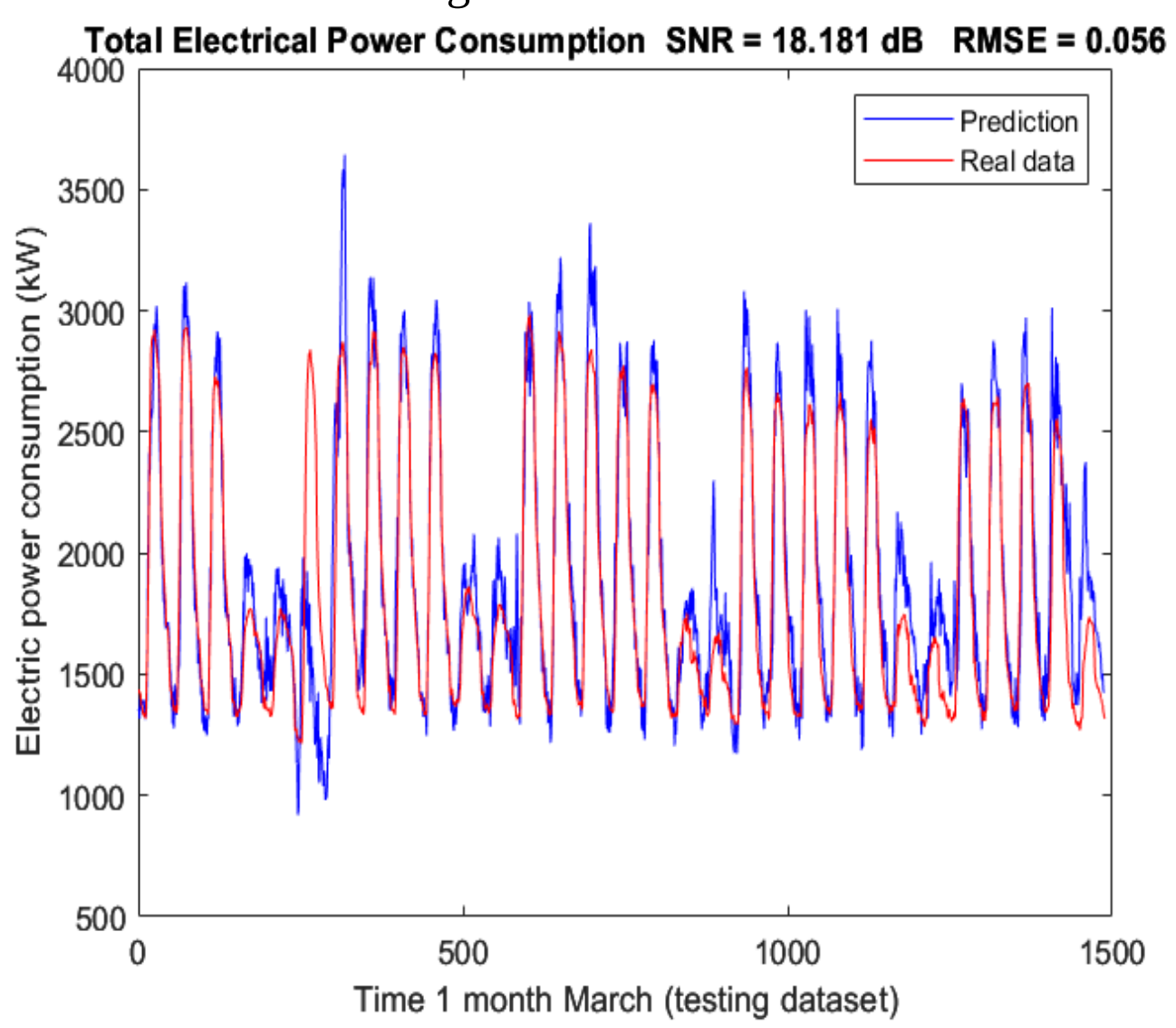

a) electric testing dataset: SNR = 18.18dB /34.63 dB, NRMSE = 0.056 / 0.0085.

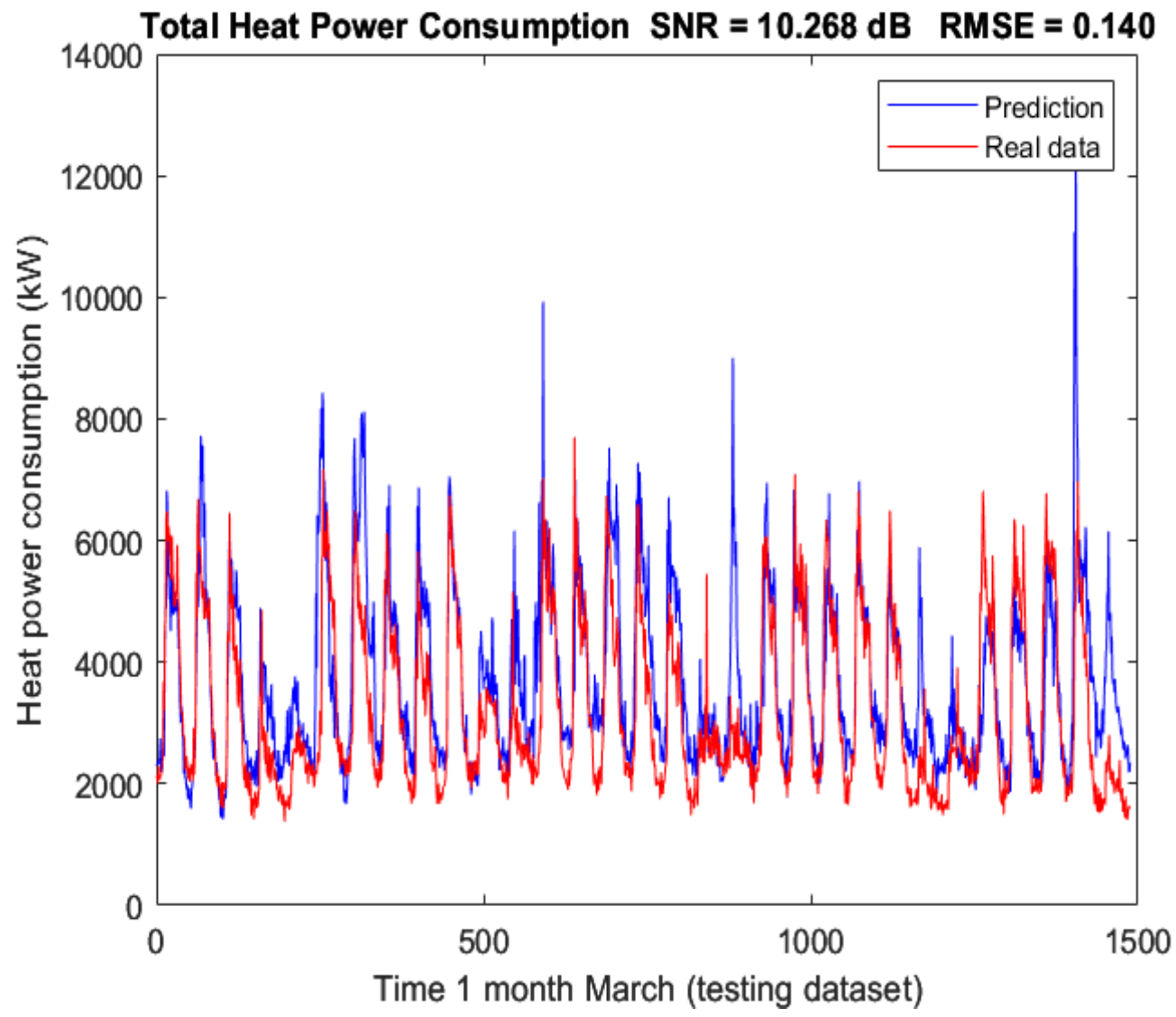

b) heat testing dataset: SNR =10.26dB /17.22dB, NRMSE = 0.14 / 0.0628.

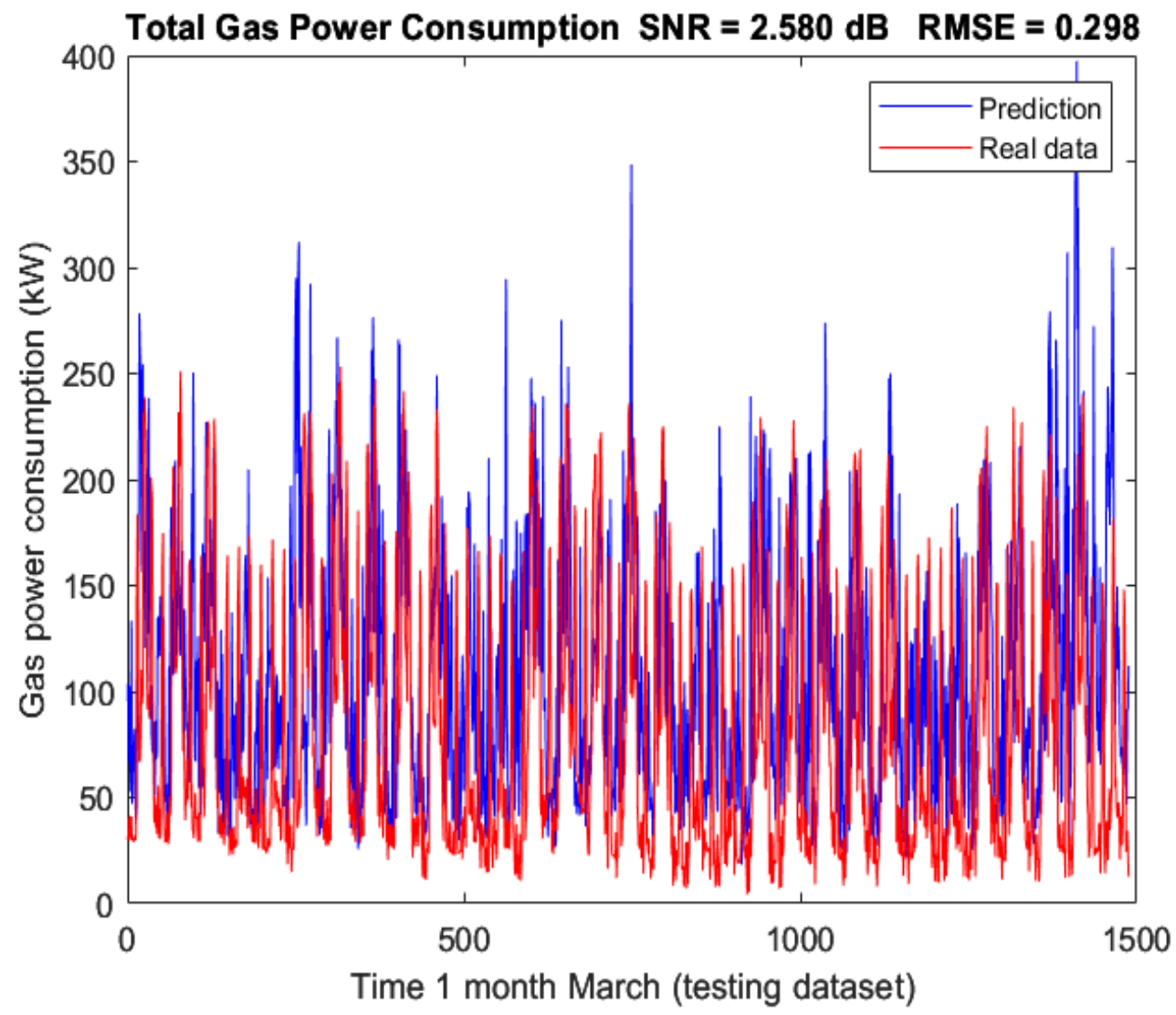

c) gas testing: SNR = 2.58dB/8.86dB, NRMSE=0.298/0.14.

Fig. 21. Joint prediction of total electric, heat and gas power consumptions for 39 buildings for testing dataset of month of March (2013) and a single CNN_4 model.

### *B.5 Short-term CNNs predictions based on a single CNN model for each energy network (multiple-building input/ output variable) (CNN_5)*

This section presents the prediction of each energy vector (electric, heat, gas) by using a single CNN model for each energy network. CNN_5 model has an image input layer of size [48 39 1] and a regression output layer corresponding to an energy network, and with the size of [1 39].

The training settings are 800 epochs during the training and the Adam optimizer with a batch training data size of 1000. The results for the training datasets are at an acceptable level, which are for the electric network (SNR = 29.75dB, NRMSE = 0.0229), the heat network (SNR =

12.71dB, NRMSE = 0.1029) and for the gas network (SNR = 9.92dB, NRMSE = 0.1261).

Fig.22 shows the testing results obtained with a single CNN_5 model, which was trained and tested simultaneously for all 39 buildings of the electric network. The underlined numbers in Fig.22 are for the results from Fig.13.

Fig.23 shows the testing results obtained with a single CNN_5 model trained and tested simultaneously for all 39 buildings for the heat network. Fig.24 shows the results for the gas network obtained with another single CNN_5 model. The heat and the gas results evaluated by SNR and NRMSE are not as good as the underlined numerical results shown in the same figures, and which are the results from section IVB.1 for when a single input variable/single output variable were used.

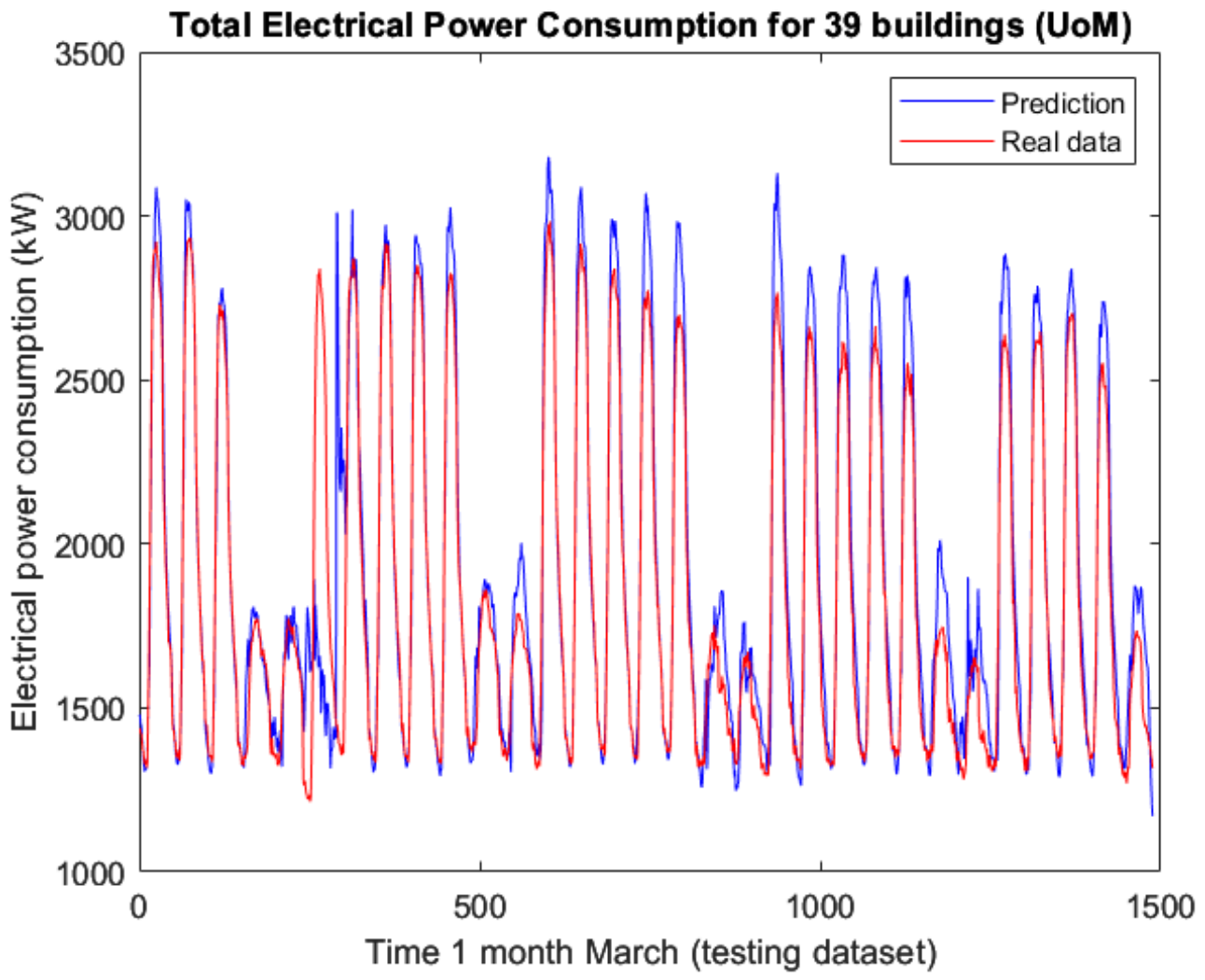

Fig.22. Prediction of total electric power consumptions for 39 buildings using training dataset January and February (2013) and testing dataset March (2013) (i.e. a single CNN_5 model for the electric energy vector): electric testing dataset SNR = 19.70dB / 26.77dB, NRMSE = 0.0664/ 0.0195.

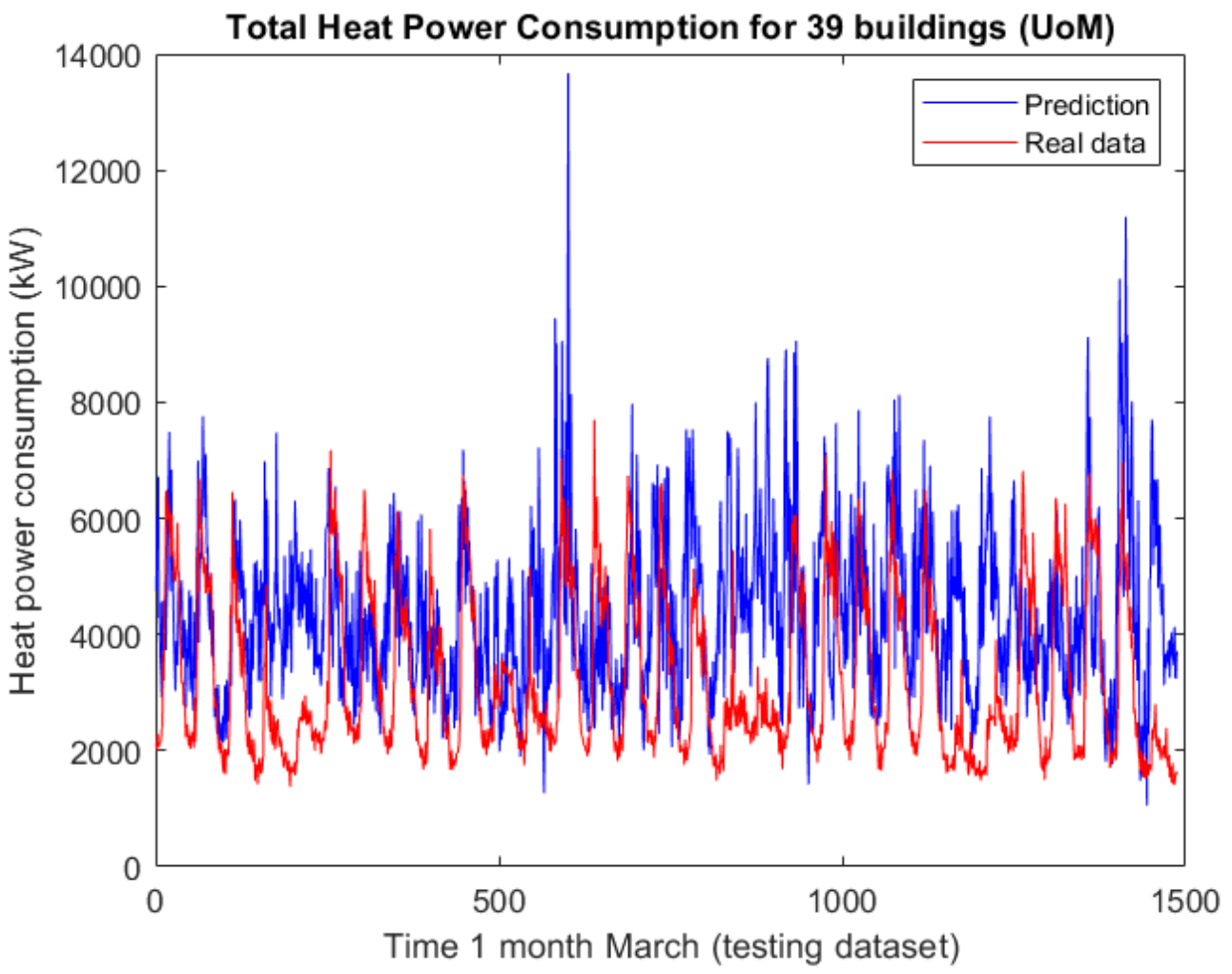

Fig. 23. Prediction of total heat power consumptions for 39 buildings using training (January and February) and testing (March) datasets (i.e. a single CNN_5 model for the heat energy vector): heat testing dataset SNR = 4.79dB 17.22dB, NRMSE = 0.2626/ 0.0628.

*B.6 Short-term CNNs predictions based on federated learning for integrated energy system (CNN_6)*

An average federated learning algorithm is developed. The CNN_1 architecture described previously, but with only a single convolutional layer, is used in the federated learning algorithm for the global and the local CNN models.

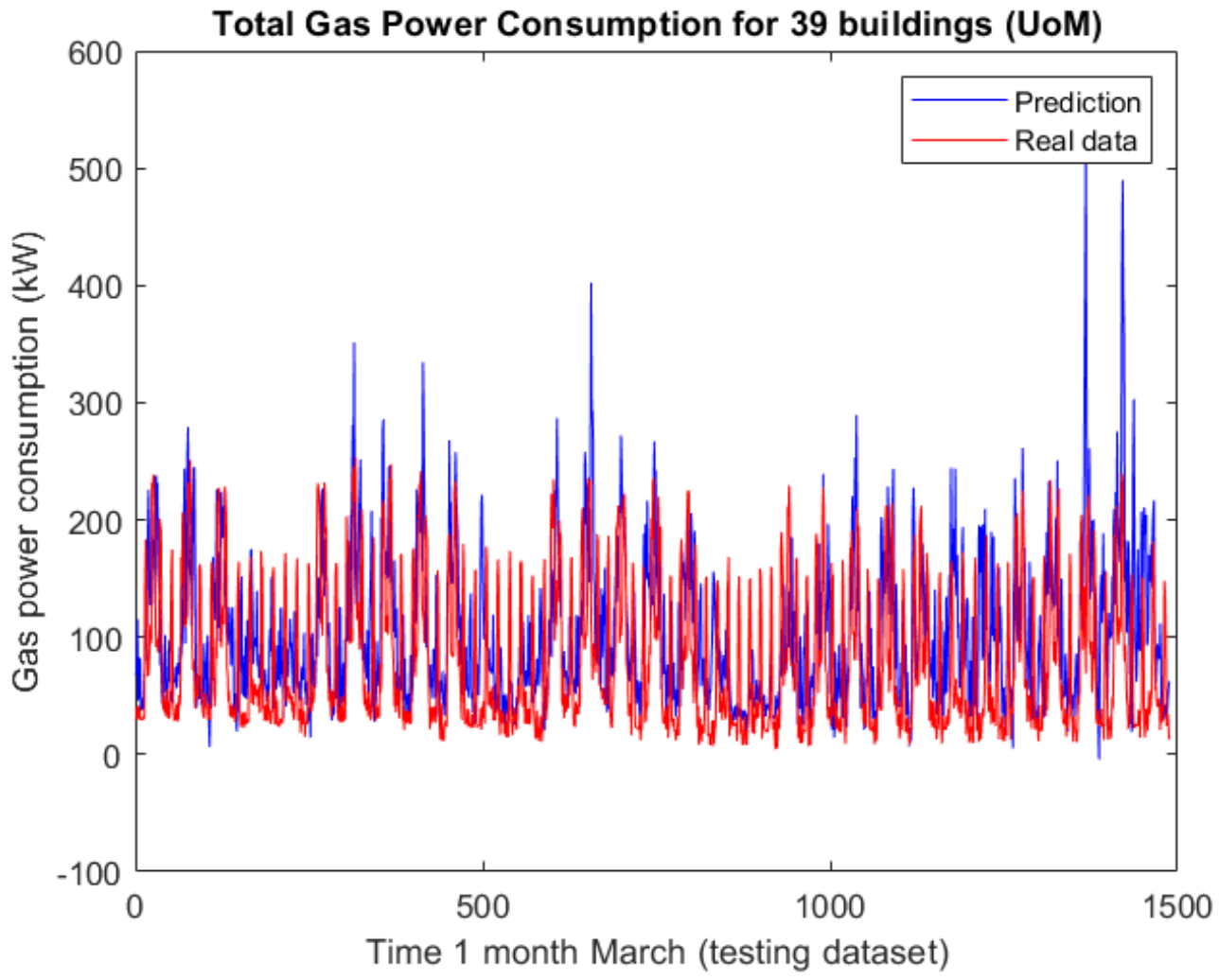

Fig. 24. Prediction of total gas power consumptions for 39 buildings using training dataset January and February (2013) and testing dataset March (2013) (i.e. a single CNN_5 model for the entire gas energy vector):gas testing dataset SNR = 3.18dB /8.68 dB, NRMSE =0.2781/ 0.1477.

The results for the training datasets are at an acceptable level, which are for the electric network (SNR = 29.75dB, NRMSE = 0.0229), the heat network (SNR = 12.71dB, NRMSE = 0.1029) and for the gas network (SNR = 9.92dB, NRMSE = 0.1261).

Fig.25 shows the testing predictions of the total electric power consumption obtained for 39 buildings (UoM) with the average federated learning algorithm (CNN_6).

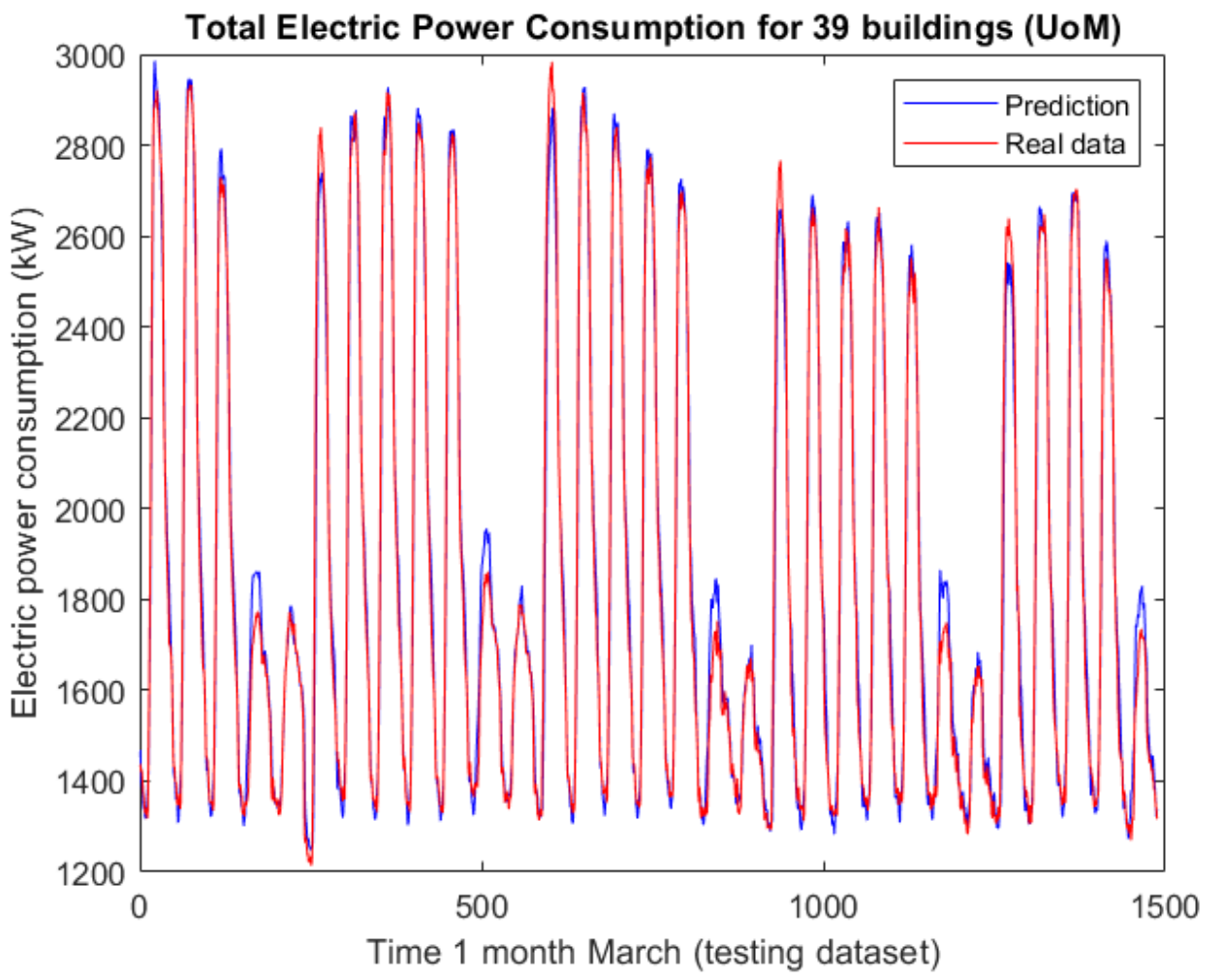

Fig.25. Predictions of the total electric power consumption obtained with federated learning algorithm (CNN_6): testing dataset NRSME=0.0210, SNR=29.72db.

The SNR and the NRMSE values are very good, which means that the average federated learning is a good predictive method for this electric network, and it should be also possible to use the resulted federated global CNN_6 model in real-life situations.

However, Fig.26 shows the predicted total heat power consumption for 39 interconnected buildings (UoM) obtained with the same average federated learning algorithm (CNN_6) for the testing dataset. In this case, it is clear that because the training heat power consumptions are more non-linear in comparison to the electric power consumptions, then it becomes unfeasible to use this type of learning.

Fig.27 shows that because the gas power consumptions are both non-linear and also scarce, then it becomes difficult to predict accurately both the individual buildings gas power consumptions and also the total gas power consumption.

The findings from Figs. 25, 26 and 27 are also confirmed by the correlation matrixes from Fig. 12, which show strong correlation between the buildings with non-zero electrical power consumptions, and weaker correlations for the heat and the gas networks buildings.

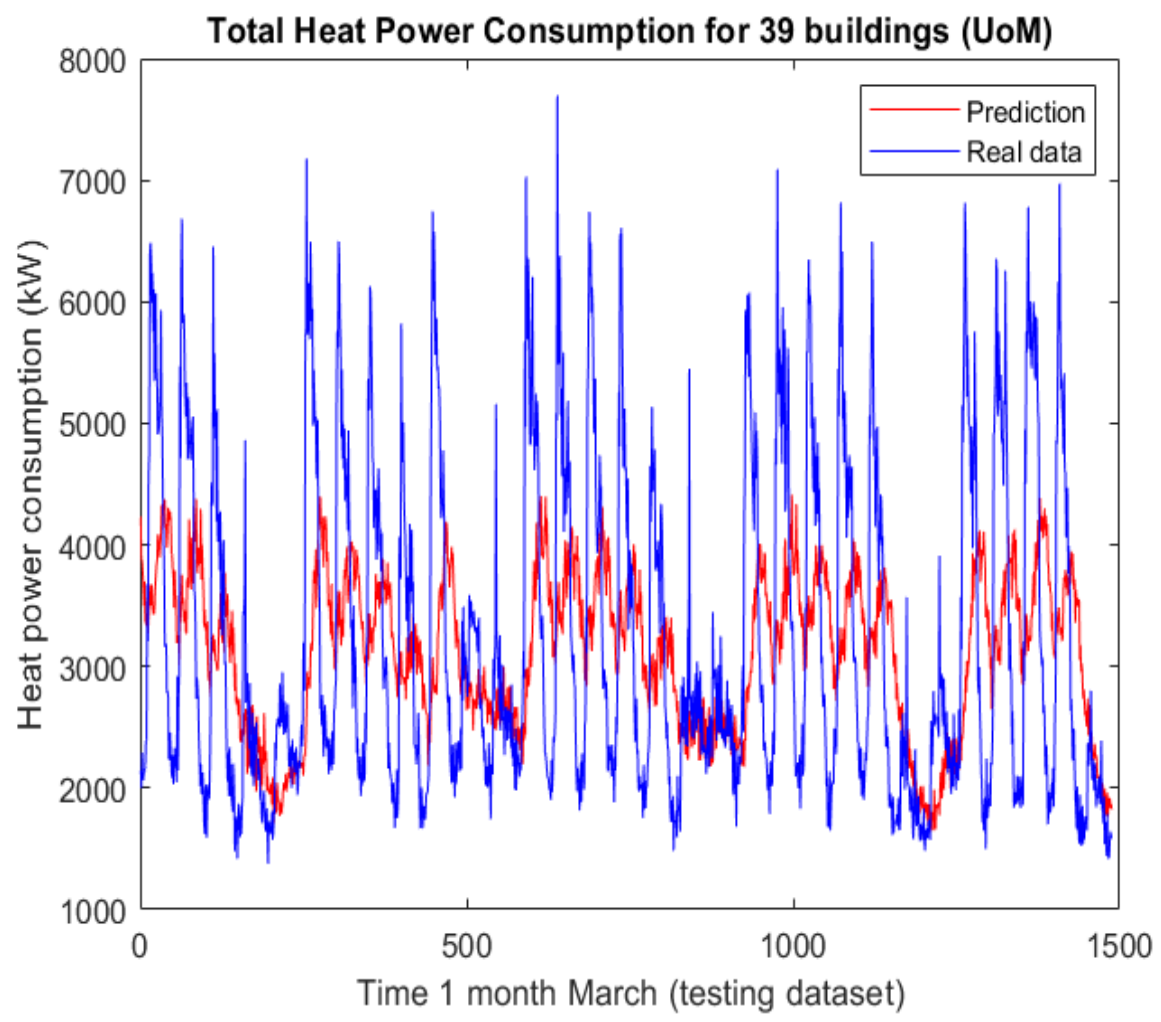

Fig.26. Predictions of the total heat power consumption obtained for 39 buildings (UoM) with the federated learning algorithm (CNN_6): testing dataset NRMSE = 0.1742, SNR =8.3593dB.

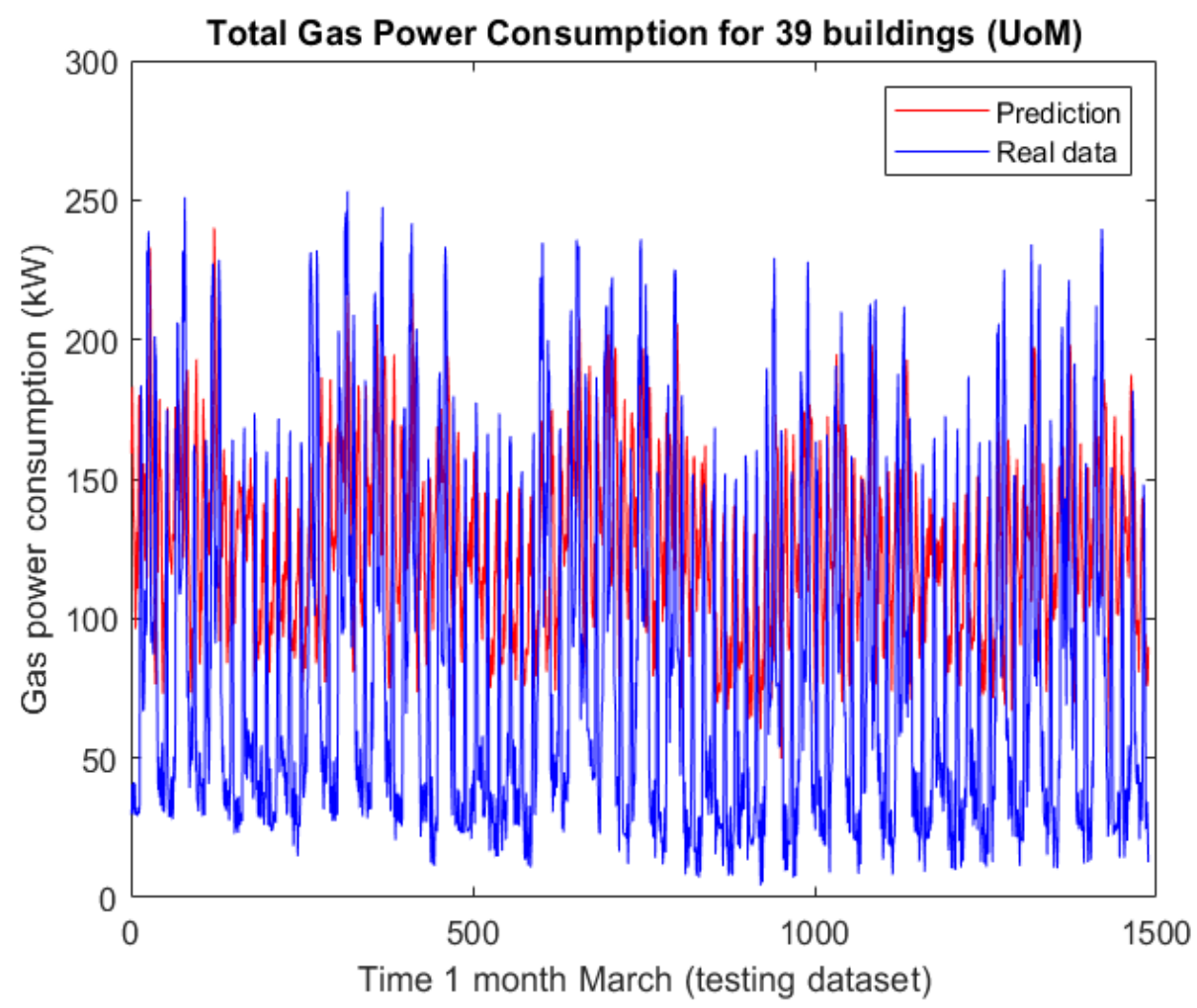

Fig. 27. Predictions of the total gas power consumption obtained with the federated learning algorithm (CNN_6): testing dataset NRSME=0.3131, SNR=2.15db.

*B.7 Summary of prediction results of the different CNN frameworks models*

The results shown in sections IVB.1, IVB.2, IVB.3, IVB.4, IVB.5 and IVB.6 are now summarized in Table IV. It can be noticed that the results from section IVB.3, which are using training, validation and testing datasets and shown on row 3 of Table IV (bolded and underlined), are better in terms of SNR and NRMSE than all the other results for both the training and the testing datasets, and comparable with the results shown on the 1st row (section IVB.1). The results from sections IVB1 and IVB.3 (i.e. rows 1 and 3 of Table IV) are obtained with 114 CNN models covering each 39 buildings and each type of electric, heat and gas energy vectors (i.e. 39 multiplied by 3).

It can be also noticed that on the second row of Table IV, the training and the testing results for the electric network are the same as the ones shown on the first row of Table IV, because there were no training/testing simulations performed. This is because for the electric network there is no case of CNN_2 model with multiple input variables and single output variable, because the electric consumption is the single input variable of the CNN_2 model, and which predicts solely the electric consumptions.

Improvements on the numerical predictions especially for the gas testing predictions can be searched by using also other well-established DL models such as the LSTM model (3 layers LSTM, 400 nodes per layer), a combination of CNN model and 1 layer LSTM model (400 nodes per LSTM layer), and a combination of CNN model and 3 layers LSTM model (400 nodes per LSTM layer). For comparison purposes, the numerical predictions obtained with these other DL models are shown in Table V and compared with the results from section IVB.1 and they exhibit similar accuracy levels as evaluated by SNR and NRMSE measures.

Finally, in the supplementary material (sections AA.2, AA.3, AA.4 and AA.5) there are shown several more results obtained for other months and training/testing conditions, and with the scope of further investigating the capabilities of the developed CNN models.

## V. Discussion & Conclusions

The contribution of the paper was first to study the correlation between the output variables and the input variables of the novel CNN prediction models. Therefore, the correlation between the energy vectors of the next 24 hours and the previous 24 hours energy vectors, weather temperature and solar radiance (i.e. input variables of CNN models) were studied. This showed high correlation between the next and the previous 24 hours power consumptions of the same type (i.e. electric-electric, heat-heat, gas-gas), and less correlations with regard to the solar radiance or the weather temperature (°C). As a novelty, the correlation study between the similar types of energy vectors (electric, heat and gas) across the different buildings, revealed high correlations between the electrical power consumptions of the different buildings. This explains that the federated learning (CNN_6) was able to predict successfully the electrical power consumptions for all the buildings and for the entire electrical network.

Second contribution was to explore whether for combined electric, heat and gas network systems can be devised novel CNN prediction models that accurately predict individually, or jointly the electric, the heat and the gas power network consumptions. The structures of the novel CNN models were devised and evaluated by using accuracy measures consisting of SNR and NRMSE indicators. The total predicted electric, heat and gas power consumptions were plotted for all the 39 building predictions. In total, there were devised six novel modelling frameworks based on CNNs. For the present example of integrated energy system, the CNN modelling technique (CNN_3) using single input variable and single output variable of the same energy vector type (e.g. heat input variable and heat output variable), and using training, testing and validation datasets showed to be a reliable and robust methodology in terms of accuracy of results and in comparison to the other modelling techniques.

Third, corroborated with the correlation studies, it was aimed to investigate whether as input variables of CNN models can be used either single or multiple input variables, with the latter for the situations where there are network connections between the electric, the heat and the gas network systems. The numerical results showed that both options are viable because, as already noted, when using a single input variable there exists for example very high correlations between the next and the previous 24 hours power consumptions for any of the three combined electric, heat or gas networks. While for the training results, there are noticed several differences between when using single or multiple input variables, and in the favour of using single input variables, then for the testing results there are not noticed any major differences.

Fourth contribution consisted of evaluating the quality of predictions, and it was clearly noticed that more scarce and non-linear was the training data then more difficult it was to predict the power consumptions. Because of these reasons, the gas network predictions were the most difficult to predict for this specific integrated electric, heat and gas network system. This is also in line with other findings from the literature.

Therefore, interesting guidelines can be provided in the following on developing similar predictive models for multi-energy demand: (I) the use of validation datasets allows the tuning of the parameters of the CNN models, which it would ensure better performances for the multi-energy testing results. Alternatively, the number of optimization epochs could be chosen in such a way that the over- or under-fitting of the predictive models could be avoided. In this work, it seemed that 400 epochs was the optimum choice for the single input/single output CNN models (CNN_1).

Other driving guidelines (II) are if more multi-scale energy data is to be used by the models, then more optimization epochs might be necessary to train the models, and with less number of convolutional layers for the hidden layers of the predictive models so that to ensure convergence: for example when a CNN model was used for the joint energy prediction (i.e. CNN_4 model) there were necessary two convolutional layers in the hidden layers and 2000 optimization epochs, in comparison to the three convolutional layers used in the hidden layers and 400 epochs for the single input/single output variable CNN_1 model.

Furthermore, (III) it should be taken into account the particularities of the integrated energy system consisting of the size of integrated energy system, the interconnections (i.e connections between the different networks) and interdependence (i.e. correlations between the different energy vectors) between the different energy vectors and how these can affect the forecasting: for example, as already described above, for the present example of integrated energy system consisting of three connected electric, heat and gas networks, the correlations were the highest between the energy vectors of the same type (i.e. electric-electric, heat-heat, gas-gas) and in comparison to other factors such as weather temperature or solar radiance.

Finally, (IV) different predictive models may produce varying performances for the different integrated energy systems depending on their particularities: for example, this work showed that federated learning for the electric network belonging to this integrated energy system is a viable option but not for the heat or the gas networks because of lower, non-linear and scarce power consumptions. This situation could be different for other integrated energy systems involving electric, heat and gas consumptions such as when for systems with lower electric and higher heat and gas power consumptions.

With regard to future work, it is aimed to be studied other realistic integrated energy systems datasets and to develop other novel data-driven predictive models of multi-energy network behavior, which to calculate future trajectories of frequency, voltage, water mass flow rates, pressures and other variables of interests. Otherwise, these variables could be also determined based on the CNNs predictions obtained in this paper.

| | Training | | | | | | Testing | | | | | |
|---|---|---|---|---|---|---|---|---|---|---|---|---|
| | Electric | | Heat | | Gas | | Electric | | Heat | | Gas | |
| | SNR | NRMSE | SNR | NRMSE | SNR | NRMSE | SNR | NRMSE | SNR | NRSME | SNR | NRSME |
| 1) Single input variable and single output variable (CNN_1 Section IV.B1) | 38.84 dB | 0.007 | 26.77 dB | 0.019 | 22.59 dB | 0.0293 | 34.63 dB | 0.0085 | 17.22 dB | 0.0628 | 8.86 dB | 0.147 |
| 2) Multiple input variables and single output variables (CNN_2 Section IV.B2) | **38.18 dB** (*the same value from third row) | **0.0077** (*the same value from third row) | 24.10 dB | 0.026 | 20.22 dB | 0.0385 | **34.87 dB** (*the same value from third row) | **0.0116** (*the same value from third row) | 17.42 dB | 0.061 | 8.86 dB | 0.14 |
| 3) Single input variable and single output variable and using validation dataset (CNN_3 Section IV.B3) | **38.18 dB** | **0.0077** | **19.68 dB** | **0.044** | **15.26 dB** | **0.0681** | **34.87 dB** | **0.0116** | **17.84 dB** | **0.0584** | **9.18 dB** | **0.151** |
| 4) Single CNN model for the entire integrated energy system (CNN_4 Section IV.B4) | 31.43 dB | 0.017 | 22.76 dB | 0.031 | 13.43 dB | 0.017 | 18.18 dB | 0.056 | 10.26 dB | 0.14 | 2.58 dB | 0.298 |
| 5) Single CNN model for each energy network (CNN_5 Section IV.B5) | 29.75 dB | 0.0229 | 12.71 dB | 0.102 | 9.92 dB | 0.1261 | 19.70 dB | 0.0664 | 4.79 dB | 0.2626 | 3.18 dB | 0.278 |
| 6) Federated learning with single input variable (CNN_6 Section IV.B6) | 29.50 dB | 0.0210 | 7.904 dB | 0.171 | 3.63 dB | 0.2602 | 29.72 dB | 0.0210 | 8.35 dB | 0.1742 | 2.15 dB | 0.313 |

Table IV. SNR and NRMSE values obtained for the training and the testing dataset for the electric, the heat and the gas networks with the following simulations: 1) Single input variable and single output variable (CNN_1: Section IV.B1) , 2) Multiple input variables and single output variables (CNN_2: Section IV.B2), 3) Single input variable and single output variable and using training/validation/testing datasets (CNN_3: Section IV.B3), 4) Single CNN model for the entire integrated energy system (CNN_4: Section IV.B4), 5) Single CNN model for each energy network (CNN_5: Section IV.B5), 6) Federated learning with single input variable (CNN_6: Section IV.B6). (* there is no case of multiple input variables and single output variable CNN model as the previous 24 hour electric consumption is the single input variable of the CNN model, which predicts the next half hour electric consumption, which is also the output variable).

| Deep Learning architectures for gas power prediction | Total gas power consumption prediction | |
|---|---|---|
| | SNR | NRMSE |
| 3 layers CNN (CNN_1: section IVB.1) | 8.68 dB | 0.1477 |
| CNN + 3 layers LSTM | 8.73 dB | 0.1440 |
| CNN + 1 layer LSTM | 8.42 dB | 0.1501 |
| 3 layers LSTM | 8.84 dB | 0.1450 |

Table V. SNR and NRMSE values obtained for the gas network testing dataset with the different DL models.

## Supplementary Material

This section includes several more results on the application of the CNN model to the available data of the integrated electric, heat and gas energy system.

*AA1. Epochs number for training stage*

For the training method (I) described in the results section IV.B, in order to find the optimum number of epochs, we chose several buildings with relevant power consumptions, which are higher than zero and non-scarce as these are the major power energy consumers in a network. We plot the NRMSE and the SNR values for the testing datasets shown function of the number of epochs used during the training. The optimum number of epochs would be the one with the smallest NRMSE and the highest SNR from a series of training trials with 200 epochs, 400 epochs, 600 epochs, 800 epochs, 1000 epochs, 1200 epochs and 2000 epochs. The difference between these training trials consists of the computational times required to train the CNN model(s), and more epochs obviously would result in a higher computational time. In conclusion, it is desired to identify the number of epochs for which the testing dataset has the smallest NRMSE and the highest SNR, and also a reduced computational time during the training.

Fig.A1 and Fig.A2 show the NRMSE and the SNR values plotted for the electrical testing datasets for two different buildings. It can be observed that in both figures the SNRs reach the highest value around 400 epochs and the smallest NRMSE values also around 400 epochs (i.e. 7 minutes computational time).

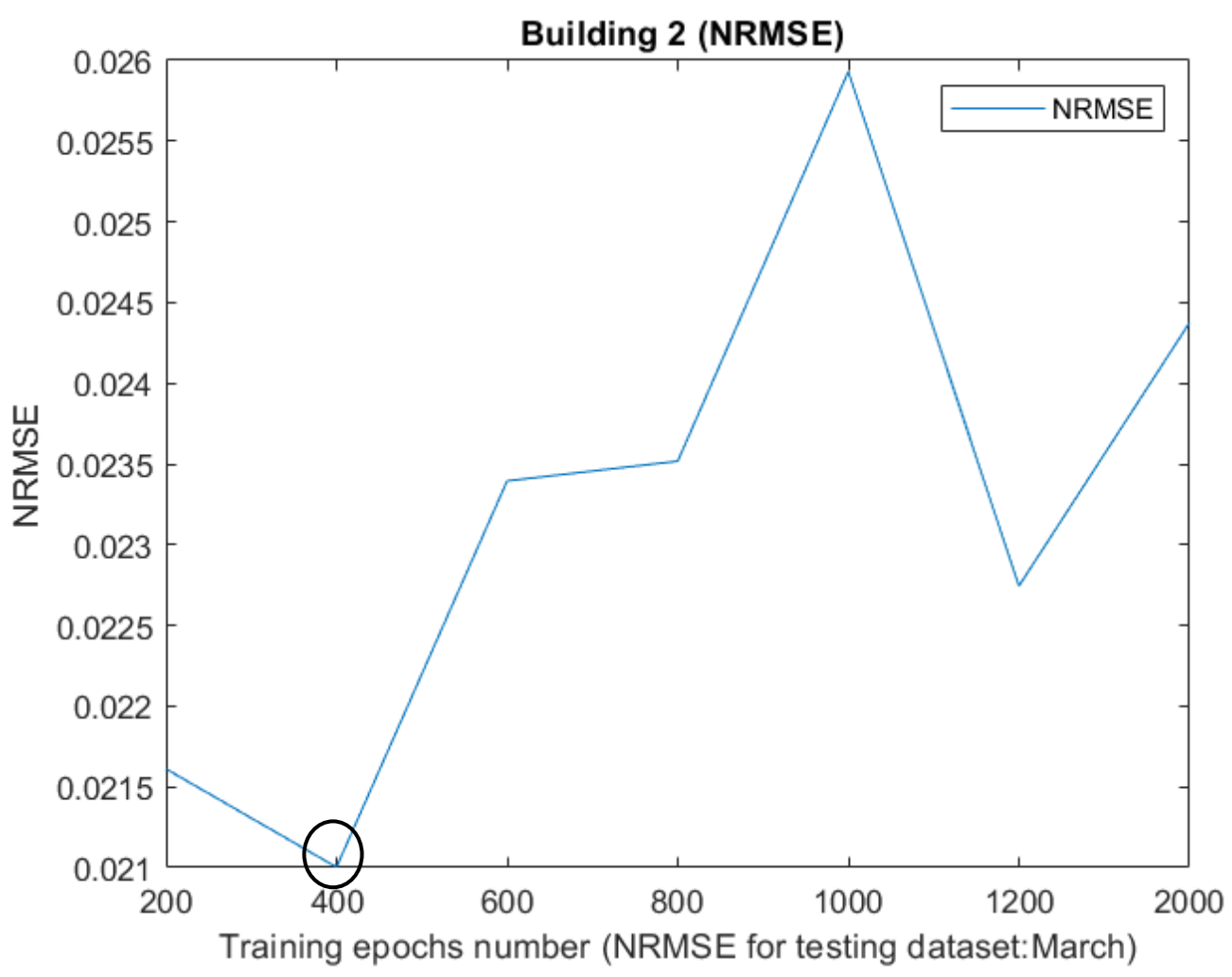

a) NRMSE

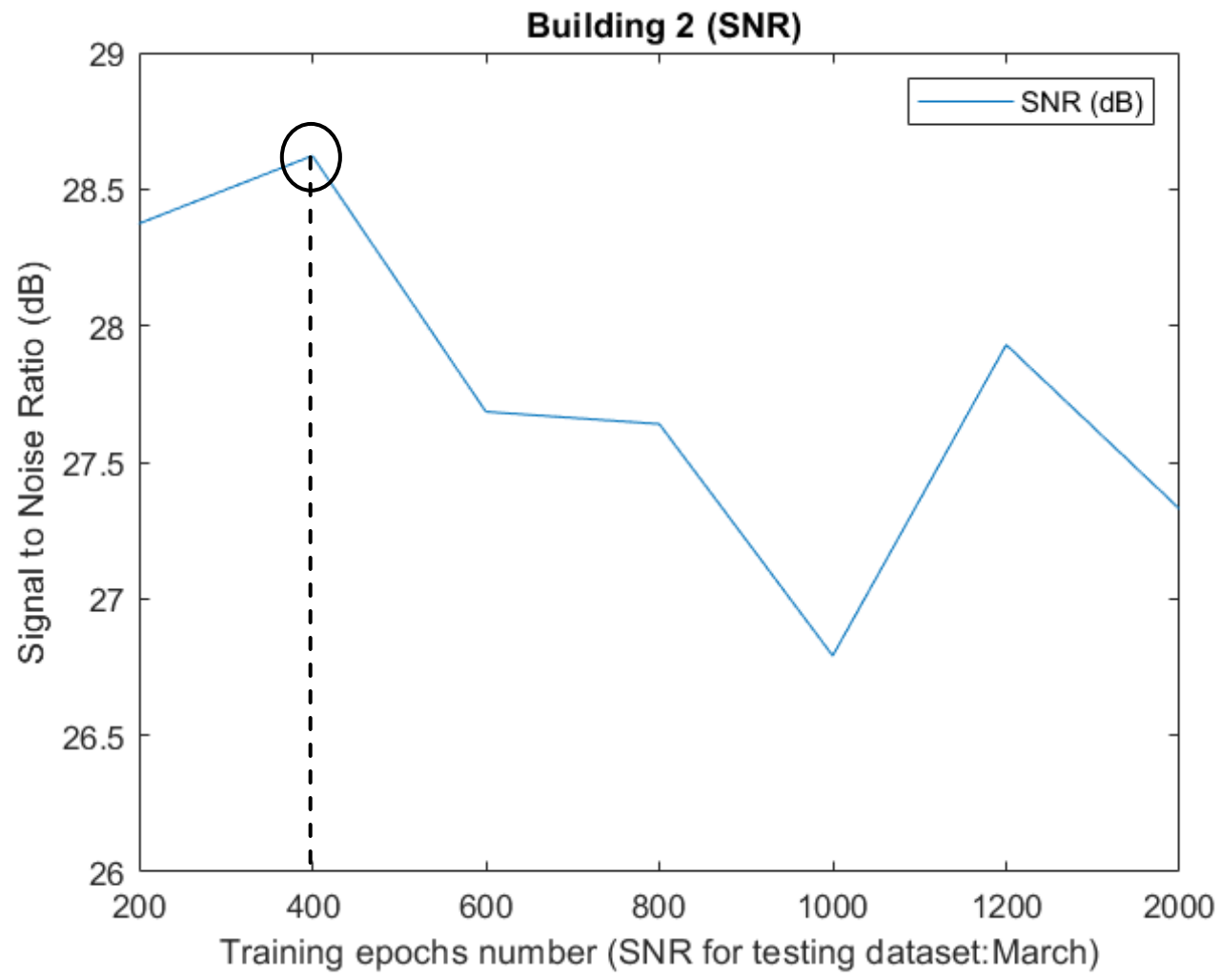

b) SNR

Fig. A1. NRMSE and SNR values for an electrical testing dataset for a building function of the number of epochs used during the training.

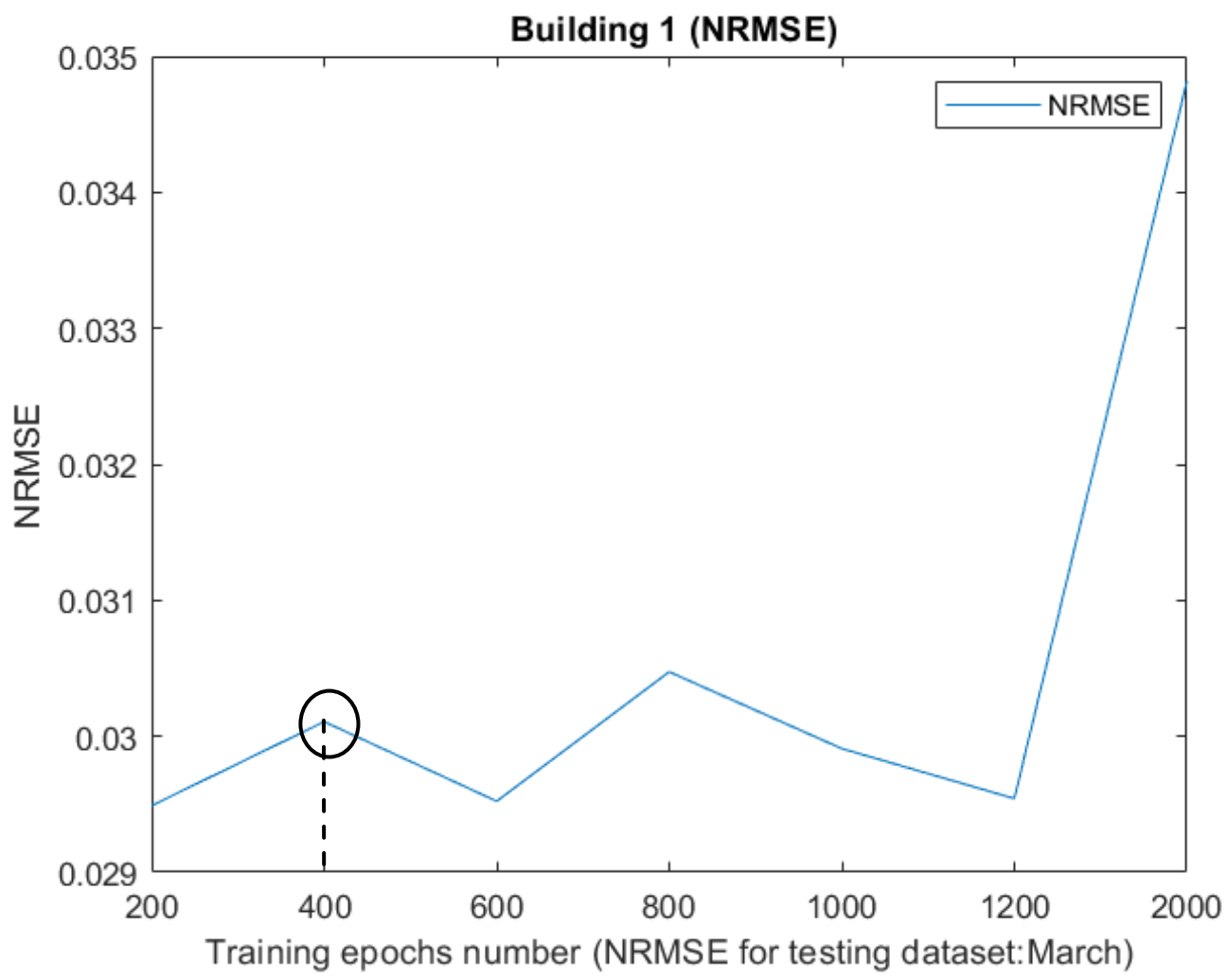

a) NRMSE

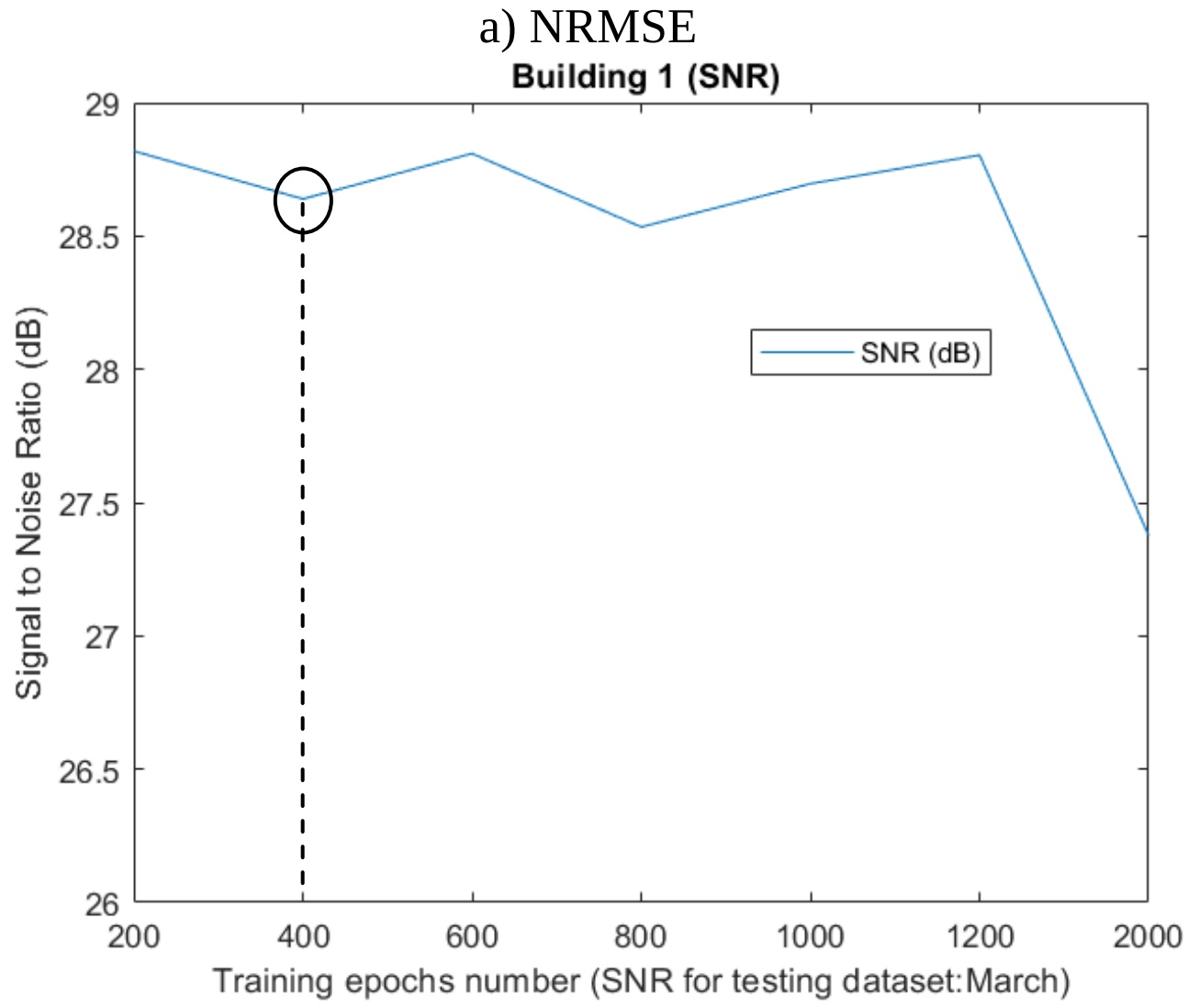

b) SNR

Fig. A2. NRMSE and SNR values for an electrical testing dataset for a building and shown function of the number of epochs used during the training.

Fig.A3 and Fig.A4 show the NRMSE and the SNR values obtained for the heat testing datasets and shown again function of the number of epochs used during the training.

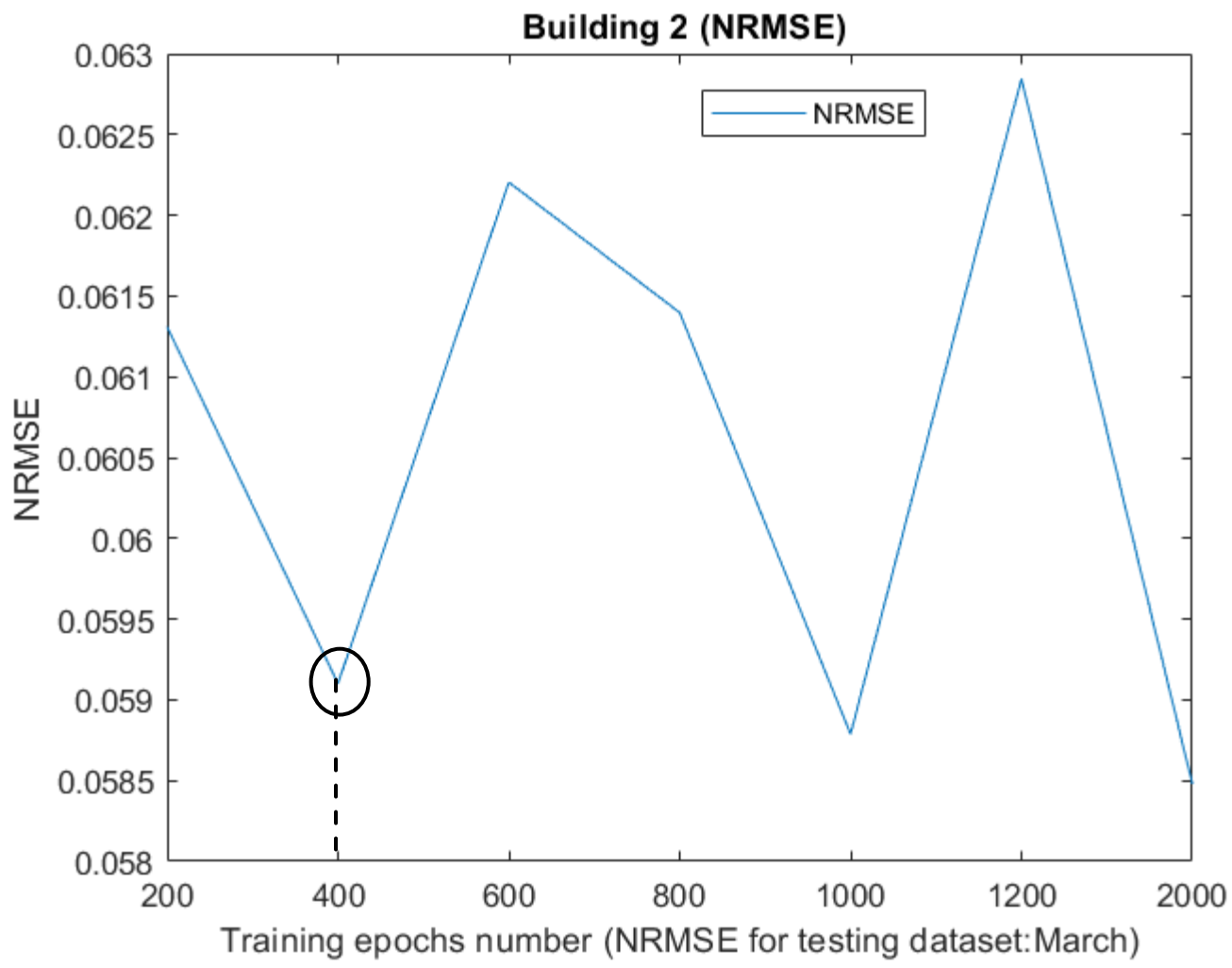

a) NRMSE

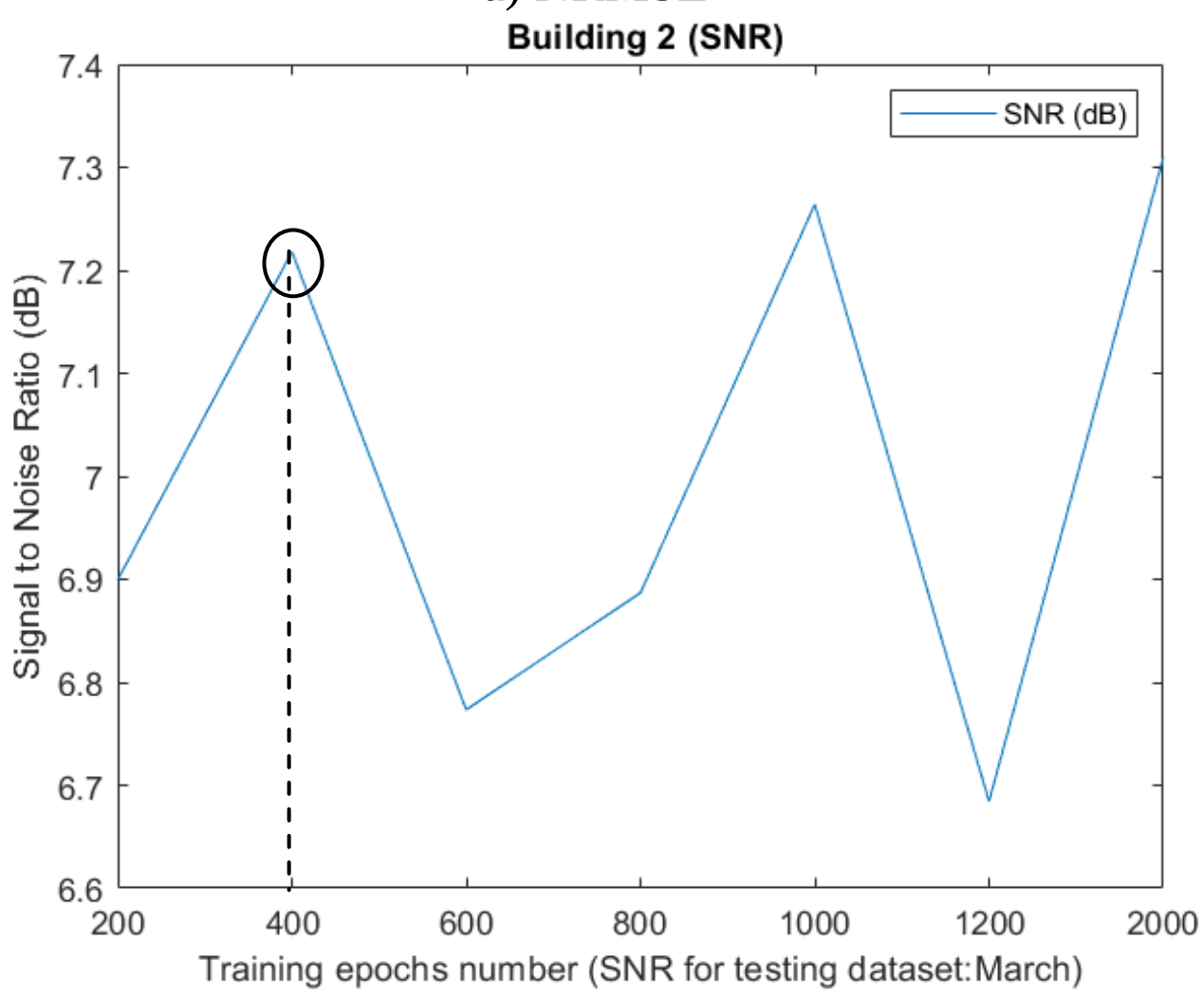

b) SNR

Fig. A3. NRMSE and SNR values for heat testing dataset for a building function of the number of epochs used for the training dataset.

It can be observed (Fig.A3) that 400 epochs is a good candidate for the training and the testing stages, although for example 2000 epochs could be as well a very good candidate solution. However, 2000 epochs would mean to increase the computational time, which would be more than half an hour.

In another example (Fig.A4), again for the heat training/testing buildings datasets, the 400 epochs do not seem to be the optimum choice but it is close to the other better options, which are 200 (<4%) and 1200 epochs (<4%).

For the gas testing datasets, there are shown in Fig.A5 the NRMSE and the SNR values obtained for the gas power consumptions predictions of a building and function of the number of epochs used during the training. Again, 400 epochs (i.e. 7 minutes computational time) seem to be a good choice although 1200 (<4%) or 2000 (<5%) epochs seem to be a better epochs number, but again also more expensive in terms of computational time (i.e. around half an hour computational time).

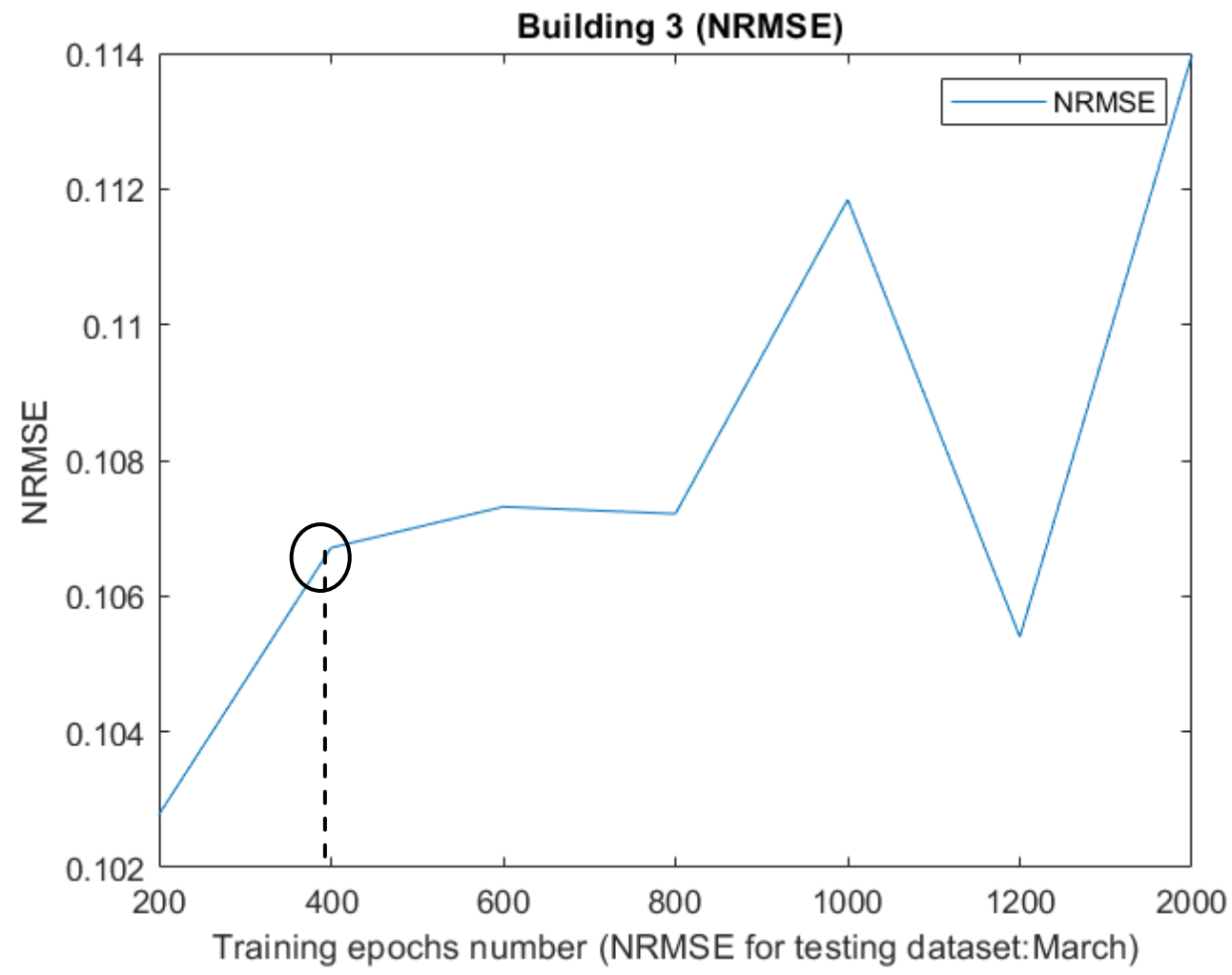

a) NRMSE

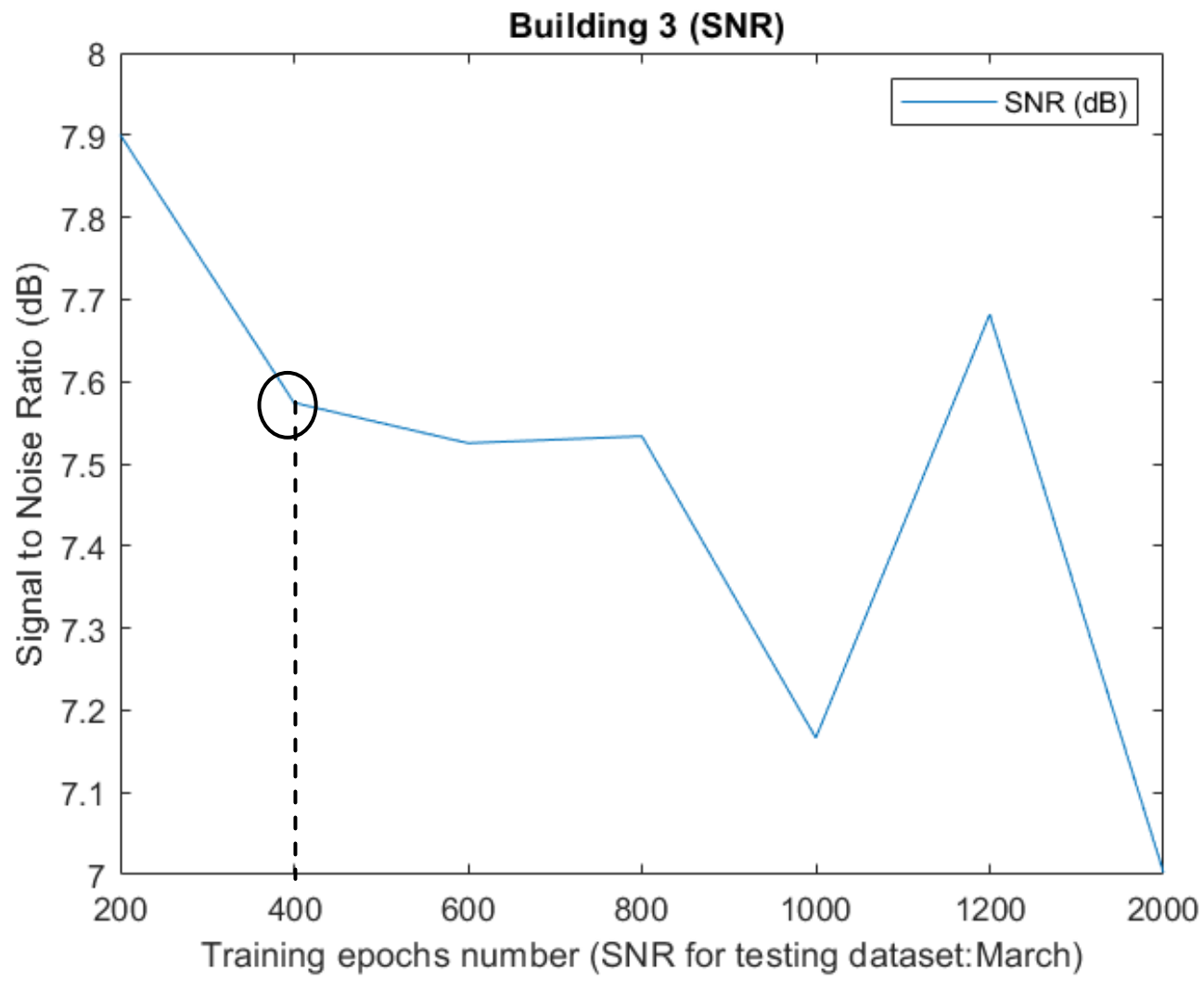

b) SNR

Fig. A4. NRMSE and SNR values for heat testing dataset for a building shown function of the number of epochs used for the training dataset.

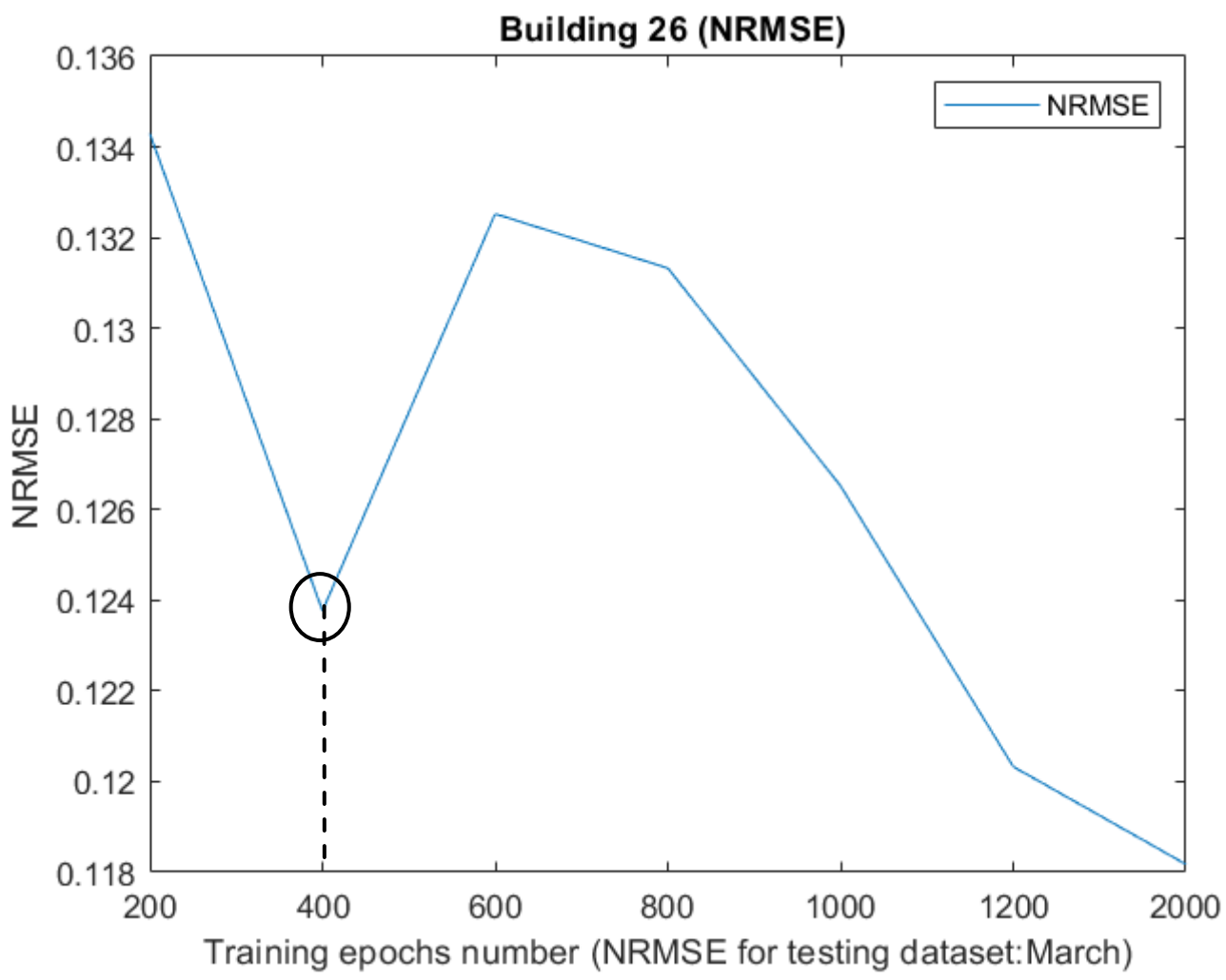

a) NRMSE

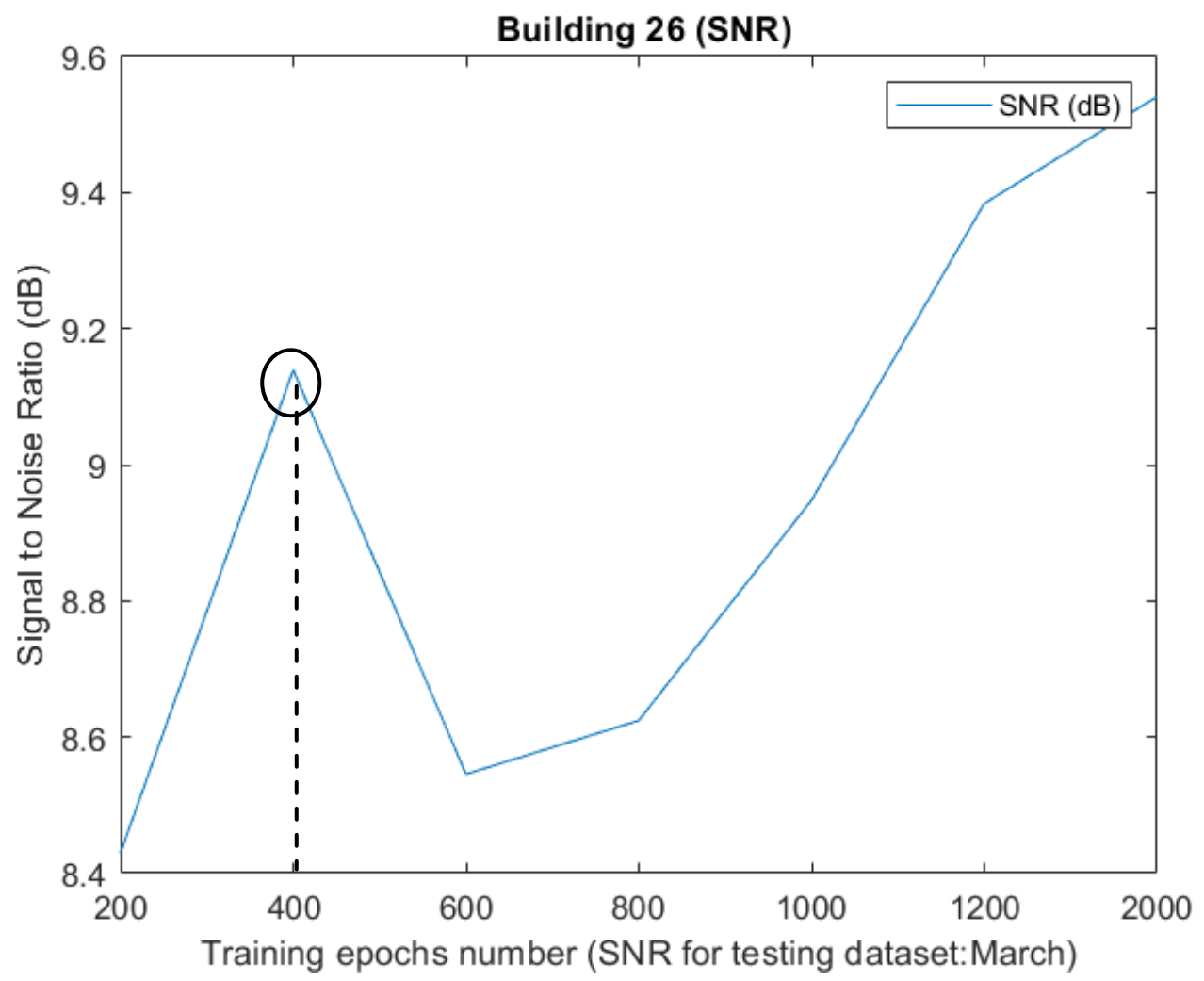

b) SNR

Fig. A5. NRMSE and SNR values for gas testing dataset for a building shown function of the number of epochs used for the training dataset.

Finally, another example is shown in Fig.A6 with the NRMSE and the SNR values obtained for the gas power consumptions predictions of another building and again function of the number of epochs used during the training. It is 400 epochs (i.e. 7 minutes computational time), which seems to be a very good choice.

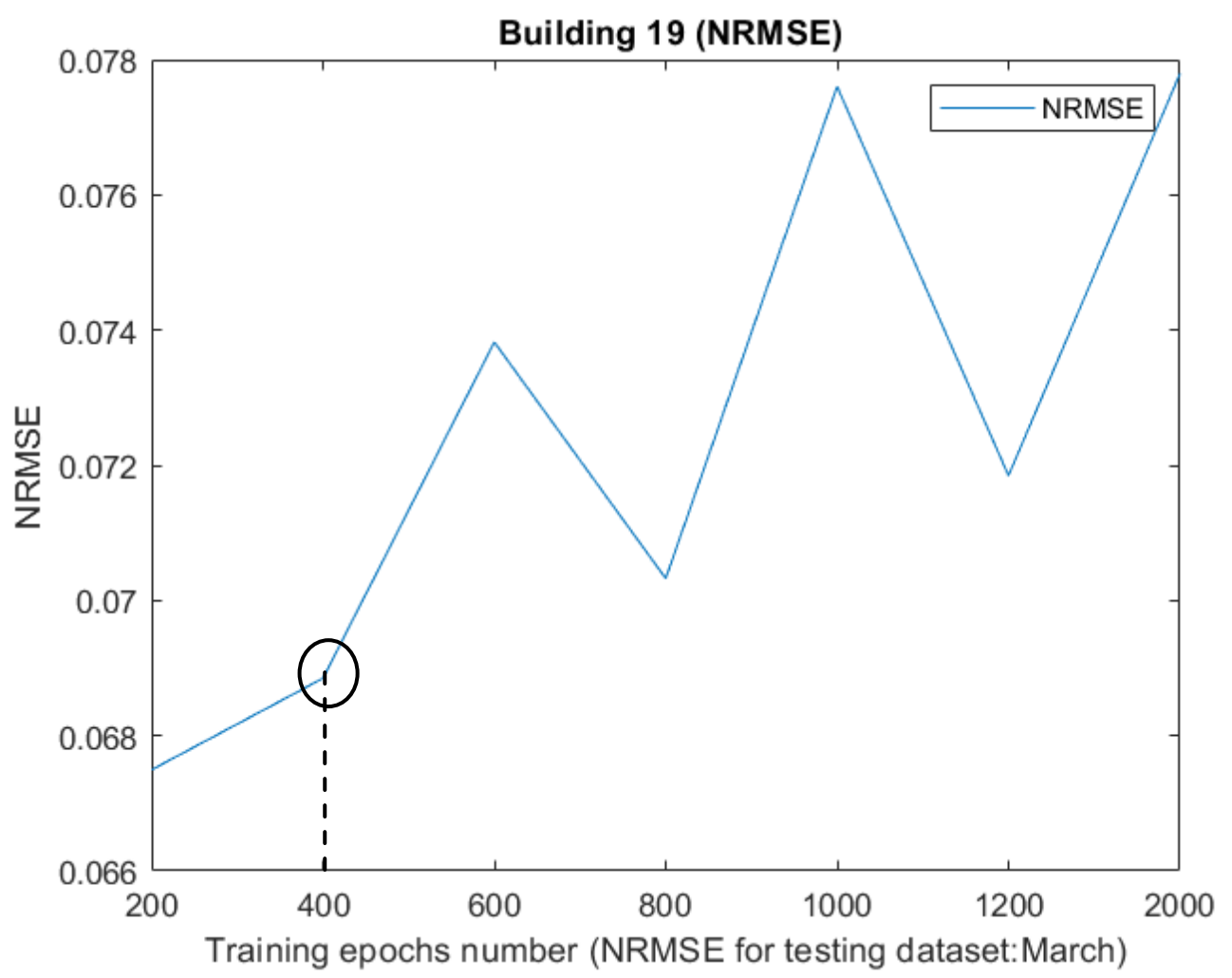

a) NRMSE

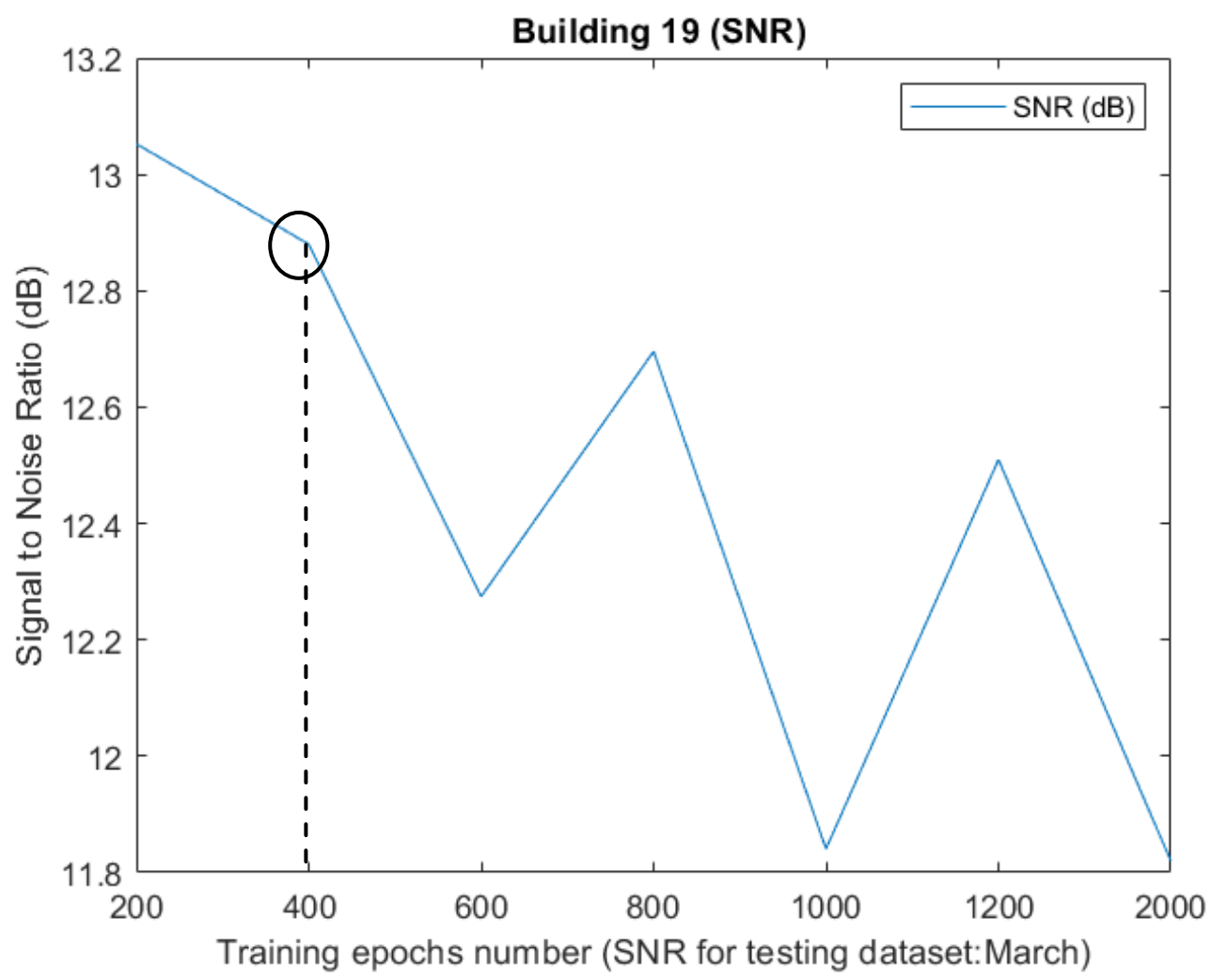

b) SNR

Fig. A6. NRMSE and SNR values for a gas testing dataset for a building function of the number used for the training dataset.

*AA.2 Short-term CNN predictions of power consumptions using weather temperature as an additional input variable*

In the main part of the paper there were presented short-term predictions obtained with CNN models, which had as input variables the previous 24 hours power consumptions. In this section, there are shown more results for when CNN models have additional input variables, namely the weather temperatures or the solar radiance.

In Fig.A7 there are shown the training (January and February 2013) and the testing (March 2013) results for predicting the electrical power consumptions for building 1 (Fig. 2) and for when the input variables are the previous 24 hours electrical power consumptions and the weather temperature data (°C).

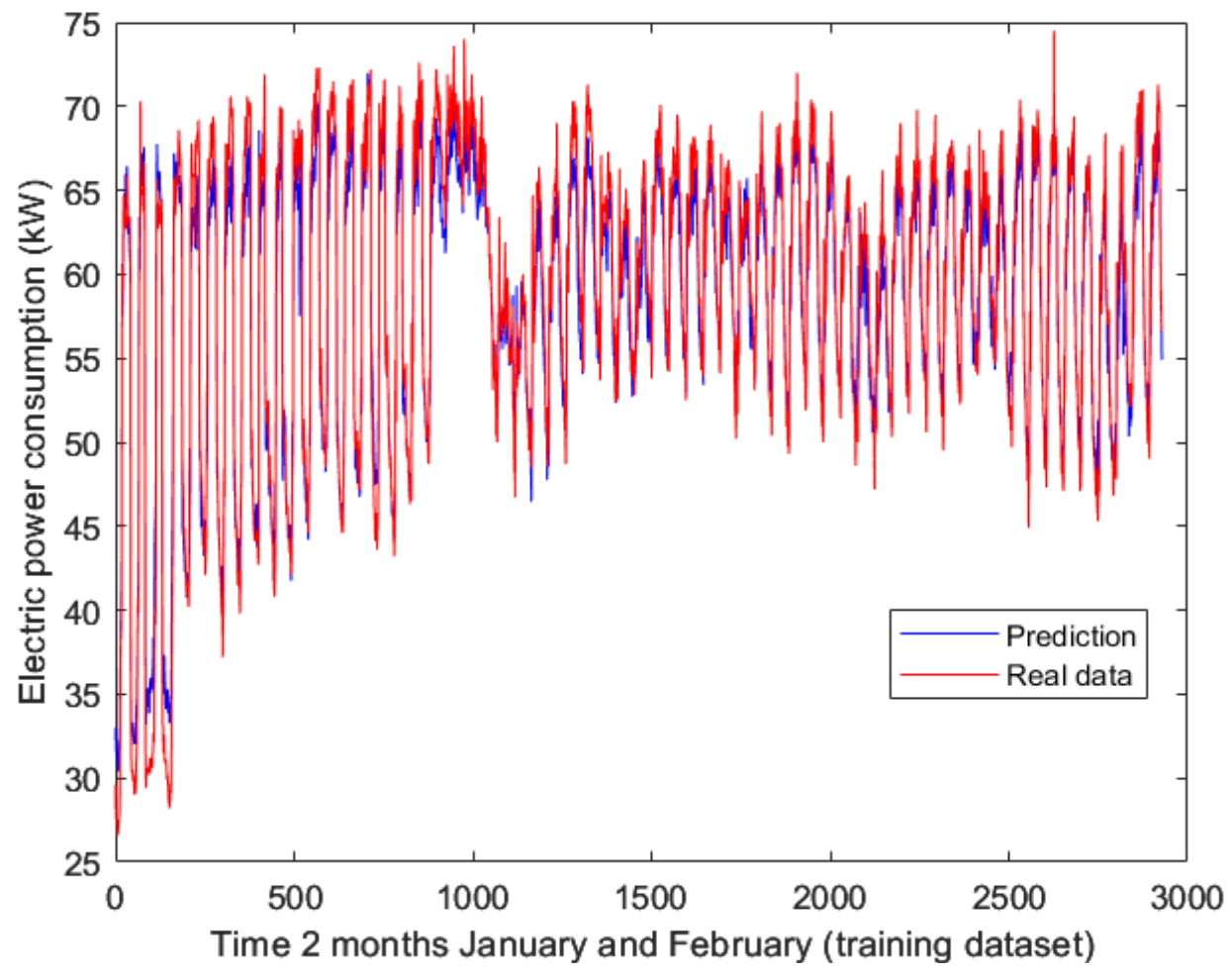

a) training results: SNR =29.21dB /29.74 dB, NRMSE =0.0288 / 0.0271

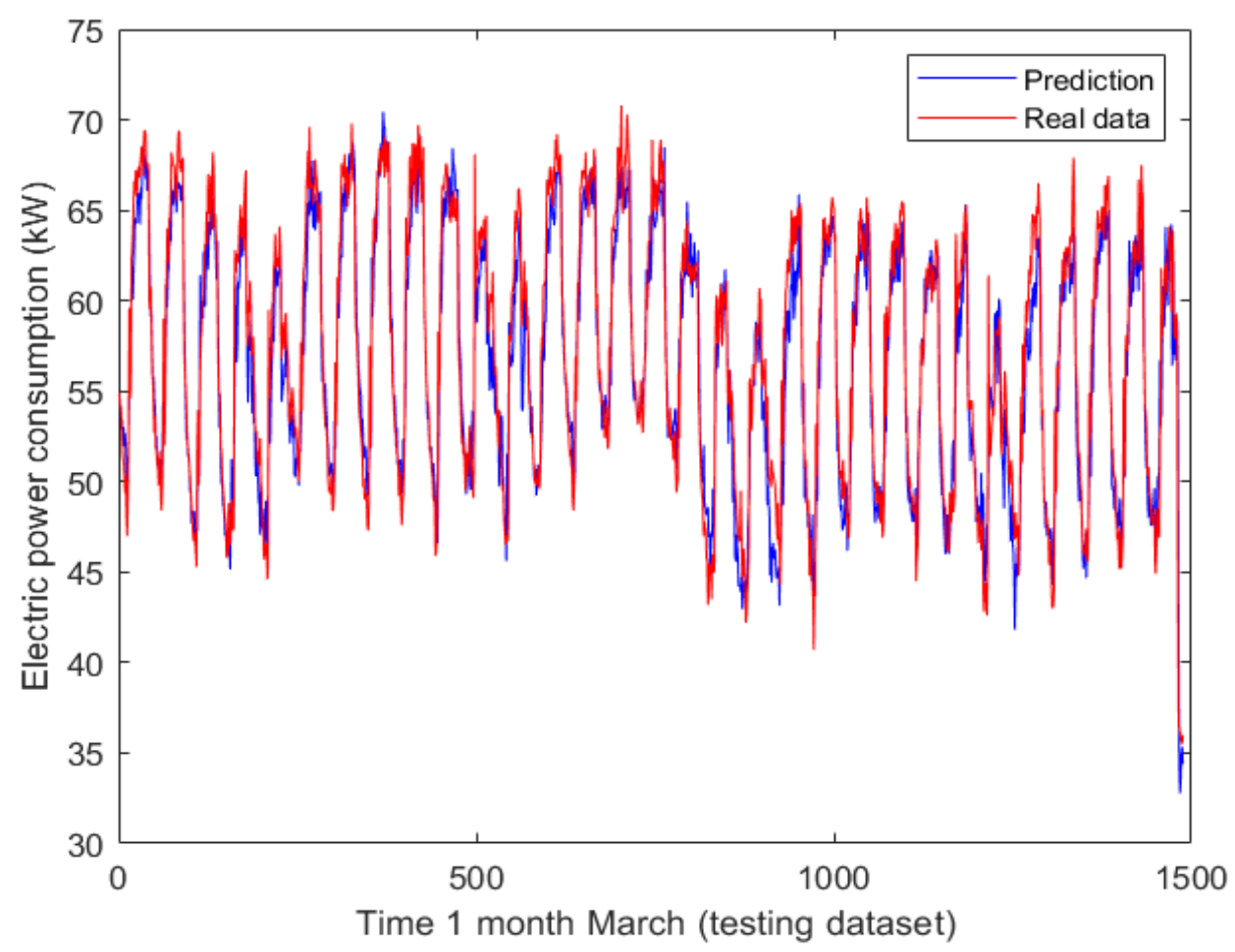

b) testing results: SNR =28.20dB / 28.84dB, NRMSE =0.0316 / 0.029.

Fig. A7. Electrical prediction results with input variables as the previous 24 hours electrical power consumptions and the weather tempereatures data (building 1).

In Fig.A7, the underlined numbers shown in the caption of the figure correspond to when only the previous 24 hours electrical power consumptions were used as input variable.

In Fig.A8, there are shown the gas prediction results with input variables as the previous 24 hours gas power consumptions and the weather temperatures, and for building 26 and for the same periods of time (January to March): the underlined underlined numerical results are the ones obtained for when only the previous 24 hours electrical power consumptions were used as input variable to the CNN model.

Overall, Figs.A7 and A8 do not show anything in support or against using weather temperatures as a second input variable for the CNN model. This could be well explained by the fact that the energy power consumptions within each of the 39 buildings are a direct reflection of the weather temperatures and the solar radiance from outside of the buildings.

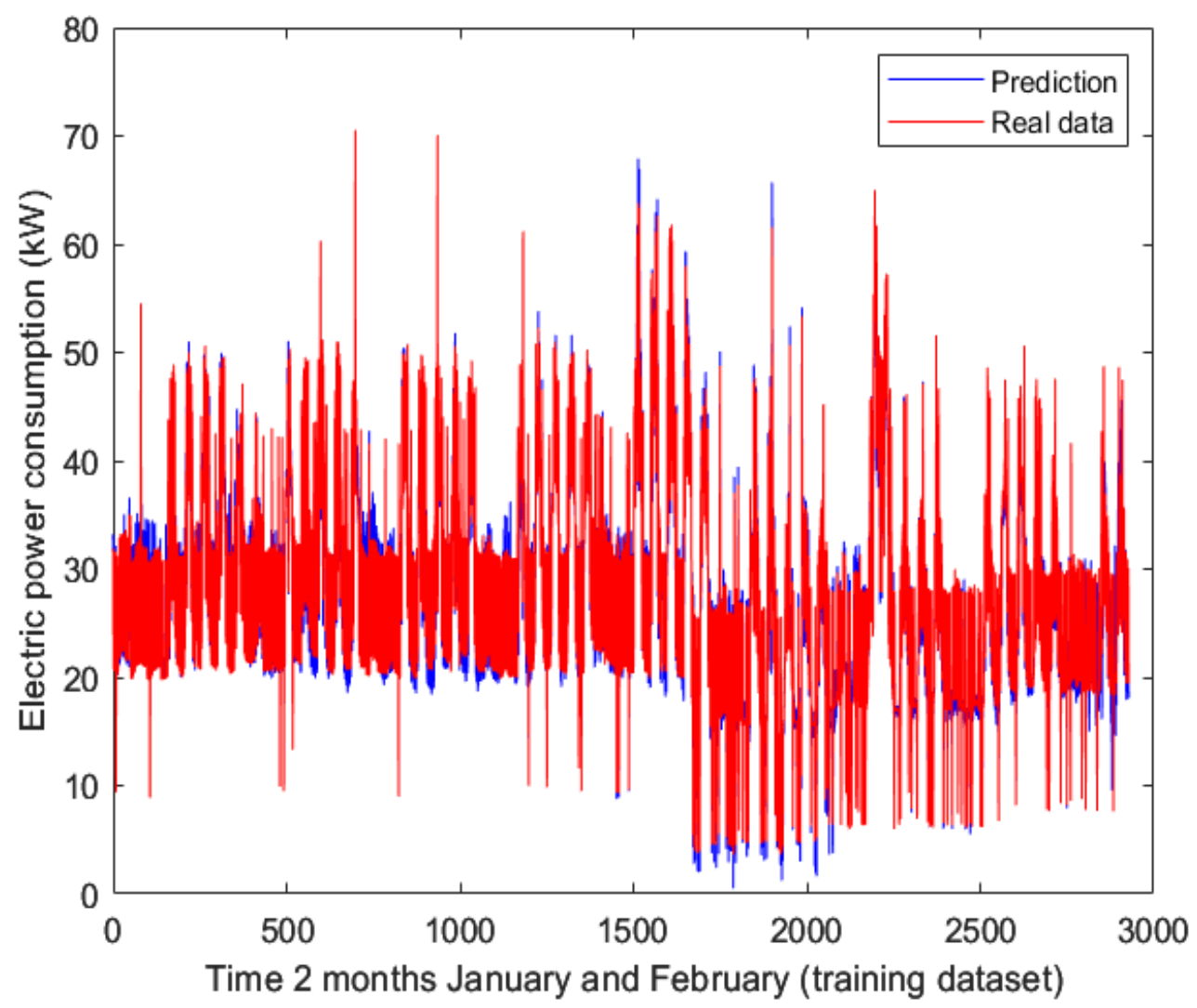

a) training results: SNR =22.69dB /21.13 dB, NRMSE =0.0321 / 0.0383.

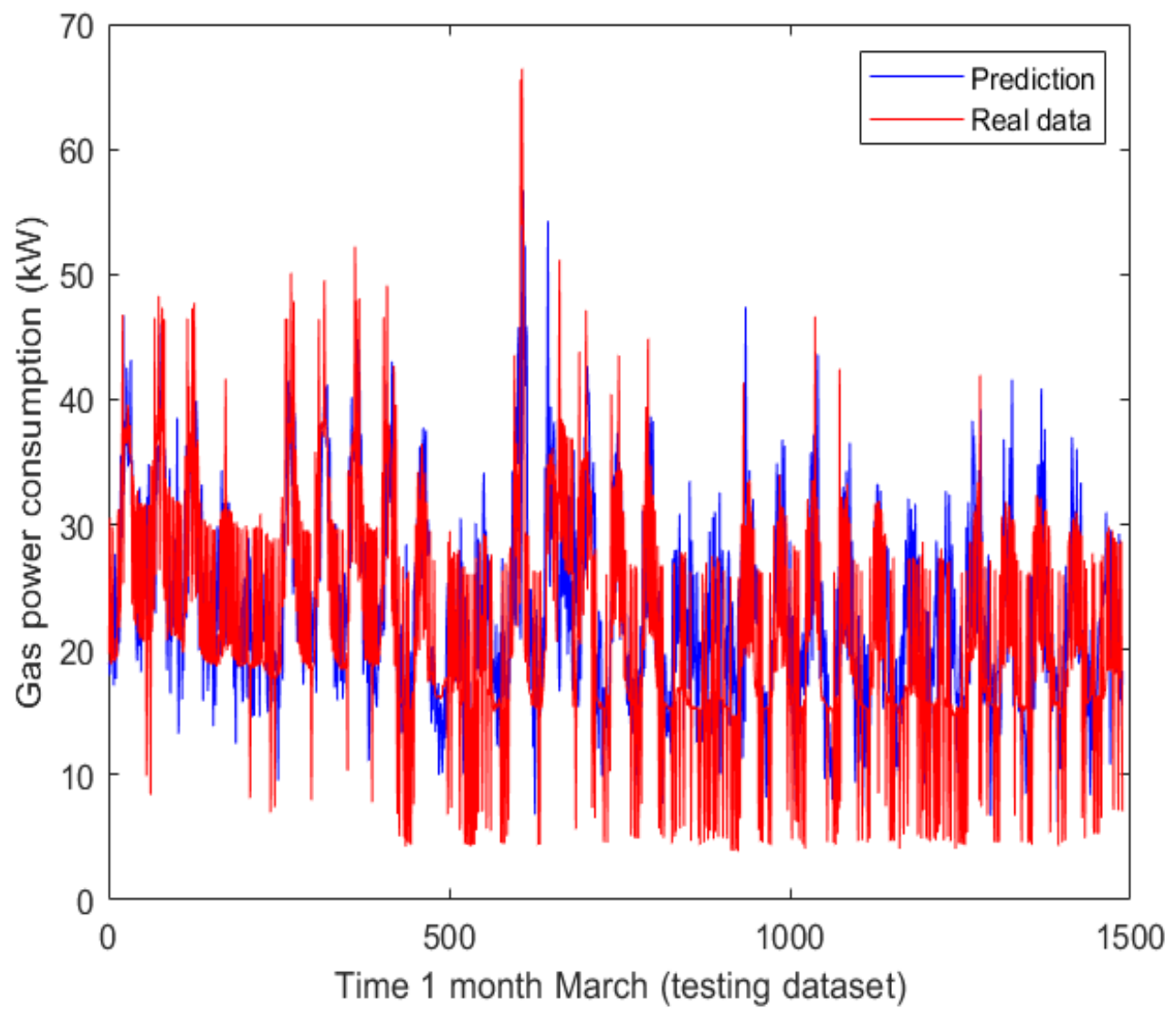

b) testing results: SNR =9.1873dB / 8.24dB, NRMSE =0.1231 / 0.1371.

Fig. A8. Gas prediction results with input variables as the previous 24 hours gas power consumptions and the weather tempereatures (building 26).

### *AA.3 Short-term CNN predictions of electrical power consumptions for a warmer period of time during a year*

The prediction results shown so far in the paper were for the colder season (January, February and March) when the power consumptions are higher than usual. Therefore, in Fig.A9 is investigated the robustness of the trained CNN models for a period of time (March, April and May 2013) when the temperatures are increasing gradually. It is also compared the situation when the input variable of the CNN model is only the previous 24 hours electrical power consumption versus when there are two input variables, the previous 24 hours electrical power consumption and the weather temperatures (Fig.A9).

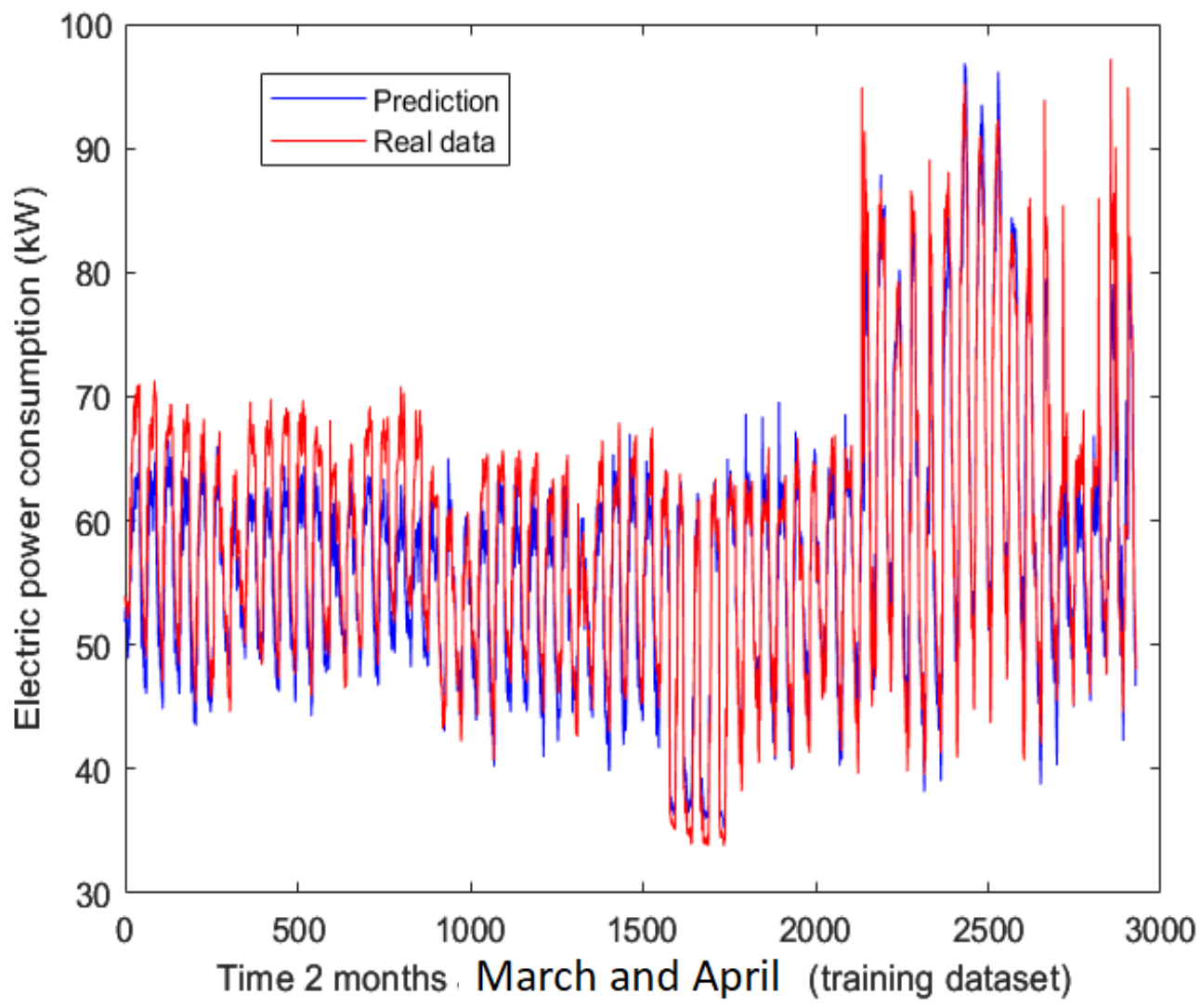

a) training results: SNR = 23.62dB /21.69dB, NRMSE =0.0405 / 0.0357.

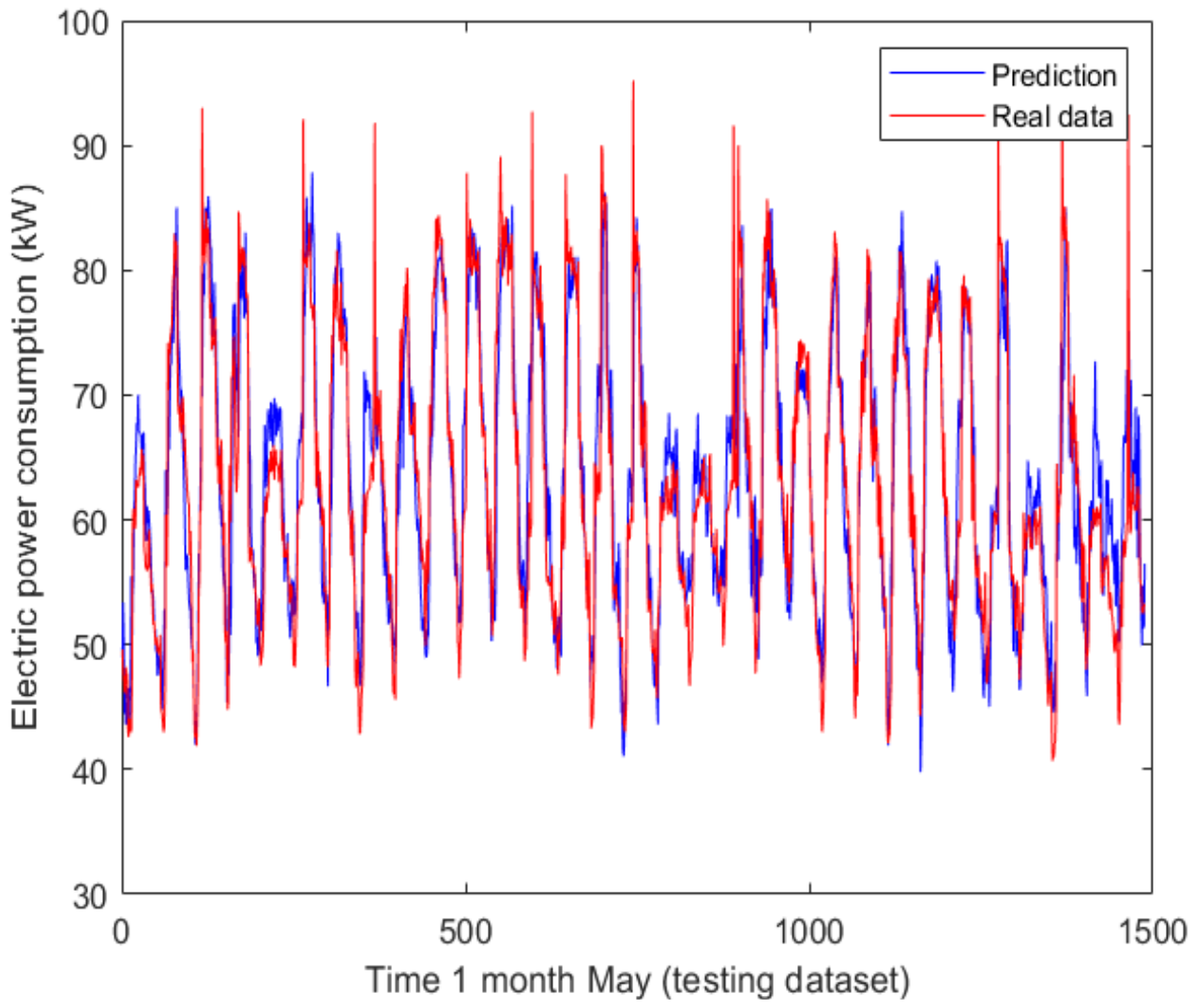

b) testing results: SNR =22.17dB/25.01dB,
NRMSE =0.0526 / 0.0367.

Fig. A9. Electrical prediction results (building 1) with input variables as the previous 24 hours electrical power consumptions and the weather tempereatures and for a period of time (March, April and May 2013) characterized by a gradual increased of the weather temperatures.

As already mentioned in Fig.A9 there are shown the electrical prediction results (building 1) obtained with the input variables as the previous 24 hours electrical power consumptions and the weather tempereatures (°C) and for a period of time (March, April and May 2013) characterized by the gradual increased of the weather temperatures. The above results are for building 1, and while the training results seem to be slightly better in terms of SNR and NRMSE, the testing results do not show better when using the weather temperatures as an additional input variable to the previous 24 hours electrical power consumptions and for this warmer period of the year (i.e. March, April and May).

*AA.4 Short-term CNN predictions of electrical power consumptions using solar radiance as an additional input variable*

Besides the weather temperatures (°C), the solar radiance can be used also as an additional input variable. Therefore, it is now compared the prediction results obtained with the input variable of the CNN model as being the previous 24 hours electrical power consumption versus when there are two input variables, the previous 24 hours electrical power consumption and the solar radiance (i.e. building 1). While the training results (Fig.A10(a)) seem again to be slightly better in terms of SNR and NRMSE, the testing results (Fig.A10(b)) do not show an improvement when using the solar radiance as an additional input variable to the previous 24 hours electrical power consumptions.

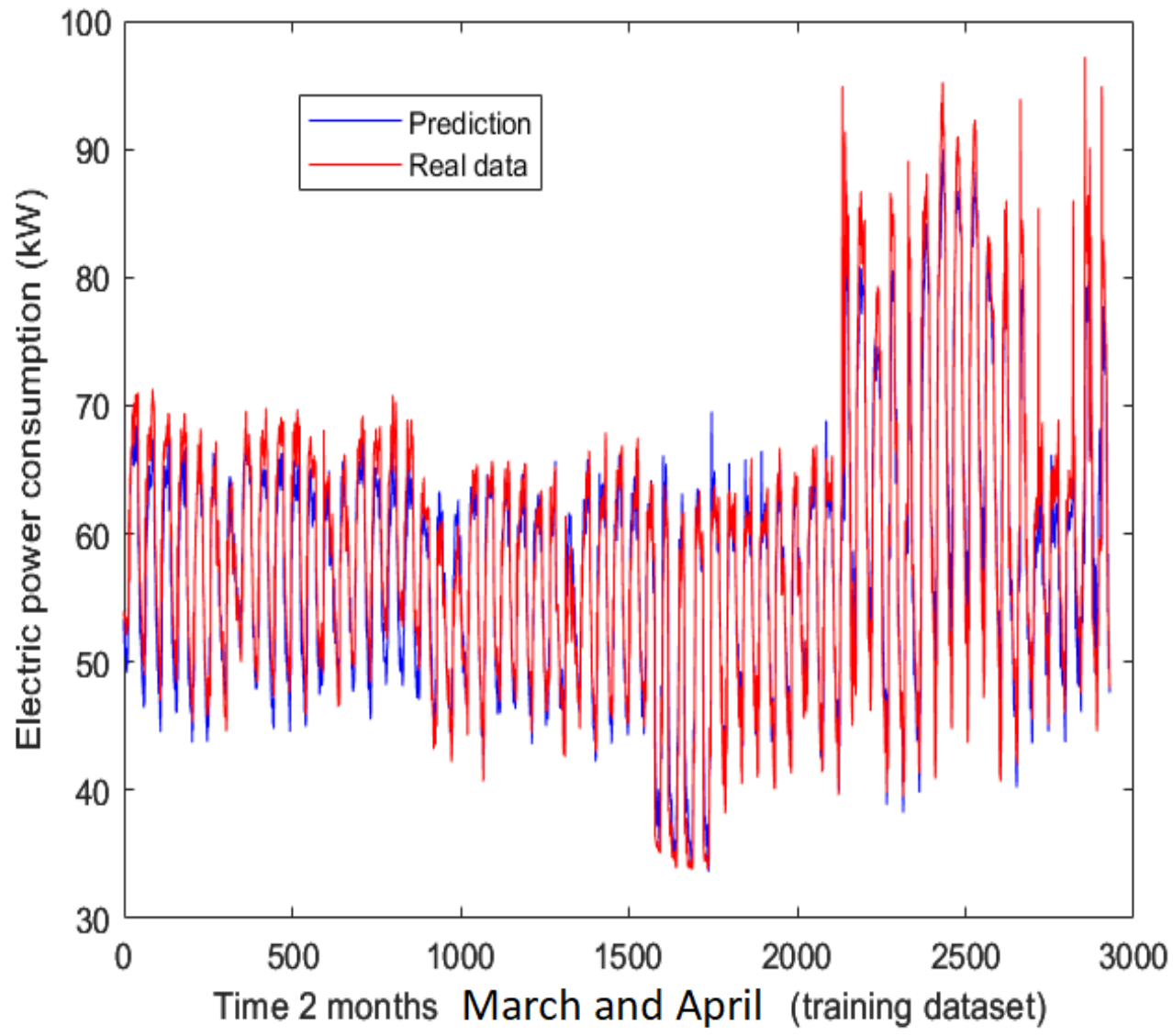

a) training results: SNR = 25.14dB / 21.69dB,
NRMSE = 0.0366/ 0.0357.

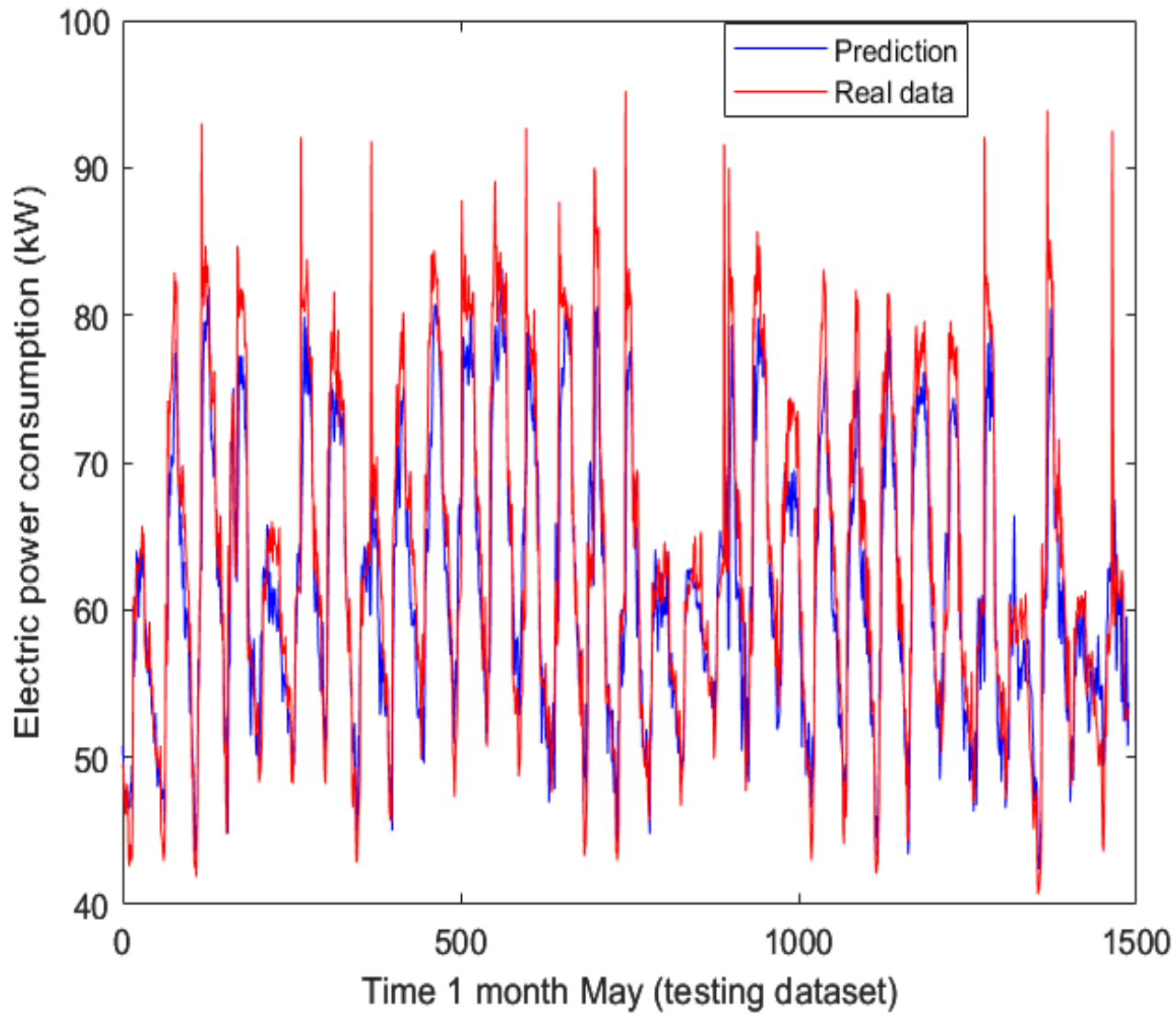

b) testing results: SNR =22.01dB/25.01dB,
NRMSE = 0.0535/0.0367.

Fig.A10. Electrical prediction results (building 1) with input variables as the previous 24 hours electrical power consumptions and the solar radiance, and for a period of time (March, April and May 2013) characterized by gradual increased of the weather temperatures (°C).

*AA.5 Short-term CNN predictions of electrical power consumptions using weather temperature as an additional input variable and with all input data scaled between 0 and 1*

This set of results is for when the input variables of the CNN models are the previous 24 hours electrical power consumptions and the weather temperatures (°C), and all the input data is scaled between 0 and 1. Fig.A11 shows the electrical power consumption prediction results obtained for building 1 and for the period of time January to March, and for when all the input variables are scaled between 0 and 1. The input variables are the previous 24 hours electrical power consumptions and the weather temperaturees (°C).

Again, the underlined numbers correspond to the CNN model results, which was using only the previous 24 hours electricl power consumptions as the input variable.

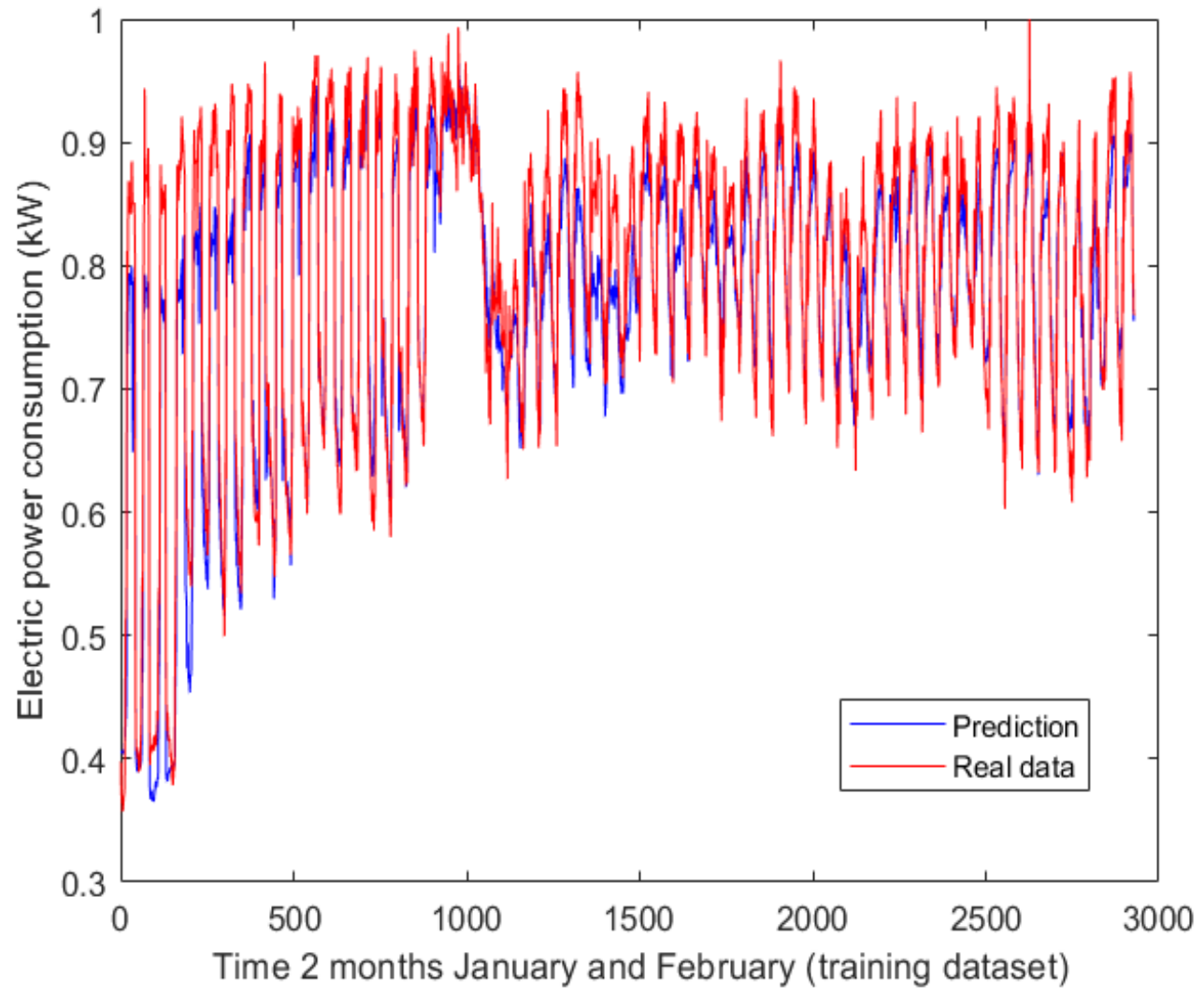

a) training results: SNR = 25.63dB dB/29.74 dB, NRMSE = 0.0433/0.0271 .

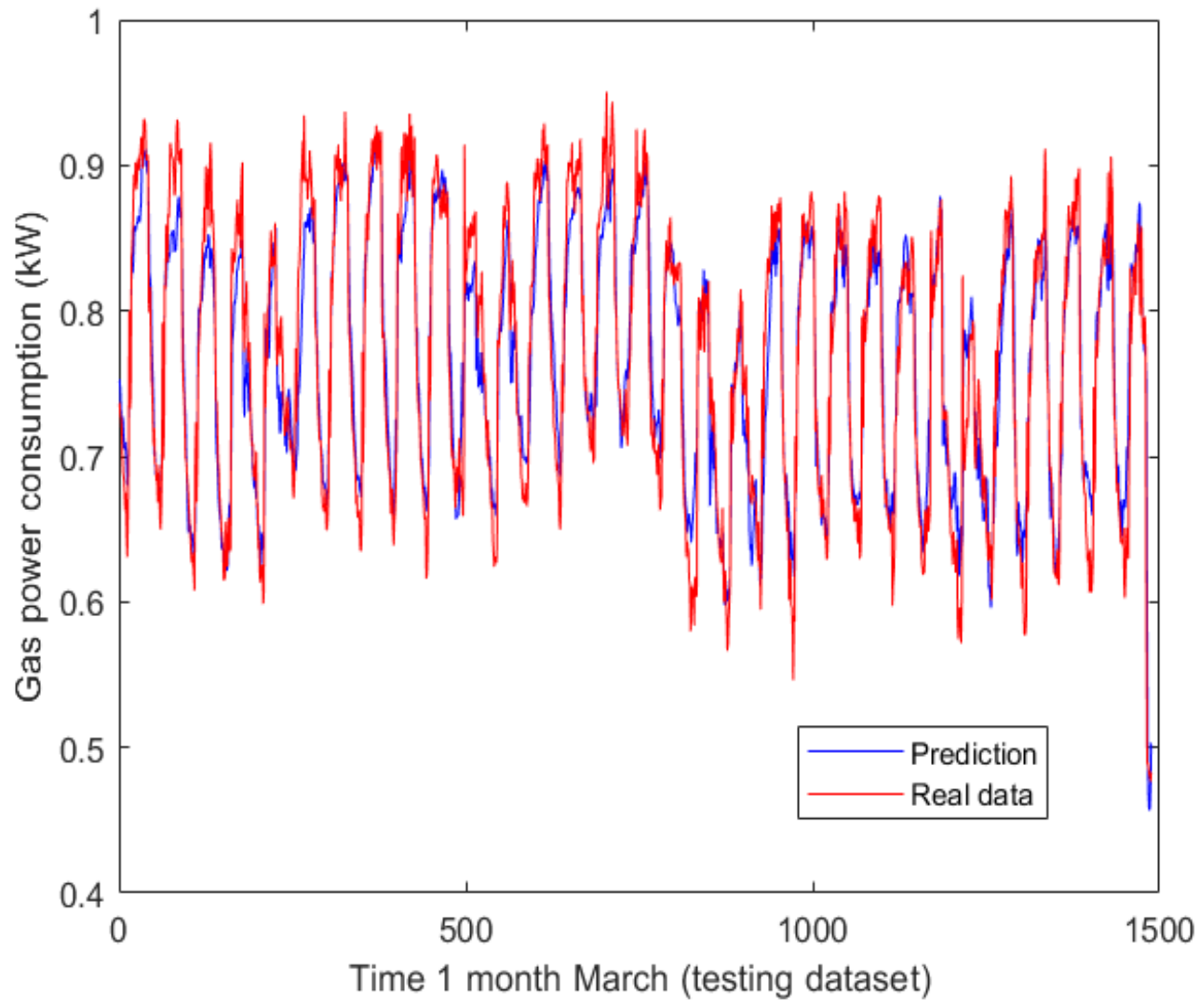

b) testing results: SNR = 27.39dB/28.84dB, NRMSE = 0.0347/0.029.

Fig. A11. Electrical power consumption prediction results (building 1) with input variables as the previous 24 hours electrical power consumptions and the weather temperatures (°C) and all input data being scaled between 0 and 1.

### *AA.6 RMSE and loss functions for training and validation datasets when using training, validation and testing datasets (option_3)*

In section IV.B1 there were presented the short-term CNN predictions when using training, validation and testing datasets. This method involved using a single input variable cosisting of the previous 24 hours power consumption, and predicting a single output variable consisting of the next half an hour power consumption of the same type as the input variable (i.e. electric-electric, heat-heat, gas-gas). For this method, it is also of interest to look to the RMSE and the loss functions for the training and the validation datasets, and as calculated by the DL toolbox (MATLAB), and which values should be also low. Therefore, Fig.A12 and Table A1 show these values during the training of two buildings and for the electrical energy vector. It can be observed that after 300 epochs, the mini-batch RMSE, the validation RMSE, the mini-batch loss and the validation loss start to stabilize around some small values.

| Ep. | Iter. | Time Elapsed (minutes) | Mini-batch RMSE | Valid. RMSE | Mini-batch Loss | Valid. Loss |
|---|---|---|---|---|---|---|
| 1 | 1 | 0.03 | 87.21 | 79.98 | 3802.9792 | 3198.2224 |
| 50 | 200 | 1:01 | 3.27 | 3.51 | 5.3452 | 6.1709 |
| 100 | 400 | 2:00 | 2.95 | 3.34 | 4.3632 | 5.5612 |
| 150 | 600 | 2.58 | 2.82 | 2.71 | 3.9730 | 3.6678 |
| 200 | 800 | 3:56 | 2.61 | 2.73 | 3.3992 | 3.7392 |
| 250 | 1000 | 4:55 | 2.35 | 3.11 | 2.7661 | 4.8469 |
| 300 | 1200 | 5:53 | 2.24 | 2.79 | 2.5192 | 3.8864 |
| 350 | 1400 | 6:51 | 2.15 | 3.20 | 2.3039 | 5.1277 |
| 400 | 1600 | 7:52 | 2.30 | 2.66 | 2.6531 | 3.5492 |

a)

| Ep. | Iter. | Time Elapsed (minutes) | Mini-batch RMSE | Valid. RMSE | Mini-batch Loss | Valid. Loss |
|---|---|---|---|---|---|---|
| 1 | 1 | 0.02 | 55.13 | 47.52 | 1519.8113 | 1129.1021 |
| 50 | 200 | 0.52 | 1.85 | 1.91 | 1.7204 | 1.8146 |
| 100 | 400 | 1.53 | 1.56 | 1.73 | 1.2177 | 1.5038 |
| 150 | 600 | 2.50 | 1.53 | 1.86 | 1.1653 | 1.7297 |
| 200 | 800 | 3.47 | 1.39 | 1.80 | 0.9688 | 1.6186 |
| 250 | 1000 | 4.46 | 1.63 | 2.09 | 1.3315 | 2.1741 |
| 300 | 1200 | 5.44 | 1.15 | 1.62 | 0.6595 | 1.3198 |
| 350 | 1400 | 6.40 | 1.03 | 1.71 | 0.5346 | 1.4579 |
| 400 | 1600 | 7.42 | 1.05 | 1.73 | 0.5494 | 1.5017 |

b)

Table. A1. RMSE and loss functions for electrical training and validation datasets when using training, validation and testing datasets (option_3) and with regard to the number of epochs: a) building 2; b) building 1 (Ep-epoch, Iter-iteration).

Furthermore, in Table A2 there are shown the RMSE and the loss functions for training and validation datasets for heating predictions for a building (i.e. building 5).

| Ep. | Iter. | Time Elapsed (minutes) | Mini-batch RMSE | Valid. RMSE | Mini- batch Loss | Valid. Loss |
|---|---|---|---|---|---|---|
| 1 | 1 | 0.09 | 678.46 | 684.78 | 230152.8750 | 234465.0469 |
| 100 | 300 | 4.30 | 45.52 | 123.31 | 1035.8372 | 7603.0923 |
| 200 | 600 | 6.36 | 15.45 | 102.14 | 119.3028 | 5216.1431 |
| 300 | 900 | 10.15 | 14.24 | 111.26 | 101.3942 | 6189.5220 |
| 400 | 1200 | 11.54 | 6.27 | 107.73 | 19.6801 | 5802.4126 |
| 500 | 1500 | 16.08 | 6.31 | 101.19 | 19.9028 | 5119.8462 |
| 600 | 1800 | 18.10 | 4.96 | 105.70 | 12.2885 | 5586.4155 |
| 700 | 2100 | 20.00 | 6.89 | 95.19 | 23.7665 | 4530.8188 |
| 800 | 2400 | 21.52 | 3.76 | 95.29 | 7.0747 | 4539.6313 |
| 900 | 2700 | 23.48 | 2.73 | 96.44 | 3.7319 | 4649.9136 |
| 1000 | 3000 | 25.41 | 2.45 | 97.85 | 3.0086 | 4787.4126 |

Table. A2. RMSE and loss functions for heat training and validation dataset when using training, validation and testing datasets (option_3) and with regard to the number of epochs (building no. 5).

Although the final values in Table.A2 may seem higher this time, in reality in percentage values with regard to the initial values, they can still be regarded as very small values such as the mini-batch RMSE (0.003%), the mini-batch loss (0.00001%) and the validation loss (0.02%). Higher validation RMSE can be observed (14%) but this is because the RMSE is not shown normalized. This is confirmed also by Fig.A13, which shows some good predictions for both the training and the testing heating datasets obtained for this building (building no. 5) for which the RMSE and the loss functions are shown in Table A2.

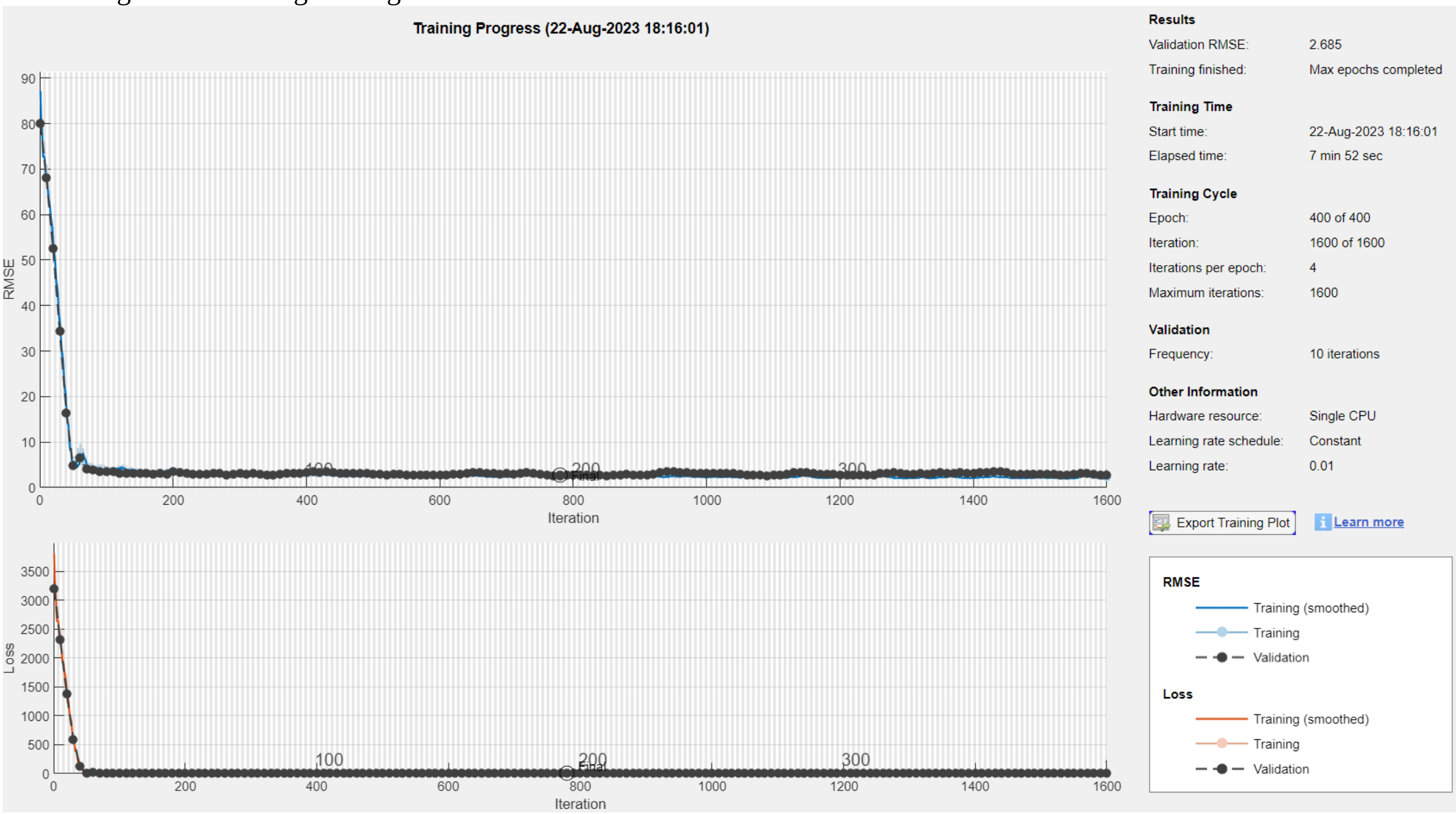

a)

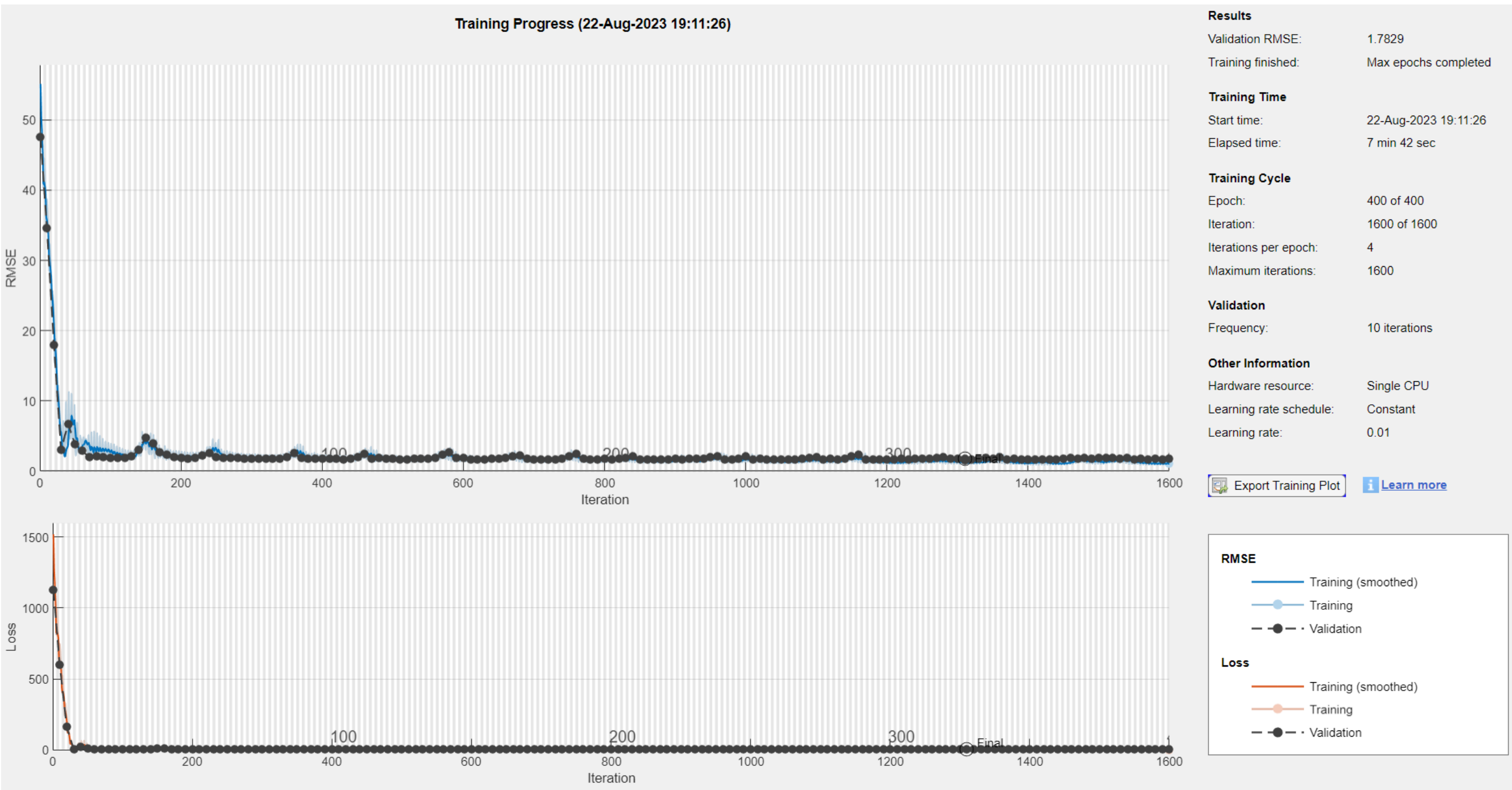

b)

Fig. A12. RMSE and loss functions for training and validation datasets during the training of two buildings and with regard to the number of epochs and iterations: a) building no. 2; b) building no. 1.

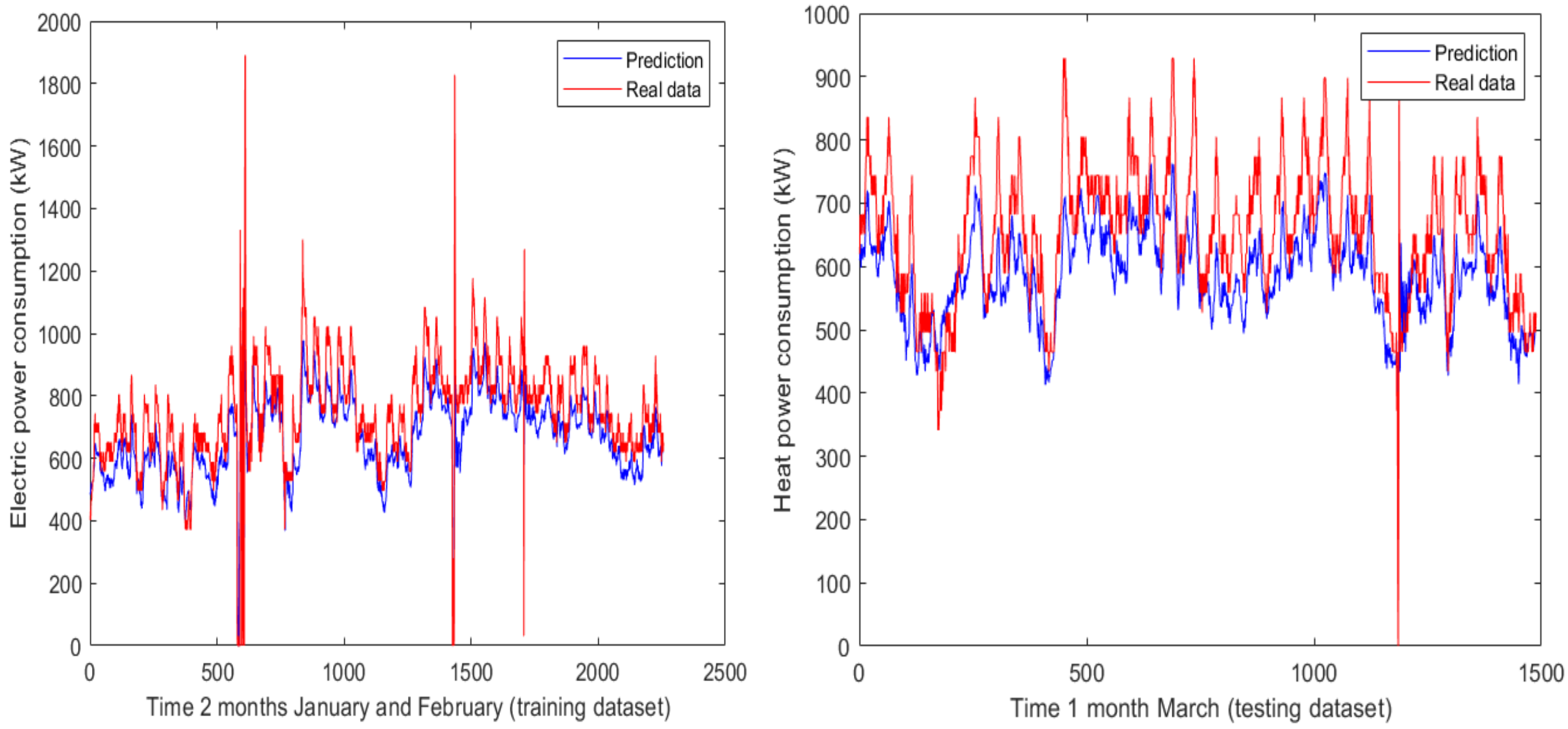

a) b)

Fig.A13. Training and testing predictions obtained using validation datasets for building no. 5: a) training (SNR =16.96 db , RMSE = 0.0565) ; b) testing (SNR =17.75 db , RMSE = 0.0453).